\documentclass{article}

\PassOptionsToPackage{numbers, compress}{natbib}

\usepackage[preprint]{neurips_2026}

\usepackage[utf8]{inputenc} 
\usepackage[T1]{fontenc}    
\usepackage{url}            
\usepackage{booktabs}       
\usepackage{amsfonts}       
\usepackage{nicefrac}       
\usepackage{microtype}      
\usepackage{xcolor}         
\usepackage[most]{tcolorbox}
\usepackage{multirow}
\usepackage{placeins}

\usepackage{colortbl}
\usepackage{siunitx}
\usepackage{xspace}
\usepackage{caption}
\usepackage{bm}
\usepackage{booktabs}
\usepackage{multirow}
\usepackage{colortbl}
\usepackage{xcolor}

\definecolor{aclblue}{rgb}{0, 0, 0.5}

\usepackage[
    colorlinks=true,
    linkcolor=aclblue,
    citecolor=aclblue,
    urlcolor=aclblue
]{hyperref}

\usepackage{cleveref} 

\usepackage{fontawesome5}

\definecolor{headerblue}{RGB}{220, 230, 245}
\definecolor{rowgray}{RGB}{247, 247, 247}
\definecolor{catcolor}{RGB}{235, 240, 250}

\newcommand{\posvalue}[1]{\textcolor{green!60!black}{+#1}}
\newcommand{\negvalue}[1]{\textcolor{red!70!black}{#1}}
\newcommand{\posvaluebad}[1]{\textcolor{red!70!black}{+#1}}
\newcommand{\negvaluegood}[1]{\textcolor{green!60!black}{#1}}
\newcommand{\neuvalue}[1]{#1}
\definecolor{easy}{RGB}{198,230,198}
\definecolor{hard}{RGB}{248,198,198}

\newcommand{\name}[1]{\textsc{UnpredictaBench}\xspace}

\definecolor{unpredictaNavy}{HTML}{001F5B}
\definecolor{unpredictaTeal}{HTML}{00A6B8}

\newcommand{\namecolored}{\textsc{\textcolor{unpredictaNavy}{Unpredicta}\textcolor{unpredictaTeal}{Bench}}}

\title{\namecolored{}: A Benchmark for Evaluating Distributional Randomness in LLMs}

\renewcommand{\thefootnote}{\fnsymbol{footnote}}

\author{%
  \textbf{Amirhossein Abaskohi}\textsuperscript{*\,\textdagger\,1} \quad
  \textbf{Amirhossein Dabiriaghdam}\textsuperscript{*\,1} \quad
  \textbf{Liang Luo}\textsuperscript{2} \\[3pt]
  \textbf{Ellie Dingqiao Wen}\textsuperscript{2} \quad
  \textbf{Lele Wang}\textsuperscript{1} \quad
  \textbf{Giuseppe Carenini}\textsuperscript{1} \quad
  \textbf{Peter West}\textsuperscript{1} \\[8pt]
  \textsuperscript{1}University of British Columbia \qquad
  \textsuperscript{2}Independent Researcher\\[6pt]
  {\small
  \textcolor{unpredictaTeal}{\faLink}\,
  \href{https://unpredictabenchmark.github.io/}{\texttt{\textcolor{unpredictaNavy}{UnpredictaBenchmark.GitHub.io}}}
  }
}

\newcounter{prompt}[section]
\renewcommand{\theprompt}{\thesection.\arabic{prompt}}

\newenvironment{promptbox}[2][]{%
    \refstepcounter{prompt}%
    \begin{tcolorbox}[
        colback=yellow!10!white,      
        colframe=yellow!60!orange,    
        fonttitle=\bfseries,       
        title={Prompt \theprompt: #2},
        enhanced,                  
        attach boxed title to top left={yshift=-2mm, xshift=5mm},
        boxed title style={colback=yellow!60!orange},
        sharp corners=south,       
        drop shadow,
        breakable,                 
        #1                         
    ]%
    \ttfamily\small
    \setlength{\baselineskip}{1.6\baselineskip}%
    \obeylines\obeyspaces\ignorespaces
}{%
    \end{tcolorbox}
}

\newtcolorbox[auto counter, number within=section]{examplebox}[2][]{%
    colback=gray!8!white,        
    colframe=gray!75!black,      
    fonttitle=\bfseries,
    fontupper=\small,
    title={Example \thetcbcounter: #2},
    enhanced,
    attach boxed title to top left={yshift=-2mm, xshift=5mm},
    boxed title style={colback=gray!75!black, coltitle=white}, 
    sharp corners=south,
    drop shadow,
    #1
}

\begin{document}

\maketitle

\renewcommand{\thefootnote}{\fnsymbol{footnote}}
\footnotetext[1]{Equal contribution.}
\footnotetext[2]{Corresponding author: \texttt{aabaskoh@cs.ubc.ca}}

\renewcommand{\thefootnote}{\arabic{footnote}}
\setcounter{footnote}{0}

\begin{abstract}

We introduce \textbf{\textsc{UnpredictaBench}}, an evaluation that tests the ability of large language models (LLMs) to capture true underlying distributions. As LLMs are increasingly used as substitutes for other entities (e.g., for humans in economic simulations), the tendency of many models to collapse towards a single plausible answer means a failure to capture the \emph{unpredictability} of real systems. Recent work on improving output diversity is insufficient for this setting: simulation requires samples that are calibrated to a target distribution, not merely varied outputs. \textsc{UnpredictaBench} isolates a simplified but fundamental version of this problem: sampling outcomes from individual target distributions, including canonical statistical distributions, distributions induced by stochastic programs, and natural-language scenarios that describe random processes. We introduce 448 such problems together with $KS@N$, a general-purpose evaluation metric that quantifies how well a model outputs approximate black-box target distributions via the Kolmogorov-Smirnov statistical test. This is the rate at which we fail to reject model samples of size N against ground-truth samples, with larger N indicating greater difficulty. Tested across open and proprietary models, we find a large spread in distributional capabilities.
For instance, when models generate samples of size 100 ($KS@100$, our standard metric), scores range from near 0 to over 20\%. No model is able to achieve over 40\% at $KS@100$, showing significant headroom in distributional sampling as a capability. Although adding reasoning can somewhat increase scores, we find no immediate solution for this issue. \textsc{UnpredictaBench} shows that even simple distributional simulation remains challenging, making it a necessary first step toward using LLMs as stand-ins for complex systems
\footnote{
Dataset is available on 
\href{https://huggingface.co/datasets/UnpredictaBench/UnpredictaBench}
{\raisebox{-0.15em}{\includegraphics[height=1em]{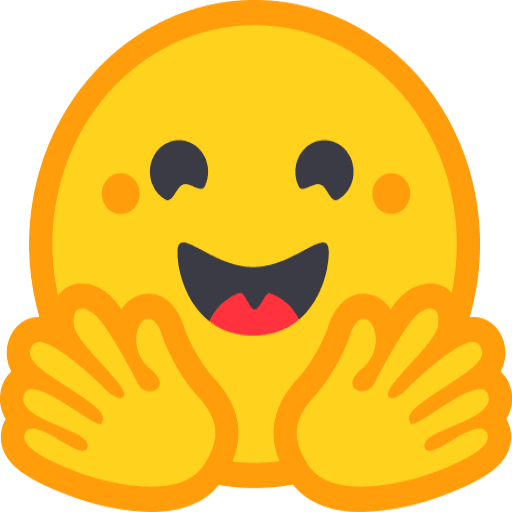}}~Hugging Face}, 
with code and ground-truth values released on 
\href{https://github.com/UnpredictaBench/UnpredictaBenchCode}
{{\textcolor{black}{\faGithub}}~GitHub}.
}. 

\end{abstract}
\section{Introduction}
\label{sec:intro}

\begin{figure}[t]
    \centering
    \caption{\textbf{(a)} Most models fail to reproduce target distributions, either lacking distributional understanding or collapsing to a narrow output range. \textsc{Nemotron-3-Super-120B}~\cite{nvidia_nemotron_3_2025} is a notable exception, capturing the multimodal Skellam structure reasonably well, whereas \textsc{OLMo-3-7B}~\citep{olmo2025olmo} places nearly all mass near zero despite the true Poisson distribution extending well beyond 20. \textbf{(b)} Since real-world systems are stochastic, applications such as economic simulation and epidemiological modeling require LLMs to reproduce randomness faithfully; distributional mismatch can yield biased estimates, overconfident predictions, and misleading conclusions.}
    \label{fig:teaser}
    \includegraphics[width=\linewidth]{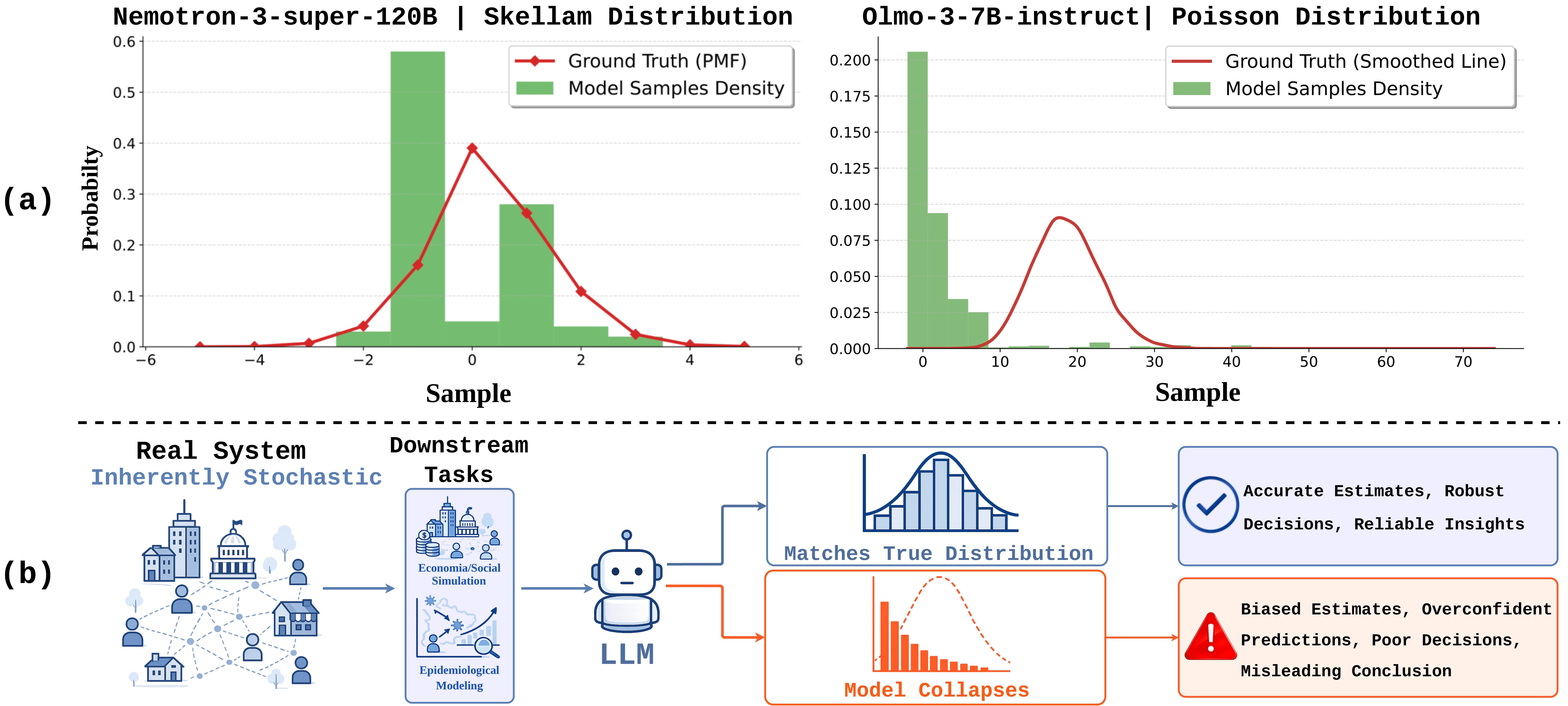}
\end{figure}

Randomness and uncertainty are core aspects of many fields of knowledge--physics, biology, statistics, and even human behavior--and although large language models (LLMs) can reason about randomness \citep{paruchuri-etal-2024-odds}, it is not clear how well they can produce it. This is particularly important as these models are increasingly used as stand-ins to simulate other systems~\cite{10.1145/3586183.3606763, pmlr-v205-ichter23a, huang2022inner}, making predictions about physical outcomes or modeling human interactions (see Figure~\hyperref[fig:teaser]{\ref*{fig:teaser}(b)}). In order for these applications to work, models must produce uncertain outcomes that are \emph{calibrated} to the underlying process, although their ability to do this is not well evaluated. 
Recent work suggests that LLMs can partially reason about distributions when estimating probabilities or percentiles~\citep{paruchuri-etal-2024-odds}, but this does not translate to faithful stochastic generation. Prior studies show failures in behavioral simulation~\citep{gu-etal-2025-llms}, real-world distribution modeling~\citep{Pleko2025EpidemiologyOL}, mixed-strategy games~\citep{guo-etal-2025-illusion}, and even simple random tasks such as coin flips, dice rolls, and random integers~\citep{van2024random,hopkins2023can,zhao2026large,coronado2025deterministic}.
Towards a systematic evaluation of this question, we introduce \textbf{\name{}}, a benchmark to test distributional randomness in LLMs. 

Verifying the stochastic correctness of LLMs in general requires broad progress in evaluation, and so the goal of \name{} is to test whether models can capture even a simple version of this problem: sampling from direct, single-output distributions. The benchmark is composed of 448 known distributions, stochastic code problems, and word problems. These include unimodal and multimodal distributions, real-world problems (e.g. race condition in multi-threading), and list shuffling. Models are tasked with generating independent samples, and evaluated with a new metric, $\bm{KS@N}$. Simply, $KS@N$ aims to capture a notion of distributional accuracy, based on the rate at which model samples are rejected against a black-box sample from the true distribution by a Kolmogorov-Smirnov test~\citep{kolmogorov1933sulla, smirnov1948table} with a fixed threshold. Increasing N naturally increases difficulty, and $KS@N$ requires only \emph{samples} from the ground truth. 

Evaluating a range of open and frontier models on \name{}, we observe high variance in performance. 
No model surpasses 40\% for $KS@100$ (our default setting), and most models spread their accuracy between 0\% and 20\%, indicating that generating a plausible sample of size 100 remains a significant challenge across the board. \texttt{Nemotron-3-super-120b-a12b}~\cite{nvidia_nemotron_3_2025} consistently ranks among the top performers across various $KS@N$ levels, whereas models like \texttt{GPT-5.4}~\cite{openai_gpt54_2026} and \texttt{Claude-sonnet-4.6}~\cite{anthropic_sonnet46_2026} average only 15.18\% and 4.7\% across all tasks, respectively—lower than much smaller open-source models such as \texttt{Qwen-3.5-2B}~\cite{qwen3.5}, which achieves 17.67\%. 
We see similar trends in related metrics including Wasserstein Distance and Jensen–Shannon divergence~\cite{lin2002divergence}. Qualitatively, we find a range of model failures, from collapse onto a reasonable mode, to total miscalibration to the true distribution (Figure~\hyperref[fig:teaser]{\ref*{fig:teaser}(a)}). Interventions such as reasoning can help, but are far from solving the problem. 
In terms of benchmark difficulty, tasks requiring models to infer the underlying distribution from code and shuffling tasks prove the most challenging, with several strong overall performers collapsing to 0\% on the latter.
\name{} accuracy correlates strongly with utility metrics from 
NoveltyBench~\cite{zhang2025noveltybench} and CREATE~\cite{wadhwa2026create}, 
confirming that distributional fidelity captures a genuine notion of model quality 
while offering a statistically grounded alternative to LLM-as-a-judge 
evaluation~\cite{zheng2023judging}.

\name{} is a first step in understanding, evaluating, and improving the ability of LLMs to capture complex sources of randomness. We should not yet expect LLMs to capture more complex distributions, such as human behavior, given their struggle in this simple setting. This benchmark also offers a roadmap for future work in this area, naturally providing increasingly difficult versions through modifications such as an increase in sample size, and providing a template for future benchmarks that can reuse elements such as $KS@N$. Overall, our \textbf{contributions} are as follows:
\textbf{(i)} We introduce \textbf{\name{}}, a benchmark of 448 test instances covering 
40 target distributions across unimodal and multimodal settings with a diverse task suite spanning textual, code, real-world, and shuffling 
scenarios, evaluating distributional randomness beyond simple numeric prompting.
\textbf{(ii)} We propose $KS@N$, a repeated-generation evaluation metric that compares empirical model outputs against ground-truth distributions, assessing stochastic fidelity rather than one-off correctness.
\textbf{(iii)} We provide a first systematic analysis of LLMs as statistical random generators across a wide range of distributions and prompting conditions, offering a unified testbed for future work on randomness and distributional generation.
\section{Related Work}
\label{sec:rw}

\textbf{Probabilistic reasoning and randomness generation.}
Prior work establishes that LLMs can perform non-trivial probabilistic reasoning with contextual support~\citep{paruchuri-etal-2024-odds}, but a consistent finding is that reasoning \emph{about} a distribution does not translate to faithfully \emph{generating from} it. \citet{gu-etal-2025-llms} show that LLMs can identify probabilistic structure but fail to sample from it accurately, \citet{Pleko2025EpidemiologyOL} show that LLMs do not faithfully encode real-world observational distributions, and \citet{Zhang2025PredictingEM} demonstrate that performance deteriorates when latent distributions must be inferred. During generation, LLMs fail even in simple settings such as uniform random number generation~\citep{hopkins2023can}, with outputs reflecting human-like biases rather than true randomness~\citep{van2024random,zhao2026large}. \citet{coronado2025deterministic} provide a broad empirical study showing model outputs are often surprisingly deterministic and biased toward specific values, and \citet{guo-etal-2025-illusion} demonstrate a cognition--behavior gap in strategic settings: models can state the correct mixed strategy yet their actual choices remain biased. Most directly related to our work, \citet{gu2026illusion} show that while frontier models can convert provided random seeds to target distributions, their ability to sample directly from specified categorical distributions is fundamentally flawed. \name{} differs from all of these by providing a unified benchmark over many distributions and tasks rather than focusing on any single setting.

\textbf{Alignment, uncertainty, and behavioral factors.}
Another body of work investigates why models exhibit poor stochastic behavior. Post-training is a key culprit: \citet{west2025base} show that base models outperform aligned models on random number generation and creativity, \citet{li2025preserving} show that cross-entropy fine-tuning systematically reduces output diversity, and \citet{zhang2026embarrassinglysimpleselfdistillationimproves} show that fine-tuning on temperature-shifted self-samples can partially recover it. Beyond training, prompt structure can heavily condition apparent stochastic behavior~\citep{bigelow2024incontext}. 
On uncertainty calibration, raw model confidence is often poorly calibrated~\citep{shi-etal-2025-reasoning} and structured by semantic similarity between candidate responses~\citep{mccabe2025estimating}. Finally, \citet{cao-etal-2025-specializing} show that fine-tuning can improve alignment with human opinion distributions in social simulation, but persistent diversity reduction remains. These findings motivate\name{}'s repeated-output evaluation: the goal is not simply to elicit diverse responses, but to testwhether model outputs are calibrated to a target distribution.

\section{\name{}}
\label{sec:benchmark}

\begin{figure}[t]
    \centering 
    \caption{
        \textbf{\name{} Pipeline.} \textbf{(a) Data Generation.} Instances are constructed from two sources: 40 distributions selected from Wikipedia, from which \texttt{GPT-5.4}~\cite{openai_gpt54_2026} generates tasks across 7 categories; and 50 human-curated real-world stochastic tasks. \textbf{(b) Evaluation.} Each task is evaluated by querying the model $100$ times independently and comparing the empirical output distribution against a ground-truth reference using three metrics.}
    \includegraphics[width=1.01\linewidth]{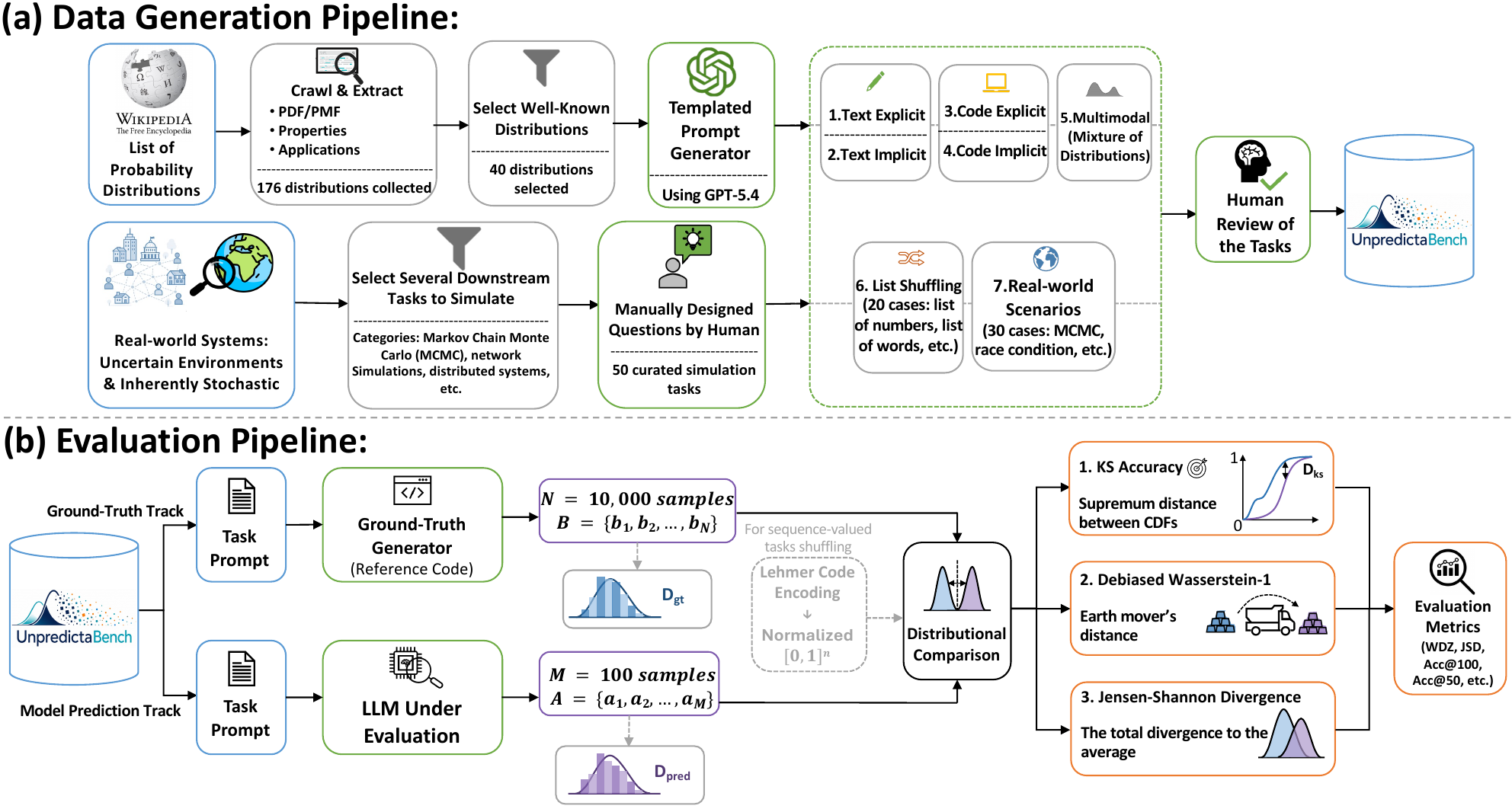}
    \label{fig:piepline}
\end{figure}

In this section, we describe the construction of \textbf{\name{}} and summarize its task design, statistics, and our evaluation strategy, as illustrated in Figure~\ref{fig:piepline}. Our goal is to evaluate whether language models can \emph{generate outputs consistent with target probability distributions}, rather than simply recognize or describe them.

\subsection{Benchmark Construction and Task Types}
\label{subsec:benchmark_creation}

We first crawled probability distributions from Wikipedia\footnote{\url{https://en.wikipedia.org/wiki/List_of_probability_distributions}}. For each distribution, we extracted detailed information including the probability density/mass function, mean, mode, median, real-world applications, and key statistical properties. In total, we collected \textbf{176 distributions}. From this pool, we selected \textbf{40 well-known distributions} (see Table~\ref{tab:distributions} in Appendix~\ref{app:distributions} for the full list of distributions), as our benchmark targets general-purpose language models rather than expert statisticians. These distributions form the basis of all benchmark tasks. 
To construct the benchmark instances, we use a templated generation pipeline where distribution information is passed to \texttt{GPT-5.4}~\cite{openai_gpt54_2026} to produce prompts across different task types. For each automatically generated task, the prompt also specifies distribution hyperparameters, chosen to \textbf{cover both concentrated and spread-out regimes}. This allows the benchmark to test whether models can adapt not only to different distribution families, but also to different parameterizations of the same distribution.
In addition, \textbf{50 tasks were manually constructed by a single annotator}: 30 Real-World Scenario tasks and 20 Shuffling tasks. All 450 generated and manually constructed tasks were then reviewed by two independent annotators, resulting in the removal of 2 tasks that failed quality checks, yielding a final benchmark of \textbf{448 instances}. The exact prompt templates used for generation and answer extraction are provided in Appendix~\ref{app:prompts}. \name{} contains seven task categories, designed to probe distributional understanding across varied representations and difficulty levels. 

\textbf{Textual Tasks: \textbf{(1)} Text Explicit and \textbf{(2)} Text Implicit.} Textual tasks present distributions in natural language. In explicit tasks, the distribution and its parameters are fully named and the model is asked to generate a sample directly. In implicit tasks, a real-world scenario is described without naming the underlying distribution, requiring the model to infer the stochastic process before sampling. Prompt templates are given in Prompts~\ref{prompt:text-explicit-c}--\ref{prompt:text-implicit-s}.

\begin{examplebox}{Text Explicit (left) and Text Implicit (right)}
\begin{minipage}[t]{0.48\linewidth}
Generate a random sample from a Poisson distribution with rate parameter $\lambda = 18$, 
representing the number of events observed in one fixed interval.
\end{minipage}
\hfill
\begin{minipage}[t]{0.48\linewidth}
A quality-control engineer inspects 6 components, each independently failing with 
probabilities 0.04, 0.05, 0.03, 0.06, 0.04, and 0.05. What is one possible total 
number of failed components in a single inspection?
\end{minipage}
\end{examplebox}

\textbf{Code Tasks: \textbf{(3)} Code Explicit and \textbf{(4)} Code Implicit.}
Code tasks require the model to predict a possible output of a stochastic Python 
program. In explicit tasks, the distribution is sampled directly via \texttt{NumPy\footnote{\url{https://numpy.org/}}}. In implicit 
tasks, the target distribution is implemented indirectly through transformations such 
as square roots, trigonometric functions, or summations applied to samples from a 
different distribution, requiring deeper reasoning about the underlying stochastic 
process. Prompt templates are given in 
Prompts~\ref{prompt:code-explicit-c}--\ref{prompt:code-implicit-s}.

\begin{examplebox}{Code Explicit (left) and Code Implicit (right)}
\begin{minipage}[t]{0.48\linewidth}
\begin{verbatim}
import numpy as np
a = 2.0; b = 2.5
u = np.random.uniform(0.0, 1.0)
sample = a * (b / a) ** u
print(sample)
\end{verbatim}
\end{minipage}
\hfill
\begin{minipage}[t]{0.48\linewidth}
\begin{verbatim}
import numpy as np
rng = np.random.default_rng()
x = rng.gamma(shape=0.6, scale=1.0)
y = rng.gamma(shape=0.6, scale=1.0)
outcome = x / (x + y)
print(float(outcome))
\end{verbatim}
\end{minipage}
\end{examplebox}

\textbf{\textbf{(5)} Multimodal Tasks.}
Multimodal tasks require sampling from distributions formed by combining two or more 
component distributions, constructed from 20 highly recognizable distributions (refer to Appendix~\ref{app:distributions}) in our 
set via mixture sampling or additive combinations. These tasks evaluate whether models 
can maintain multi-modal coverage rather than collapsing to a single mode. Prompt 
templates are given in Prompts~\ref{prompt:multimodal-c} and~\ref{prompt:multimodal-s}.

\begin{examplebox}{Multimodal Example}
Generate one random sample from a 2-component mixture of exponential distributions: 
with probability 0.55, draw from Exponential($\lambda{=}8.0$); with probability 0.45, 
draw from Exponential($\lambda{=}1.6$). What is the sampled value?
\end{examplebox}

\textbf{\textbf{(6)} Shuffling Tasks.}
Shuffling tasks evaluate permutation-level randomness by asking the model to produce a 
uniform random shuffle of a given list of up to five elements. Lists span four types: 
numerical values, counting words (e.g., first, second), arbitrary words, and mixed 
lists. Outputs are encoded via Lehmer codes prior to evaluation 
(Section~\ref{subsec:lehmer}).

\begin{examplebox}{Shuffling Example}
Considering this list: \texttt{["first", "second", "third"]}, what is one possible 
uniform random shuffle? Respond with exactly one list only.
\end{examplebox}

\textbf{\textbf{(7)} Real-World Scenario Tasks.}
To evaluate whether models can simulate inherently uncertain environments, we include 
30 manually curated real-world scenario tasks covering six categories of practical 
nondeterminism: \textbf{(i)} MCMC sampling dynamics, \textbf{(ii)} multi-outcome decision-making, 
\textbf{(iii)} race conditions and multi-threaded execution, \textbf{(iv)} hashing and collision 
behavior, \textbf{(v)} network simulations with stochastic delays, and \textbf{(vi)} distributed systems 
with asynchronous communication. These tasks require models to implicitly reason about 
underlying stochastic processes rather than simply pattern-match to a named 
distribution. Examples are provided in Appendix~\ref{app:realworld_examples}.

\subsection{Statistics}
\label{subsec:stats}

Table~\ref{tab:unpredictabench-overview} summarizes the key statistics of 
\name{}. The benchmark comprises \textbf{448 instances} in English, 
of which \textbf{398 are \texttt{GPT-5.4}-authored} and \textbf{50 are human-authored}. Of the 
398 automatically generated tasks, half use concentrated parameter settings and half 
use spread out parameter settings, \textbf{80 are multimodal} while the remaining 
\textbf{318 are unimodal}, and the distribution across task types is: 
\textbf{159 Text Explicit}, \textbf{79 Text Implicit}, \textbf{80 Code Explicit}, and 
\textbf{80 Code Implicit}. The 30 human-authored Real-World tasks span six categories: 
OS concurrency (6), garbage collection (6), network simulations (5), distributed 
systems (5), hashing (4), and MCMC (4). The 20 Shuffling tasks cover four list types: 
integer (6), ordinal (6), word (5), and decimal (3) lists, with an average list length 
of \textbf{2.95 elements}.
The right panel of 
Table~\ref{tab:unpredictabench-overview} shows the coverage of target distribution.

\begin{table}[t]
\centering
\small
\caption{Overview of \name{}. \textbf{Left:} dataset-level statistics.\textbf{ Right:} category coverage of the probability distributions used in the benchmark.}
\label{tab:unpredictabench-overview}

\begin{minipage}[t]{0.46\textwidth}
\centering
\captionsetup{type=table}
\begin{tabular}{@{}lr@{}}
\toprule
\textbf{Statistic} & \textbf{Value} \\
\midrule
Instances & 448 \\
Language & English \\
Human-authored prompts & 50 \\
GPT-5.4-authored prompts & 398 \\
Min prompt length (char) & 164 \\
Median prompt length (char) & 501 \\
Mean prompt length (char) & 513.7 \\
Max prompt length (char) & 1788 \\
\bottomrule
\end{tabular}
\end{minipage}
\hfill
\begin{minipage}[t]{0.50\textwidth}
\centering
\captionsetup{type=table}
\begin{tabular}{@{}lr@{}}
\toprule
\textbf{Target Dist. Category} & \textbf{Count} \\
\midrule
Abs.\ continuous, semi-infinite & 11 \\
Abs.\ continuous, bounded interval & 6 \\
Abs.\ continuous, whole real line & 5 \\
Discrete, finite support & 6 \\
Discrete, infinite support & 5 \\
Joint distributions & 5 \\
Mixed discrete--continuous & 1 \\
Non-numeric & 1 \\
\bottomrule
\end{tabular}
\end{minipage}
\vspace{-5pt}
\end{table}
\subsection{Evaluation Strategy}
\label{sec:evaluation_strategy}

To assess how well a model reproduces a target distribution, we compare a set $A$ of $N=100$ independent samples drawn from the model's predictive distribution $\mathcal{D}_{\mathrm{pred}}$ against a reference set $B$ of $M=10{,}000$ samples drawn from the ground-truth distribution $\mathcal{D}_{\mathrm{gt}}$. Because the reference set is itself sampled, the evaluation could in principle depend on the particular ground-truth draw used for comparison. We therefore conduct a sensitivity analysis in Appendix~\ref{app:ground_truth_sensitivity}, where we repeat the evaluation with multiple independently sampled reference sets and confirm that the results are stable.


For sequence-valued tasks such as shuffling, we first encode each permutation $\pi$ via its \emph{Lehmer code}~\citep{Lehmer1960TeachingCT} $L(\pi)$, a bijective mapping from permutations to integer sequences that preserves all ordering information, and normalize each coordinate to $[0,1]$. We then use the first coordinate $Z_1(\pi)$ as a scalar proxy for the permutation distribution, enabling direct application of our scalar metrics. We focus on $Z_1$ because the first Lehmer coordinate has the largest support: for a permutation of length $n$, $L_1$ can take $n$ distinct values, whereas later coordinates have progressively smaller support. As a result, matching the distribution of $Z_1$ is a stricter one-dimensional diagnostic than matching later coordinates, since it requires the model to reproduce a richer marginal distribution over possible initial ranks. While no single coordinate fully characterizes the joint distribution over permutations, $Z_1$ provides a challenging and interpretable scalar summary of ordering behavior, making it suitable for comparison with our scalar-valued tasks. Full details of the Lehmer encoding are provided in Appendix~\ref{subsec:lehmer}.

Our primary evaluation metric is $\bm{KS@N}$, which we treat as an \textbf{accuracy metric} for distributional fidelity. For each problem $i$ in a set of $l$ stochastic tasks, we apply a two-sample Kolmogorov--Smirnov test between $A$ and $B$ and obtain a $p$-value $p_{\mathrm{ks},i}$. We then define:
\begin{equation}
\small
\mathrm{KS@}N = \frac{1}{l}\sum_{i=1}^{l} \mathbf{1}\left[p_{\mathrm{ks},i} \geq 
p_{\mathrm{threshold}}\right],
\end{equation}
the fraction of problems for which the model's samples are \emph{not} rejected as inconsistent with the ground truth. We set $p_{\mathrm{threshold}} = 0.0001$ to ensure a low false-negative rate, and verify that using the true distribution as $\mathcal{D}_{\mathrm{pred}}$ achieves $\mathrm{KS@}N = 1.0$ across all values of $N$ considered. Larger $N$ increases difficulty by demanding closer calibration to the true distribution. We additionally report two complementary metrics: the Debiased Wasserstein-1 Distance Z-score (WDZ), which expresses the observed earth mover's distance in standard deviations above the permutation null baseline and captures tail behavior and systematic shifts in location and scale; and Jensen--Shannon Divergence (JSD), which captures density-level shape mismatches. See Appendix~\ref{app:metrics} for their full definitions.
\section{Experiments and Results}
\label{sec:experiments}


\subsection{Experimental Settings}

\noindent
\textbf{Model. } In this study, we evaluate a diverse set of models spanning multiple architectures and scales, covering both open-weight and proprietary systems.
Open-weight families include {OLMo-3}~\cite{olmo2025olmo};
{Qwen-3}~\cite{qwen3technicalreport};
{Qwen-3.5}~\cite{qwen3.5};
{Nemotron-3}~\cite{nvidia_nemotron_3_2025};
{Ministral-3}~\cite{mistralai_mistral3_2025};
{Llama-3.1}, and {Llama-3.2}~\cite{grattafiori2024llama3herdmodels};
{Phi-3.5}~\cite{microsoft_phi35_2024};
and {DeepSeek-v3.2}~\cite{deepseekai2025deepseekv32}.
Proprietary models include {Claude-sonnet-4.6}~\cite{anthropic_sonnet46_2026};
{GPT-4o}~\cite{openai_gpt4o_2024};
{GPT-5.4}~\cite{openai_gpt54_2026};
{Mercury-2}~\cite{inceptionlabs_mercury2_2026};
and {Grok-4.1-fast}~\cite{xai_grok41_2025}. 

\noindent
\textbf{Sampling and Generation Settings. }
In all experiments, we use a fixed temperature of $T{=}1.0$ unless otherwise specified. Temperature $T{=}1.0$ is a natural default as it preserves the model's trained output distribution without artificially concentrating or flattening it. Reasoning is disabled by setting the reasoning effort to \texttt{none} except in the reasoning experiments (Section~\ref{sec:exp:reasoning}), where we set reasoning effort to \texttt{xhigh}, allocating up to 95\% of tokens for reasoning with a maximum of 4,096 tokens. Each model is queried independently 100 times per problem instance with \texttt{max\_tokens=64} for standard, Shuffling, and real-world tasks. For the list prompting ablation (Section~\ref{sec:ablations:list}), we set \texttt{max\_tokens=640} and \texttt{max\_tokens=2512} when requesting 10 and 35 elements, respectively. For Shuffling and real-world tasks, prompts are specified per problem and bundled with each task in the benchmark, since these tasks differ in style. Text and code tasks instead use two static prompts, given in Prompt~\ref{prompt:text-sampling} and Prompt~\ref{prompt:code-sampling}.

\noindent
\textbf{Answer Parsing and Retry Protocol. }
For answer extraction, models are prompted to return their answer in a structured format: a number, string, or list enclosed in \texttt{\{\{asked\_value\}\}} depending on the task type, enabling reliable parsing. If extraction fails, we retry using \texttt{GPT-4o-mini} as a fallback extractor (refer to Prompts~\ref{prompt:extractor}-\ref{prompt:extractor_realworld} for the instructions used for extraction). If the model fails to produce a valid output (e.g., omitting the final value or returning a malformed list) across 5 consecutive retries, that run is skipped for that problem instance. For the reasoning token extraction and list prompting experiments, model calls are repeated and outputs accumulated until at least 100 values are collected; if more than 100 are obtained, only the first 100 are used.

\noindent
\textbf{Evaluation Infrastructure and Cost. }
All primary experiments were conducted via OpenRouter\footnote{\href{https://openrouter.ai/}{https://openrouter.ai}}, a cloud-based model aggregation platform providing unified API access to open and proprietary models. A small number of models unavailable on OpenRouter were evaluated locally on a workstation equipped with an Intel Core i9 CPU, 64\,GB of RAM, and an NVIDIA RTX 3090 GPU with 24\,GB VRAM. Local-only models include: \texttt{Llama-3.2-1B-instruct}, \texttt{Phi-3.5-mini-instruct}, \texttt{Qwen3.5-2B}, \texttt{OLMo-3-7B-instruct}, and \texttt{Ministral-3-3B-instruct-2512}. Each individual experiment required approximately 1--10 minutes of wall-clock time via OpenRouter, with variation attributable to cloud provider load, model size, and architecture. The total API cost across all reported experiments was approximately \$300\,USD.

\subsection{Overall Model Performance on UnpredictaBench}

\begin{figure}[t]
    \centering
    \caption{KS@100 (\%) of all evaluated models grouped by model family. Each bar represents a single model, color-coded by its originating family (see legend). \textbf{Nemotron-3 Super 120B leads all models at 32.64\%}, with a substantial drop-off to the next tier.}
    \label{fig:model_accuracy_overall}
    \includegraphics[width=\linewidth]{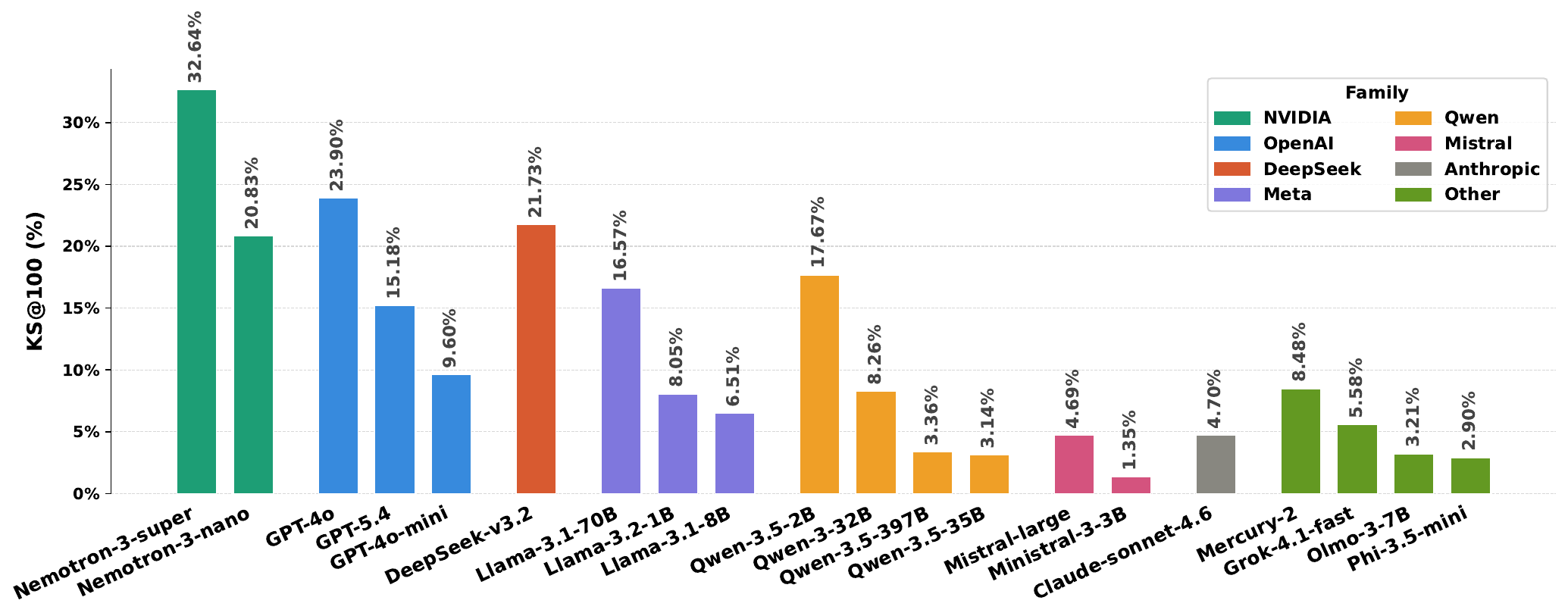}
    \vspace{-20pt}
\end{figure}

Figure~\ref{fig:model_accuracy_overall} presents the KS@100 scores for all models evaluated, grouped by model family. The results reveal a striking performance gap between the top-performing systems and the broader field. \textbf{Nemotron-3 Super 120B achieves the highest KS@100 at 32.64\%, nearly doubling the score of the third-ranked model}, underscoring the advantage conferred by scale within the NVIDIA family, where even the smaller Nemotron-3 Nano 30B (20.83\%) remains competitive with frontier models. Among frontier models, GPT-4o (23.90\%) and DeepSeek V3.2 (21.73\%) form a tight cluster immediately behind the NVIDIA leaders, while GPT-5.4 (15.18\%) and GPT-4o Mini (9.60\%) trail considerably, suggesting that model tier within a family matters as much as the family itself. The open-weight Llama 3.1 70B (16.57\%) and Qwen3.5 2B (17.67\%) are noteworthy: the latter in particular demonstrates that \textbf{a compact 2B model can rival systems an order of magnitude larger}, hinting at the outsized role of training data and instruction tuning over raw parameter count. At the lower end of the spectrum, several models cluster below 5\%, including Ministral-3 3B (1.35\%), Phi-3.5 Mini (2.90\%), and OLMo-3 7B (3.21\%). Surprisingly, \textbf{Claude Sonnet 4.6 (4.70\%) and Mistral Large 2512 (4.69\%) fall into this lower tier despite their considerable sizes}, which may reflect a mismatch between our benchmark's task distribution and the optimization objectives of these models. We hypothesize 
that this is because MoE routing tends to activate a sparse and consistent subset of 
experts for a given input type, effectively reducing the diversity of computation paths 
and producing outputs that are no more varied than those of a much smaller dense model.

\textbf{Category-Level Analysis. } Table~\ref{tab:main} presents a fine-grained breakdown of model performance across four task categories (\textbf{Code}, \textbf{Text}, \textbf{RealWorld}, and \textbf{Shuffling}), alongside Jensen--Shannon Divergence (JSD) and the Wasserstein Distance Z-score (WDZ). Crucially, \textbf{KS@100 is broadly consistent with both JSD and WDZ across models and categories}: models scoring higher on KS@100 consistently exhibit low JSD and WDZ as well, validating that the \textbf{KS-based metric captures genuine distributional alignment}. JSD captures global distributional similarity, while WDZ emphasizes tail behavior; both align with the KS@100 ranking, supporting metric robustness. Evaluation stability across three repeated runs is reported in 
Appendix~\ref{app:error_analysis}, where we analyze run-to-run variance using the 
standard deviation across runs.

\textbf{Performance across Sample Sizes.} 
Table~\ref{tab:acc_ks} reports KS@N across seven evaluation thresholds ($N \in \{1, 2, 5, 10, 20, 50, 100\}$) for all evaluated models. All models achieve perfect KS@1, confirming that every model can produce at least one plausible value within the support of the target distribution. Performance degrades almost monotonically as $N$ increases, reflecting the growing statistical difficulty of producing a full sample that is indistinguishable from the ground truth. The drop is particularly steep between KS@20 and KS@100 for most models, with the gap between the strongest model (Nemotron-3 Super 120B, 32.64\%) and the weakest (Qwen-3.5-35B-a3b and Phi-3.5-mini-instruct, both at 2.90\%--3.14\%) widening considerably at larger $N$. Notably, Claude Sonnet 4.6 achieves a strong KS@2 (97.76\%) but falls sharply to 4.70\% at KS@100, one of the steepest drop-offs in the table, further evidence that \textbf{single-sample plausibility is a poor proxy for distributional fidelity}.

Shuffling and Code tasks emerge as the most demanding categories, with no model exceeding 40\% in either. In Shuffling, \textbf{several strong overall performers collapse to 0\%} (GPT-5.4, Mercury-2, Claude Sonnet 4.6, Qwen3.5-35B-a3b), while DeepSeek V3.2, OLMo-3 7B, Llama-3.2-1B, and Qwen3.5 2B all sustain $\sim$37\%, suggesting these models retain a notion of distributional randomness over longer output sequences that is independent of overall scale. RealWorld tasks, by contrast, yield the highest individual scores: \textbf{Llama-3.2-1B achieves a remarkable 59.09\%} despite ranking poorly overall, a pattern we attribute to the narrower effective output range and near-uniform structure of many real-world distributions. \textbf{Nemotron-3 Super 120B drops sharply to 3.33\% on RealWorld} despite its overall dominance, revealing that strong distributional knowledge in structured domains does not transfer to real-world stochastic settings. Models heavily optimized for precise reasoning, notably Claude Sonnet 4.6 and Qwen3.5-35B-a3b, score near the bottom across all categories, consistent with the hypothesis that \textbf{deterministic fine-tuning suppresses the output diversity} \citep{west2025base} required for distributional matching. Finally, Mercury-2's diffusion-based generation does not appear to confer any natural advantage here, with the model collapsing to 0\% on Shuffling and underperforming across Text and RealWorld tasks.
We further break down performance by distribution 
(Appendix~\ref{app:dist_difficulty}), prompting format 
(Appendix~\ref{app:explicit_vs_implicit}), distribution modality 
(Appendix~\ref{app:multimodal_analysis}), and distributional spread 
(Appendix~\ref{app:concentrated_spred}). 


\begin{table}[t]
\renewcommand{\arraystretch}{1.5}
\caption{Per-category performance across Code, Text, RealWorld, and Shuffling tasks, 
reporting KS@100, JSD, and WDZ. JSD measures global distributional overlap while 
WDZ emphasizes tail behavior. \emph{Random Machine} is a Python pseudorandom number 
generator matching the ground-truth sampling procedure, serving as the theoretical 
performance ceiling. \textbf{Bold} values indicate the best score in each column. Full detailed results for all models are provided in Appendix~\ref{app:more_models}.}
\label{tab:main}
\resizebox{\linewidth}{!}{%
\begin{tabular}{l ccc ccc ccc ccc}
\toprule
\multirow{2}{*}{\textbf{Model}}
  & \multicolumn{3}{c}{\textbf{Code}}
  & \multicolumn{3}{c}{\textbf{Text}}
  & \multicolumn{3}{c}{\textbf{RealWorld}}
  & \multicolumn{3}{c}{\textbf{Shuffling}} \\
\cmidrule(lr){2-4}
\cmidrule(lr){5-7}
\cmidrule(lr){8-10}
\cmidrule(lr){11-13}
& \textbf{KS@100↑} & \textbf{JSD↓} & \textbf{WDZ↓}
& \textbf{KS@100↑} & \textbf{JSD↓} & \textbf{WDZ↓}
& \textbf{KS@100↑} & \textbf{JSD↓} & \textbf{WDZ↓}
& \textbf{KS@100↑} & \textbf{JSD↓} & \textbf{WDZ↓} \\
\midrule

\texttt{Nemotron-3-super-120B-a12b}   & \textbf{28.13}                            & 0.30         & 5.89         & \textbf{40.34}                            & 0.48         & 7.44         & 3.33                             & 0.16         & 35.36        & 21.05        & 0.15         & 10.13        \\
\texttt{GPT-5.4}                      & 25.63                            & 0.19         & 9.03         & 10.50                            & 0.30         & 10.58        & 6.67                             & 0.25         & 19        & 0            & 0.26         & 16.81        \\
\texttt{DeepSeek-v3.2} & \textbf{28.13} & \textbf{0.18} & 12.28 & 14.29 & 0.26 & 13.05 & 36.67 & 0.15 & 13.18 & \textbf{36.84} & 0.13 & 8.58 \\
\texttt{Llama-3.1-70B-instruct}       & 23.13                            & 0.21         & 15.01        & 9.66                             & 0.26         & 12.96        & 33.33                            & 0.15         & 12.23        & 21.05        & 0.15         & 10.21        \\
\texttt{Mercury-2}                    & 10.63                            & 0.31         & 12.60        & 7.14                             & 0.27         & \textbf{10.12}        & 13.33                            & 0.20         & 13.97        & 0            & 0.20         & 12.91        \\
\texttt{OLMo-3-7B-instruct}           & 7.50                             & 0.29         & 20.74        & 5.46                             & 0.46         & 34.93        & 3.33                             & 0.26         & 59.16        & \textbf{36.84}        & 0.13         & 7.64         \\
\texttt{Claude-Sonnet-4.6}            & 5.03                             & 0.34         & 13.27        & 5.04                             & \textbf{0.25 }        & 13.40        & 3.33                             & 0.31         & 19.61        & 0            & 0.22         & 18.17        \\
\texttt{Qwen-3.5-35b-a3b}             & 3.75                             & 0.52         & 19.35        & 2.94                             & 0.55         & 16.62        & 3.57                             & 0.43         & 25.46        & 0            & 0.31         & 18.11        \\
\texttt{Llama-3.2-1B-instruct}        & 10                            & 0.57         & 14.94        & 4.20                             & 0.52         & 29.82        & \textbf{59.09}                            & 0.20         & \textbf{11.35}        & \textbf{36.84}        & 0.12         & \textbf{6.26}         \\
\texttt{Qwen-3.5-2B}                  & 14.38                            & 0.35         & 11.30        & 12.61                            & 0.29         & 10.87        & 31.03                            & 0.17         & 12.32        & \textbf{36.84}        & \textbf{0.11}         & 3.18         \\
\texttt{Qwen-3-32B}                   & 11.88                            & 0.24         & 14.30        & 5.46                             & 0.27         & 12.81        & 16.67                            & 0.23         & 17.33        & 0         & 0.19         & 13.21        \\
\texttt{Ministral-3-3B-instruct-2512} & 21.25                            & \textbf{0.18}         & \textbf{1.61}         & 17.23                            & 0.37         & 12.11        & 6.67                             & \textbf{0.13}         & 36.14        & 5.26         & 0.15         & 9.67         \\ \hline
\textbf{Random Machine}                                     & 100                              & 0.02         & -0.08        & 100                              & 0.02         & -0.08        & 100                              & 0.02         & -0.08        & 100          & 0.02         & -0.09        \\ \bottomrule
\end{tabular}
}
\vspace{-12pt}
\end{table}

\begin{table}[ht]
\centering
\caption{KS@N (\%) on the full UnpredictaBench dataset for all evaluated models at seven sample sizes $N \in \{1, 2, 5, 10, 20, 50, 100\}$. KS@N measures the fraction of problems for which the model's $N$ samples are not rejected under the KS test at threshold $p < 0.0001$. All models achieve perfect KS@1 by construction, as a single sample is almost never rejected. Performance generally degrades monotonically with $N$, with the steepest drops occurring between KS@20 and KS@100. \textbf{Bold} values indicate the best score in each column.}
\label{tab:acc_ks}
\resizebox{\textwidth}{!}{%
\begin{tabular}{l ccccccc}
\toprule
\textbf{Model}
  & \textbf{KS@1} & \textbf{KS@2} & \textbf{KS@5}
  & \textbf{KS@10} & \textbf{KS@20} & \textbf{KS@50} & \textbf{KS@100} \\
\midrule

\texttt{Nemotron-3-super-120B-a12b}      & \textbf{100} & 94.87 & 96.88 & 88.62 & \textbf{77.15} & \textbf{51.86} & \textbf{32.64} \\
\texttt{GPT-4o}                          & \textbf{100} & 97.10 & 96.21 & \textbf{89.03} & 74.21 & 42.46 & 23.90 \\
\texttt{DeepSeek-v3.2}                 & \textbf{100} & 93.75 & 94.64 & 85.27 & 68.47 & 37.62 & 21.73 \\
\texttt{Nemotron-3-nano-30B-a3b}         & \textbf{100} & 92.19 & 95.54 & 86.61 & 73.41 & 40.31 & 20.83 \\
\texttt{GPT-5.4}                         & \textbf{100} & \textbf{98.44} & \textbf{98.44} & 88.10 & 63.01 & 30.80 & 15.18 \\
\texttt{Llama-3.1-70B-instruct}      & \textbf{100} & 94.42 & 93.75 & 85.71 & 65.99 & 30.42 & 16.56 \\
\texttt{Qwen3.5-2B}                        & \textbf{100} & 87.04 & 89.49 & 79.66 & 58.63 & 34.66 & 17.67 \\
\texttt{GPT-4o-mini}                     & \textbf{100} & 93.75 & 90.40 & 78.98 & 54.10 & 20.78 & 9.60  \\
\texttt{Mercury-2}                    & \textbf{100} & 91.52 & 92.86 & 79.22 & 56.57 & 20.78 & 8.48  \\
\texttt{Qwen3-32B}                         & \textbf{100} & 95.31 & 94.20 & 79.90 & 53.70 & 18.77 & 8.26  \\
\texttt{Grok-4.1-fast}                     & \textbf{100} & 93.53 & 89.51 & 69.40 & 35.14 & 9.40  & 5.58  \\
\texttt{Claude-sonnet-4.6}            & \textbf{100} & 97.76 & 94.86 & 76.66 & 33.82 & 8.73  & 4.70  \\
\texttt{Mistral-large-2512}           & \textbf{100} & 94.42 & 91.07 & 73.84 & 43.82 & 10.50 & 4.69  \\
\texttt{Llama-3.1-8B-instruct}       & \textbf{100} & 80.80 & 81.47 & 68.52 & 44.15 & 17.27 & 6.51  \\
\texttt{Qwen3.5-35B-a3b}                   & \textbf{100} & 89.24 & 79.62 & 49.19 & 11.89 & 4.03  & 3.14  \\
\texttt{Qwen3.5-397B-a17b}                 & \textbf{100} & 94.87 & 87.28 & 54.57 & 16.79 & 4.70  & 3.36  \\
\texttt{Llama-3.2-1B-Instruct}       & \textbf{100} & 52.44 & 65.83 & 39.77 & 23.83 & 12.13 & 8.06  \\
\texttt{Phi-3.5-mini-instruct}        & \textbf{100} & 89.51 & 81.92 & 63.82 & 30.47 & 8.28  & 2.90  \\
\texttt{Olmo-3-7B-Instruct}             & \textbf{100} & 89.29 & 84.38 & 70.54 & 46.17 & 18.63 & 7.45  \\
\texttt{Ministral-3-3B-Instruct-2512} & \textbf{100} & 86.83 & 90.40 & 77.01 & 54.21 & 29.29 & 15.41 \\

\bottomrule
\end{tabular}%
}
\end{table}

\subsection{The Effect of Instruction Tuning}

Table~\ref{tab:base_vs_instruct} compares base and instruction-tuned variants of three models on the Text+Code subset of UnpredictaBench, excluding the Shuffling and RealWorld subsets. The results show that \textbf{instruction tuning provides only slight benefit for distributional understanding and, in most cases, actively reduces output diversity}. While $KS@100$ improves modestly across all three models, the gains are small, suggesting that the base model's knowledge of the target distribution is largely preserved but not meaningfully enhanced by instruction tuning. Notably, JSD and WDZ reveal a more nuanced trend: instruction tuning sometimes worsens these metrics because base models, while more diverse, occasionally generate out-of-support values, increasing distributional distance. Instruction tuning reduces such errors, but often at the cost of diversity.

\begin{table}[]
\centering
\caption{Base vs.\ instruction-tuned model variants across KS@50, KS@100, JSD, and 
WDZ, evaluated on \name{} excluding the Shuffling and RealWorld 
subsets. $\Delta$ denotes the change from base to instruct. \textcolor{green!60!black}{Green} indicates the base model outperforms the instruction-tuned.}
\label{tab:base_vs_instruct}
\resizebox{\textwidth}{!}{%
\begin{tabular}{l cc cc cc cc}
\toprule
\multirow{2}{*}{\textbf{Model}}
  & \multicolumn{2}{c}{\textbf{KS@50↑}}
  & \multicolumn{2}{c}{\textbf{KS@100↑}}
  & \multicolumn{2}{c}{\textbf{JSD↓}}
  & \multicolumn{2}{c}{\textbf{WDZ↓}} \\
\cmidrule(lr){2-3} \cmidrule(lr){4-5} \cmidrule(lr){6-7} \cmidrule(lr){8-9}
  & \textbf{Score} & \textbf{$\Delta$}
  & \textbf{Score} & \textbf{$\Delta$}
  & \textbf{Score} & \textbf{$\Delta$}
  & \textbf{Score} & \textbf{$\Delta$} \\
\midrule

\texttt{Qwen3.5-2B-base}
  & 0.3543 & \posvalue{0.0402}
  & 0.1658 & \posvalue{0.0327}
  & 0.2965 & \negvaluegood{$-$0.0208}
  & 11.4678 & \posvaluebad{0.3844} \\

\texttt{Llama-3.2-1B}
  & 0.2312 & \posvalue{0.1759}
  & 0.1131 & \posvalue{0.0854}
  & 0.4478 & \negvaluegood{$-$0.0965}
  & 6.0773  & \negvaluegood{$-$16.3010} \\

\texttt{Ministral-3B-base}
  & 0.3241 & \posvalue{0.0151}
  & 0.1784 & \posvalue{0.0126}
  & 0.3814 & \posvaluebad{0.1096}
  & 6.8250  & \negvaluegood{$-$0.0330} \\

\bottomrule
\end{tabular}%
}
\vspace{-5pt}
\end{table}

\subsection{The Effect of Reasoning}
\label{sec:exp:reasoning}

Table~\ref{tab:reasoning_comparison} compares KS@N when evaluated on the 
model's \textit{Final Output} versus numbers extracted \textit{From Reasoning} tokens on the Text+Code subset of UnpredictaBench, excluding the Shuffling and RealWorld categories. Overall, 
reasoning improves final output performance across all four models, consistent with the 
hypothesis that the core challenge is not merely output diversity but also 
\textit{understanding the distribution described in the prompt}. However, \textbf{the 
benefit is model-specific and the two number sources tell very different stories}. For 
Nemotron-3 Super 120B and DeepSeek V3.2, extracting numbers from reasoning tokens 
causes a sharp performance drop ($-$33.17 and $-$15.33 at KS@20 respectively), 
suggesting these models repeatedly revisit the same candidate values during deliberation 
rather than broadly exploring the support. \textbf{The reasoning process explores less 
than it appears to.} By contrast, Qwen3-32B benefits from reasoning in both conditions, 
with its reasoning tokens yielding a gain of $+$9.30 at KS@50. Qwen3.5-35B-a3b 
presents the most striking case: its final output yields essentially zero improvement 
over baseline at all sample sizes, because the model defaults to repeating a single 
number in its final answer. Yet its reasoning tokens reveal a substantially broader set 
of candidates it considers but never reports, yielding a large gain when extracted 
directly ($+$35.18 at KS@20). \textbf{This model knows more than it says.}

\begin{table}[]
\centering
\caption{KS@N for \textit{Final Output} vs.\ numbers extracted \textit{From 
Reasoning} tokens, evaluated on \name{} excluding the Shuffling and 
RealWorld subsets. $\Delta$ denotes the change relative to the no-reasoning baseline. 
Shaded rows correspond to the \textit{From Reasoning} condition.}
\label{tab:reasoning_comparison}
\resizebox{\textwidth}{!}{%
\begin{tabular}{ll cc cc cc}
\toprule
\multirow{2}{*}{\textbf{Model}} & \multirow{2}{*}{\textbf{Source}} 
  & \multicolumn{2}{c}{\textbf{KS@20}} 
  & \multicolumn{2}{c}{\textbf{KS@50}} 
  & \multicolumn{2}{c}{\textbf{KS@100}} \\
\cmidrule(lr){3-4} \cmidrule(lr){5-6} \cmidrule(lr){7-8}
& & \textbf{Score} & \textbf{$\Delta$} 
  & \textbf{Score} & \textbf{$\Delta$} 
  & \textbf{Score} & \textbf{$\Delta$} \\
\midrule

\multirow{2}{*}{\texttt{Nemotron-3-super-120B-a12b}}
  & Final Output
    & 83.67 & \posvalue{3.27}
    & 59.55 & \posvalue{3.27}
    & 36.43 & \posvalue{1.01} \\
  & \cellcolor{gray!20}From Reasoning
    & \cellcolor{gray!20}49.50 & \cellcolor{gray!20}\negvalue{$-$33.17}
    & \cellcolor{gray!20}27.89 & \cellcolor{gray!20}\negvalue{$-$28.39}
    & \cellcolor{gray!20}13.32 & \cellcolor{gray!20}\negvalue{$-$22.11} \\
\midrule

\multirow{2}{*}{\texttt{DeepSeek-v3.2}}
  & Final Output
    & 75.13 & \posvalue{5.03}
    & 41.21 & \posvalue{5.03}
    & 23.37 & \posvalue{3.52} \\
  & \cellcolor{gray!20}From Reasoning
    & \cellcolor{gray!20}52.51 & \cellcolor{gray!20}\negvalue{$-$15.33}
    & \cellcolor{gray!20}23.37 & \cellcolor{gray!20}\negvalue{$-$12.81}
    & \cellcolor{gray!20}9.05  & \cellcolor{gray!20}\negvalue{$-$10.80} \\
\midrule

\multirow{2}{*}{\texttt{Qwen3-32B}}
  & Final Output
    & 60.21 & \posvalue{4.92}
    & 23.51 & \posvalue{4.92}
    & 10.34 & \posvalue{2.30} \\
  & \cellcolor{gray!20}From Reasoning
    & \cellcolor{gray!20}57.29 & \cellcolor{gray!20}\posvalue{2.01}
    & \cellcolor{gray!20}27.89 & \cellcolor{gray!20}\posvalue{9.30}
    & \cellcolor{gray!20}9.05  & \cellcolor{gray!20}\posvalue{1.01} \\
\midrule

\multirow{2}{*}{\texttt{Qwen3.5-35B-a3b}}
  & Final Output
    & 17.09 & \neuvalue{0.00}
    & 4.27  & \neuvalue{0.00}
    & 3.27  & \neuvalue{0.00} \\
  & \cellcolor{gray!20}From Reasoning
    & \cellcolor{gray!20}47.49 & \cellcolor{gray!20}\posvalue{35.18}
    & \cellcolor{gray!20}17.84 & \cellcolor{gray!20}\posvalue{13.57}
    & \cellcolor{gray!20}5.78  & \cellcolor{gray!20}\posvalue{2.51} \\

\bottomrule
\end{tabular}%
}
\vspace{-5pt}
\end{table}

\subsection{Qualitative Analysis}
\label{sec:qualitative}

Figure~\hyperref[fig:teaser]{\ref*{fig:teaser}(a)} illustrates two representative failure modes observed across the benchmark. On the Skellam distribution, Nemotron-3-Super-120B covers part of the target support, including both negative and positive values, but assigns probability mass incorrectly and collapses much of its density onto a small number of bins. On the Poisson task, OLMo-3-7B produces a right-skewed set of samples concentrated at small values, while the true distribution peaks at substantially larger counts and maintains support beyond 20. Together, these examples show that models often fail not only by collapsing to an overly narrow output range, but also by producing samples that occupy the rough numerical range of the target while misrepresenting its probability structure.

\begin{figure}[t]
    \centering
    \caption{Llama-3.2-1B-base (top) and -instruct (bottom) on a Beta distribution as text (left) and a Poisson-Binomial distribution as code (right). All values are max-scaled for visibility.}
    \includegraphics[width=\linewidth]{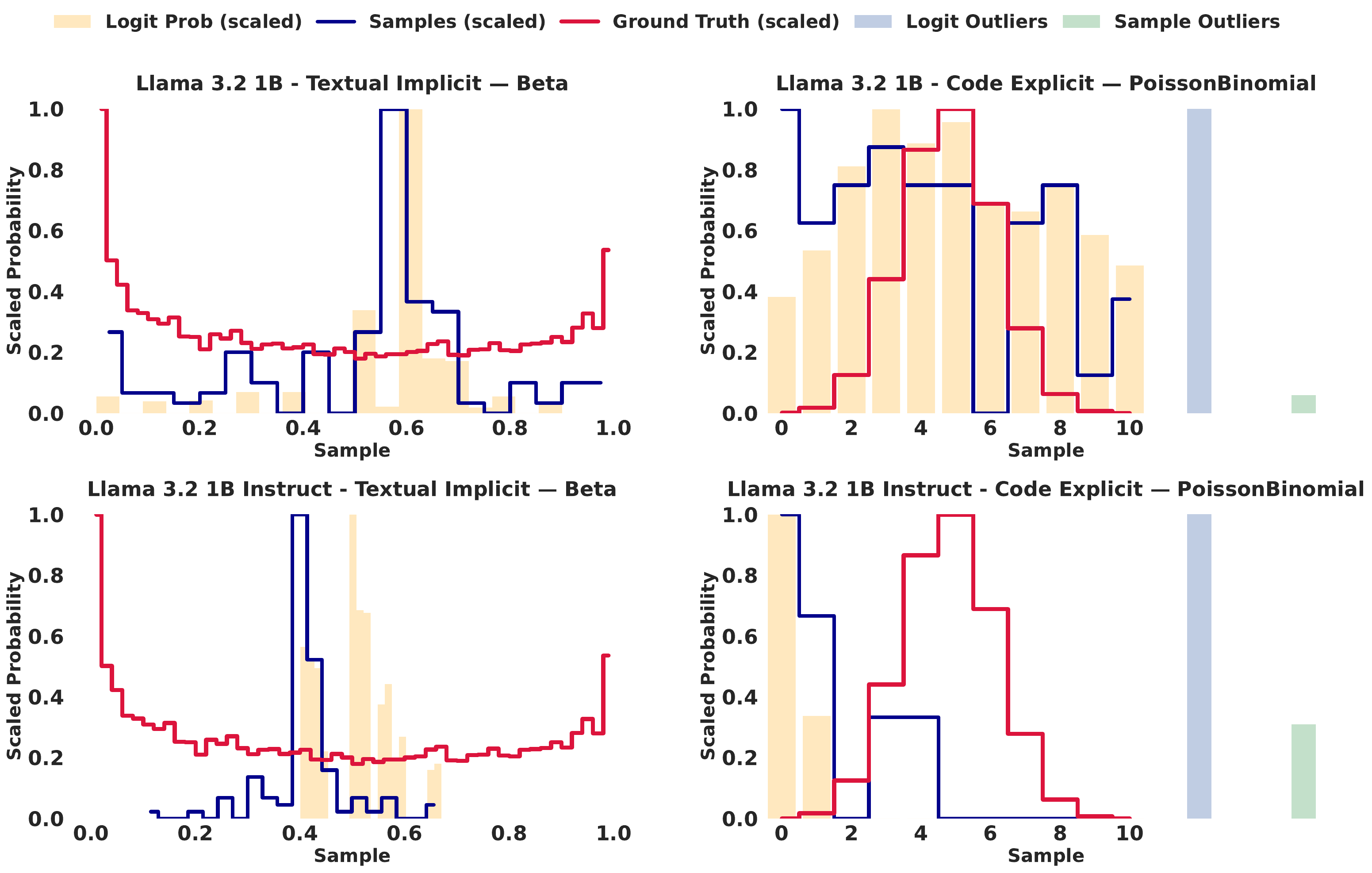}
    \vspace{-10pt}
    \label{fig:logit_analysis}
\end{figure}

Figure~\ref{fig:logit_analysis} deepens this picture by examining Llama 3.2 1B (base 
and instruct) on a Beta distribution and a Poisson-Binomial task, overlaying the 
ground truth, model samples, and logit probability mass $P(y) \propto 
\prod_t P(t_t \mid t_{<t}, x)$, all max-scaled for visibility. 
\textbf{Logit and sample distributions are consistently closely aligned}; and this 
is not trivially expected. Since each call involves independent stochastic decoding, 
one might expect logit distributions to vary across calls, producing broad sample 
diversity. Instead, \textbf{the model's internal beliefs are stable across calls and 
the failure is already visible in the logits}: the diversity problem is not a decoding 
artifact but a reflection of what the model fundamentally believes is plausible. On the discrete task (Poisson Binomial), the 
base model covers the support substantially more broadly than the instruct variant, 
which collapses toward lower values; illustrating 
\textbf{instruction tuning suppresses output diversity} by penalizing unusual outputs 
during RLHF-style~\cite{lambert2026reinforcementlearninghumanfeedback} training. Moreover, we observe more outliers in the Poisson Binomial task, likely due to the distribution’s complexity. On the continuous distribution Beta, both variants fail to capture the U-shaped ground truth. Additional qualitative analysis is provided in Appendix~\ref{app:additional_qualitative}. We further analyze two behaviors in the appendix. Appendix~\ref{app:retry} reports instruction following, measuring how often a model fails to return a structurally valid output when prompted (note that this differs from producing values that fall within the target distribution.). Appendix~\ref{app:uniqueness} reports output diversity, measuring how many of a model's 100 runs yield a previously unseen number.

\subsection{Alignment with Novelty and Creativity Benchmarks}
\label{sec:cross_dataset}

We compare \textsc{UnpredictaBench} against NoveltyBench and CREATE to examine whether distributional fidelity relates to broader notions of creativity and novelty. NoveltyBench reports Distinct\textsubscript{10}, measuring lexical diversity, and Utility\textsubscript{10}, measuring the combined usability and diversity of generated outputs. CREATE reports utility under two temperature settings, $p{=}0.7$ and $p{=}0.9$. Unlike both benchmarks, which rely on LLM-as-a-judge evaluation, \textsc{UnpredictaBench} provides a statistically grounded metric that directly measures distributional fidelity without requiring a judge model.

\begin{figure}[t]
    \centering
    \caption{Pearson correlation between \textsc{UnpredictaBench} KS@100 and metrics from NoveltyBench and CREATE across seven models. Each scatter plot compares one external benchmark metric against KS@100, with a fitted regression line.}
    \label{fig:correlation_cross_dataset}
    \includegraphics[width=\linewidth]{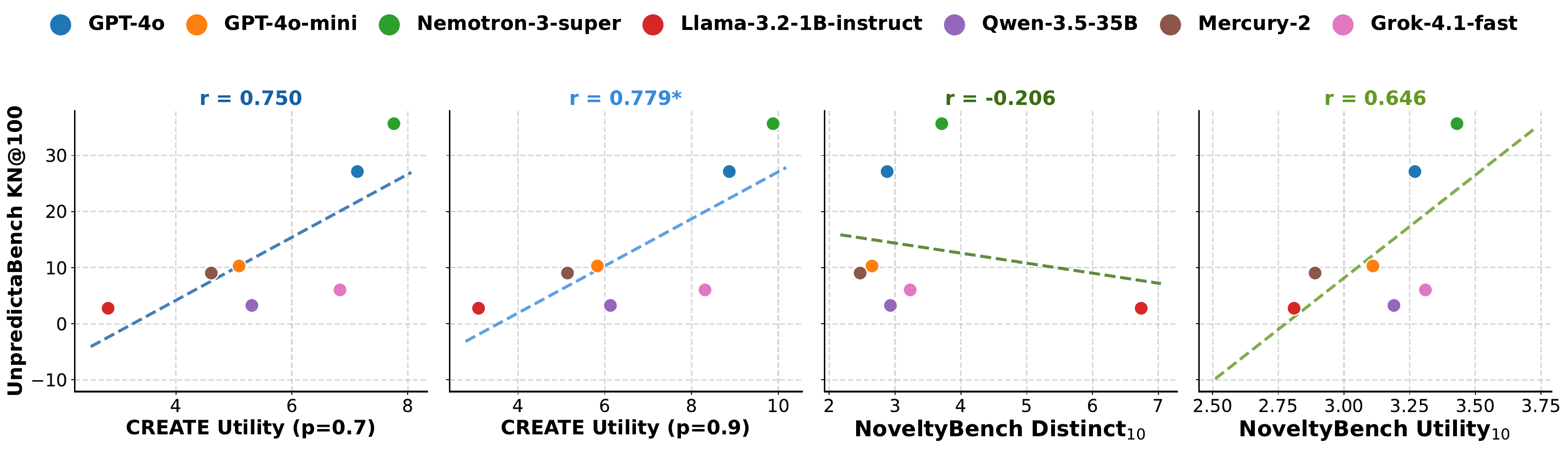}
    \vspace{-15pt}
\end{figure}

As shown in Figure~\ref{fig:correlation_cross_dataset}, KS@100 correlates positively with utility metrics from both benchmarks, including CREATE Utility at $p{=}0.7$ ($r=0.75$) and $p{=}0.9$ ($r=0.78^*$), as well as NoveltyBench Utility\textsubscript{10} ($r=0.65$). This suggests that distributional fidelity captures a meaningful aspect of creative generation. In contrast, NoveltyBench Distinct\textsubscript{10} correlates negatively with KS@100 ($r=-0.21$), consistent with our finding that raw diversity without distributional understanding is insufficient. As shown in Table~\ref{tab:cross_dataset}, Nemotron-3 Super 120B leads across all benchmarks, while Llama-3.2-1B-instruct achieves the highest Distinct\textsubscript{10} score but ranks near the bottom on utility and KS@100, illustrating that lexical diversity and distributional fidelity are distinct properties. \textbf{Mercury-2 is a notable outlier}: its diffusion-based architecture yields diverse numerical outputs on our structured stochastic tasks, but struggles with the open-ended linguistic diversity required by creativity benchmarks.

\begin{table}[t]
    \centering
    \caption{Cross-benchmark comparison of model performance on NoveltyBench, CREATE, and \textsc{UnpredictaBench} (ours). NoveltyBench reports Distinct\textsubscript{10} and Utility\textsubscript{10}; CREATE reports utility at two temperature settings; \textsc{UnpredictaBench} reports KS@100.}
    \label{tab:cross_dataset}
    \resizebox{\linewidth}{!}{%
    \begin{tabular}{l cc cc c}
    \toprule
    \multirow{2}{*}{\textbf{Model}}
      & \multicolumn{2}{c}{\textbf{NoveltyBench}}
      & \multicolumn{2}{c}{\textbf{CREATE}}
      & \textbf{\textsc{UnpredictaBench}} \\
    \cmidrule(lr){2-3} \cmidrule(lr){4-5} \cmidrule(lr){6-6}
      & \textbf{Distinct\textsubscript{10}↑}
      & \textbf{Utility\textsubscript{10}↑}
      & \textbf{Utility ($p{=}0.7$)↑}
      & \textbf{Utility ($p{=}0.9$)↑}
      & \textbf{KS@100↑} \\
    \midrule
    \texttt{GPT-4o}
      & 2.88 & 3.27
      & 7.13 & 8.87
      & 27.13 \\
    \texttt{GPT-4o-mini}
      & 2.65 & 3.11
      & 5.09 & 5.83
      & 10.30 \\
    \texttt{Nemotron-3-super-120B-a12b}
      & 3.71 & 3.43
      & 7.76 & 9.88
      & 35.67 \\
    \texttt{Llama-3.2-1B-instruct}
      & 6.74 & 2.81
      & 2.83 & 3.09
      & 2.76 \\
    \texttt{Qwen-3.5-35B-a3b}
      & 2.93 & 3.19
      & 5.31 & 6.13
      & 3.26 \\
    \texttt{Mercury-2}
      & 2.47 & 2.89
      & 4.61 & 5.14
      & 9.04 \\
    \texttt{Grok-4.1-fast}
      & 3.23 & 3.31
      & 6.83 & 8.31
      & 6.03 \\
    \bottomrule
    \end{tabular}%
    }
\end{table}
\section{Ablations}
\label{sec:ablations}

\subsection{The Effect of Temperature}
\label{sec:ablations:temp}

Table~\ref{tab:ablation_temp} reports performance across five temperature settings for three models. The results reveal a consistent and intuitive pattern: \textbf{higher temperatures improve KS@100 across all models}, as increased sampling diversity brings model outputs closer to the ground-truth distribution. For Nemotron-3 Super 120B, KS@100 peaks around $T{=}1.2$ (39.57\% average) and remains strong at $T{=}1.5$, while dropping sharply at $T{=}0.1$ (5.23\%), confirming that near-greedy decoding is particularly harmful for stochastic tasks. Ministral-3 3B follows the same trend, though interestingly its best performance occurs at $T{=}1.0$ on Code and $T{=}1.2$ on Text, suggesting a task-dependent optimal temperature. \textbf{OLMo-3 7B is a notable exception}: its WDZ remains persistently high across all temperatures and even increases slightly with temperature, indicating that higher diversity comes at the cost of greater tail deviation. This suggests that for weaker models, raising temperature amplifies out-of-support outputs rather than improving distributional coverage, echoing the base-versus-instruct trade-off discussed in the previous section. Taken together, these results suggest that \textbf{temperature is an important but model-dependent lever}: strong models benefit substantially from higher temperatures, while weaker 
models may require more targeted interventions.

\begin{table}[]
    \renewcommand{\arraystretch}{1.5}
    \caption{Effect of sampling temperature on model performance across Code and Text task categories, along with their average. For each setting we report KS@100 (higher is better), JSD (lower is better), and WDZ (closer to zero is better). All models are evaluated at five temperatures: $T \in \{0.1, 0.7, 1.0, 1.2, 1.5\}$. Higher temperatures generally improve KS@100 by increasing diversity, though weaker models may show larger tail deviations in WDZ.}
    \label{tab:ablation_temp}
    \resizebox{\linewidth}{!}{%
    \begin{tabular}{l c ccc ccc ccc}
\toprule
\multirow{2}{*}{\textbf{Model}}
  & \multirow{2}{*}{\textbf{Temp}}
  & \multicolumn{3}{c}{\textbf{Code}}
  & \multicolumn{3}{c}{\textbf{Text}}
  & \multicolumn{3}{c}{\textbf{Average}} \\
\cmidrule(lr){3-5}
\cmidrule(lr){6-8}
\cmidrule(lr){9-11}
& 
& \textbf{KS@100↑} & \textbf{JSD↓} & \textbf{WDZ↓}
& \textbf{KS@100↑} & \textbf{JSD↓} & \textbf{WDZ↓}
& \textbf{KS@100↑} & \textbf{JSD↓} & \textbf{WDZ↓} \\
\midrule
\texttt{Nemotron-3-super-120b-a12b}      & 1.5                            & 44.96                            & 0.54                     & 7.26                       & 33.75                            & 0.36                     & 5.74                       & 39.35                            & 0.45                     & 6.50                      \\
\texttt{Nemotron-3-super-120b-a12b}      & 1.2                            & 46.64                            & 0.51                     & 7.34                       & 32.50                            & 0.33                     & 5.82                       & 39.57                            & 0.42                     & 6.58                      \\
\texttt{Nemotron-3-super-120b-a12b}      & 1.0                            & 40.34                            & 0.48                     & 7.44                       & 28.13                            & 0.30                     & 5.89                       & 34.23                            & 0.39                     & 6.67                      \\
\texttt{Nemotron-3-super-120b-a12b}      & 0.7                            & 21.85                            & 0.44                     & 7.60                       & 18.13                            & 0.27                     & 6.05                       & 19.99                            & 0.36                     & 6.83                      \\
\texttt{Nemotron-3-super-120b-a12b}      & 0.1                            & 5.46                             & 0.40                     & 7.82                       & 5.00                             & 0.23                     & 6.24                       & 5.23                             & 0.32                     & 7.03                      \\ \hline
\texttt{Ministral-3-3B-instruct-2512} & 1.5                            & 15.55                            & 0.43                     & 11.84                      & 19.38                            & 0.22                     & 1.52                       & 17.46                            & 0.32                     & 6.68                      \\
\texttt{Ministral-3-3B-instruct-2512} & 1.2                            & 15.55                            & 0.40                     & 11.98                      & 23.13                            & 0.20                     & 1.57                       & 19.34                            & 0.30                     & 6.77                      \\
\texttt{Ministral-3-3B-instruct-2512} & 1.0                            & 17.23                            & 0.37                     & 12.11                      & 21.25                            & 0.18                     & 1.61                       & 19.24                            & 0.27                     & 6.86                      \\
\texttt{Ministral-3-3B-instruct-2512} & 0.7                            & 10.08                            & 0.33                     & 12.30                      & 12.50                            & 0.15                     & 1.69                       & 11.29                            & 0.24                     & 6.99                      \\
\texttt{Ministral-3-3B-instruct-2512} & 0.1                            & 1.26                             & 0.29                     & 12.56                      & 5.00                             & 0.12                     & 1.81                       & 3.13                             & 0.20                     & 7.19                      \\
\midrule

\texttt{OLMo-3-7B-instruct}
  & 1.5 & 9.24 & 0.52 & 34.58 & 9.38 & 0.34 & 20.38 & 9.31 & 0.43 & 27.48 \\
\texttt{OLMo-3-7B-instruct}
  & 1.2 & 6.30 & 0.49 & 34.74 & 9.38 & 0.31 & 20.56 & 7.84 & 0.40 & 27.65 \\
\texttt{OLMo-3-7B-instruct}
  & 1.0 & 5.46 & 0.46 & 34.93 & 7.50 & 0.29 & 20.74 & 6.48 & 0.37 & 27.84 \\
\texttt{OLMo-3-7B-instruct}
  & 0.7 & 3.36 & 0.42 & 35.21 & 5.66 & 0.26 & 20.99 & 4.51 & 0.34 & 28.10 \\
\texttt{OLMo-3-7B-instruct}
  & 0.1 & 2.10 & 0.37 & 35.59 & 2.52 & 0.22 & 21.25 & 2.31 & 0.30 & 28.42 \\
  
\bottomrule
\end{tabular}
    }
\end{table}

\subsection{Effect of Sampling Budget}

Table~\ref{tab:scaling_sampling} examines what happens when models are given larger generation budgets of 500 and 1000 samples, evaluated at increasing subset sizes. Two complementary trends emerge. First, \textbf{generating more samples consistently improves short-horizon KS@100}: KS@100 increases for all models as the generation budget grows from 100 to 1000, suggesting that with more attempts, models are more likely to produce outputs that locally resemble the target distribution. Second, and more revealingly, \textbf{evaluating over the full generated set exposes deeper distributional failures}: KS@500 and KS@1000 are consistently lower than KS@100 within the same generation budget, meaning that while models can appear well-calibrated over a small sample, their biases become statistically detectable under stricter evaluation. This confirms that \textbf{our choice of $N{=}100$ for the main benchmark is conservative}: models that pass at KS@100 may still fail under more demanding scrutiny, and the true ceiling of current models is lower than the headline numbers suggest. Ministral-3 3B is the strongest model across all settings, maintaining the highest KS@100 at every budget and evaluation threshold, while Llama-3.2-1B and Phi-3.5 Mini remain near the bottom regardless of how many samples are drawn, indicating that \textbf{scaling the sampling budget cannot compensate for a fundamental lack of distributional understanding}.

\begin{table}
\centering
\caption{Effect of increasing generation budget on distributional fidelity. For each generation budget (100, 500, and 1000 samples), we report KS-based KS@100 evaluated at different subset sizes. KS@100 within a larger budget measures short-horizon fidelity, while KS@500 and KS@1000 apply stricter statistical scrutiny over the full generated set. Larger budgets improve short-horizon KS@100 but consistently reveal deeper distributional biases when evaluated at scale, demonstrating that models cannot fully escape their distributional limitations by generating more samples.}
\label{tab:scaling_sampling}
\setlength{\tabcolsep}{6pt}
\renewcommand{\arraystretch}{1.2}
\resizebox{\linewidth}{!}{%
\begin{tabular}{l c @{\hspace{1.0em}} cc @{\hspace{1.0em}} ccc}
\toprule
\multirow{2}{*}{\textbf{Model}} 
& \multicolumn{1}{c}{\textbf{100 Samples}} 
& \multicolumn{2}{c}{\textbf{500 Samples}} 
& \multicolumn{3}{c}{\textbf{1000 Samples}} \\

\cmidrule(lr){2-2}
\cmidrule(lr){3-4}
\cmidrule(lr){5-7}

& \textbf{KS@100} 
& \textbf{KS@100} 
& \textbf{KS@500} 
& \textbf{KS@100} 
& \textbf{KS@500} 
& \textbf{KS@1000} \\

\midrule

\texttt{Llama-3.2-1B-instruct}
& 2.76 
& 3.61 
& 2.14 
& 4.17 
& 2.98 
& 1.79 \\

\texttt{Phi-3.5-mini-instruct}
& 2.51 
& 3.38 
& 1.97 
& 3.92 
& 2.76 
& 1.68 \\

\texttt{OLMo-3-7B-instruct}
& 6.28 
& 7.54 
& 5.49 
& 8.31 
& 6.67 
& 4.88 \\

\texttt{Ministral-3-3B-instruct-2512}
& \textbf{16.58} 
& \textbf{18.12} 
& \textbf{15.24} 
& \textbf{19.41} 
& \textbf{17.16} 
& \textbf{14.68} \\

\bottomrule
\end{tabular}
}
\end{table}

\subsection{The Effect of Asking for a List of Samples Instead of One}
\label{sec:ablations:list}

Table~\ref{tab:output_count_comparison} compares the standard single-output protocol 
against prompting models to generate lists of 10 or 35 values per call, merging 
repeated calls until 100 numbers are accumulated (truncating to the first 100 if more are produced). We capped list size at 35 because models consistently fail to follow instructions beyond this threshold, skipping numbers or truncating their output prematurely. The results show that \textbf{asking for lists generally improves KS@100}, consistent with the intuition that generating multiple values in a single forward pass encourages the model to diversify across the support rather than anchoring on a single point. However, \textbf{the benefit is strongly model-dependent and does not hold uniformly across evaluation thresholds}. For Nemotron-3 Super 120B and Ministral-3 3B, requesting 10 outputs yields meaningful gains at KS@100 ($+$17.12 and $+$14.32 respectively) but slightly hurts short-horizon performance at KS@20, suggesting that list generation improves global coverage at the cost of local coherence. Increasing to 35 outputs partially reverses these gains, indicating a sweet spot around 10 values per call for these models. OLMo-3 7B benefits consistently across both list sizes and all evaluation thresholds, suggesting it handles list generation well regardless of list length. \textbf{Llama-3.2-1B is the exception}: list prompting hurts performance at nearly every threshold and list size, with 35 outputs causing a sharp drop ($-$3.81 at KS@100), likely because the model struggles to maintain distributional diversity over longer lists and instead repeats values or drifts out of support. Taken together, these results suggest that \textbf{list prompting is a simple but model-sensitive intervention} that can meaningfully improve distributional fidelity for capable models without any additional training.

\begin{table}[ht]
\centering
\caption{Comparison of single-output and list-output prompting strategies at list sizes of 10 and 35, evaluated at KS@20, KS@50, and KS@100. To reach 100 total samples, model calls are repeated and their outputs merged; if a call produces more than the requested count, only the first 100 values are used. List size is capped at 35 as models reliably fail to follow instructions for larger lists, skipping or truncating their output. $\Delta$ denotes the change relative to the single-output baseline. Shaded rows correspond to the 35-output condition.}
\label{tab:output_count_comparison}
\resizebox{\textwidth}{!}{%
\begin{tabular}{ll cc cc cc}
\toprule
\multirow{2}{*}{\textbf{Model}} & \multirow{2}{*}{\textbf{Source}}
  & \multicolumn{2}{c}{\textbf{KS@20}}
  & \multicolumn{2}{c}{\textbf{KS@50}}
  & \multicolumn{2}{c}{\textbf{KS@100}} \\
\cmidrule(lr){3-4} \cmidrule(lr){5-6} \cmidrule(lr){7-8}
& & \textbf{Score} & \textbf{$\Delta$}
  & \textbf{Score} & \textbf{$\Delta$}
  & \textbf{Score} & \textbf{$\Delta$} \\
\midrule

\multirow{3}{*}{\texttt{Nemotron-3-super-120B-a12b}}
  & 10 outputs
    & 80.15 & \negvalue{$-$2.22}
    & 68.84 & \posvalue{12.01}
    & 52.01 & \posvalue{17.12} \\
  & \cellcolor{gray!20}35 outputs
    & \cellcolor{gray!20}66.83 & \cellcolor{gray!20}\negvalue{$-$15.54}
    & \cellcolor{gray!20}55.53 & \cellcolor{gray!20}\negvalue{$-$1.31}
    & \cellcolor{gray!20}41.21 & \cellcolor{gray!20}\posvalue{6.31} \\
\midrule

\multirow{3}{*}{\texttt{Ministral-3-3B-instruct-2512}}
  & 10 outputs
    & 59.55 & \negvalue{$-$0.75}
    & 44.97 & \posvalue{13.57}
    & 30.65 & \posvalue{14.32} \\
  & \cellcolor{gray!20}35 outputs
    & \cellcolor{gray!20}54.27 & \cellcolor{gray!20}\negvalue{$-$6.03}
    & \cellcolor{gray!20}38.19 & \cellcolor{gray!20}\posvalue{6.78}
    & \cellcolor{gray!20}26.88 & \cellcolor{gray!20}\posvalue{10.55} \\
\midrule

\multirow{3}{*}{\texttt{OLMo-3-7B-instruct}}
  & 10 outputs
    & 62.31 & \posvalue{14.07}
    & 39.20 & \posvalue{21.36}
    & 23.87 & \posvalue{17.59} \\
  & \cellcolor{gray!20}35 outputs
    & \cellcolor{gray!20}55.03 & \cellcolor{gray!20}\posvalue{6.78}
    & \cellcolor{gray!20}35.68 & \cellcolor{gray!20}\posvalue{17.84}
    & \cellcolor{gray!20}24.37 & \cellcolor{gray!20}\posvalue{18.09} \\
\midrule

\multirow{3}{*}{\texttt{Llama-3.2-1B-instruct}}
  & 10 outputs
    & 18.09 & \negvalue{$-$2.16}
    & 10.05 & \posvalue{3.09}
    & 4.02 & \negvalue{$-$1.04} \\
  & \cellcolor{gray!20}35 outputs
    & \cellcolor{gray!20}11.06 & \cellcolor{gray!20}\negvalue{$-$9.20}
    & \cellcolor{gray!20}4.77 & \cellcolor{gray!20}\negvalue{$-$2.19}
    & \cellcolor{gray!20}1.26 & \cellcolor{gray!20}\negvalue{$-$3.81} \\
\bottomrule
\end{tabular}%
}
\end{table}

\section{Conclusion}
\label{sec:conclusion}

We introduced \textsc{UnpredictaBench}, a benchmark for evaluating the ability of LLMs to generate samples consistent with true underlying statistical distributions. Across 448 test instances spanning 40 distributions and four task categories, we find that no current model comes close to solving the benchmark, with even the strongest model achieving only 32.64\% at KS@100. Models fail in two distinct ways: lacking a meaningful internal representation of the target distribution, or understanding its rough shape but collapsing to a narrow set of outputs. Instruction tuning exacerbates the latter, while reasoning, temperature, and list prompting help modestly but fall far short of closing the gap. Our cross-dataset analysis shows that \textsc{UnpredictaBench} aligns with utility metrics from creativity benchmarks while offering a statistically grounded alternative to LLM-as-a-judge evaluation. \textbf{The gap between current models and the Random Machine ceiling remains large and unsolved.} 

\section*{Limitations and Broader Impact}
\label{app:limit}

\paragraph{Positive Impact.}
\textsc{UnpredictaBench} targets a capability with direct relevance to simulation, scientific modeling, and decision-making: faithful distributional generation. Many downstream uses of LLMs, including economic, epidemiological, and multi-agent simulations, depend on outputs that reflect a true underlying distribution rather than collapsing onto a few dominant modes. By providing a statistically grounded benchmark and the reusable KS@N\text{KS}@N
KS@N metric, this work offers a concrete target for improving model calibration in stochastic settings, potentially reducing biased estimates and overconfident predictions in applications that require sampling. The benchmark also isolates two distinct failure modes, weak distributional understanding and insufficient output diversity, giving practitioners a clearer diagnostic for where a model breaks down. Our results surface actionable findings that can guide future model development and inform when a model is suitable for simulation-style deployment.

\paragraph{Negative Impact \& Limitations.}
All prompts are in English and 89\% are \textsc{GPT-5.4}-generated, which may introduce phrasing biases and limit generalizability to multilingual or human-authored settings. Code tasks are Python-only, so our conclusions may not transfer to other languages or programming paradigms. The ground-truth distributions reflect the reference samples we construct, and alternative formulations of a task could yield different targets. \textsc{UnpredictaBench} is strictly an evaluation benchmark and is not designed to be used as training data; optimizing directly against it risks overfitting to our specific tasks and metrics, and strong benchmark performance should not be interpreted as real-world deployment readiness. The dataset contains no personal or sensitive information and is released under \texttt{CC BY 4.0} on \href{https://huggingface.co/datasets/UnpredictaBench/UnpredictaBench}
{\raisebox{-0.15em}{\includegraphics[height=1em]{figures/huggingface_logo.png}}~Hugging Face}. The code and ground-truth values are released on \href{https://github.com/UnpredictaBench/UnpredictaBenchCode}
{{\textcolor{black}{\faGithub}}~GitHub}.

\bibliographystyle{plainnat}
\bibliography{references}

\clearpage


\appendix
\section{\textsc{UnpredictaBench} Distributions List}
\label{app:distributions}

\textsc{UnpredictaBench} covers 40 probability distributions across 8 categories, 
listed in Table~\ref{tab:distributions}. The highlighted distributions were used for the multimodal subtask.

\begin{table}[!ht]
\centering
\small
\caption{All 40 probability distributions included in \textsc{UnpredictaBench}, grouped by category.}
\label{tab:distributions}
\renewcommand{\arraystretch}{1.3}
\setlength{\tabcolsep}{8pt}
\begin{tabular}{clcl}
\toprule
\rowcolor{headerblue}
\textbf{\#} & \textbf{Distribution} & \textbf{\#} & \textbf{Distribution} \\
\midrule
\rowcolor{catcolor}
\multicolumn{4}{l}{\textbf{Absolutely Continuous $\cdot$ Bounded Interval}} \\
\cellcolor{yellow!50}1  & \cellcolor{yellow!50}Beta                & \cellcolor{yellow!50}4  & \cellcolor{yellow!50}Triangular \\
\rowcolor{rowgray}
\cellcolor{yellow!50}2  & \cellcolor{yellow!50}Arcsine             & \cellcolor{yellow!50}5  & \cellcolor{yellow!50}Truncated Normal \\
\cellcolor{yellow!50}3  & \cellcolor{yellow!50}Reciprocal          & \cellcolor{yellow!50}6  & \cellcolor{yellow!50}Uniform \\
\midrule
\rowcolor{catcolor}
\multicolumn{4}{l}{\textbf{Absolutely Continuous $\cdot$ Semi-infinite $[0, \infty)$}} \\
\cellcolor{yellow!50}7  & \cellcolor{yellow!50}Erlang              & \cellcolor{yellow!50}13 & \cellcolor{yellow!50}Weibull \\
\rowcolor{rowgray}
\cellcolor{yellow!50}8  & \cellcolor{yellow!50}$F$                 & \cellcolor{yellow!50}14 & \cellcolor{yellow!50}Chi-Squared \\
\cellcolor{yellow!50}9  & \cellcolor{yellow!50}Fr\'{e}chet         & \cellcolor{yellow!50}15 & \cellcolor{yellow!50}Exponential \\
\rowcolor{rowgray}
\cellcolor{yellow!50}10 & \cellcolor{yellow!50}Gamma               & \cellcolor{yellow!50}16 & \cellcolor{yellow!50}Inverse Gaussian \\
\cellcolor{yellow!50}11 & \cellcolor{yellow!50}Pareto              & \cellcolor{yellow!50}17 & \cellcolor{yellow!50}Log-Normal \\
\rowcolor{rowgray}
\cellcolor{yellow!50}12 & \cellcolor{yellow!50}Rayleigh Mixture    &    & \\
\midrule
\rowcolor{catcolor}
\multicolumn{4}{l}{\textbf{Absolutely Continuous $\cdot$ Whole Real Line}} \\
\cellcolor{yellow!50}18 & \cellcolor{yellow!50}Gumbel              & 21 & Logistic \\
\rowcolor{rowgray}
\cellcolor{yellow!50}19 & \cellcolor{yellow!50}Laplace             & 22 & Normal \\
\cellcolor{yellow!50}20 & \cellcolor{yellow!50}Student's $t$       &    & \\
\midrule
\rowcolor{catcolor}
\multicolumn{4}{l}{\textbf{Discrete $\cdot$ Finite Support}} \\
\rowcolor{rowgray}
23 & Bernoulli           & 26 & Binomial \\
24 & Poisson Binomial    & 27 & Discrete Uniform \\
\rowcolor{rowgray}
25 & Beta-Binomial       & 28 & Hypergeometric \\
\midrule
\rowcolor{catcolor}
\multicolumn{4}{l}{\textbf{Discrete $\cdot$ Infinite Support}} \\
29 & Poisson             & 32 & Geometric \\
\rowcolor{rowgray}
30 & Skellam             & 33 & Negative Binomial \\
31 & Compound Poisson    &    & \\
\midrule
\rowcolor{catcolor}
\multicolumn{4}{l}{\textbf{Joint Distributions}} \\
\rowcolor{rowgray}
34 & Dirichlet           & 37 & Multivariate $t$ \\
35 & Multinomial         & 38 & Negative Multinomial \\
\rowcolor{rowgray}
36 & Multivariate Normal &    & \\
\midrule
\rowcolor{catcolor}
\multicolumn{4}{l}{\textbf{Mixed Discrete-Continuous}} \\
39 & Rectified Gaussian  &    & \\
\midrule
\rowcolor{catcolor}
\multicolumn{4}{l}{\textbf{Non-Numeric}} \\
40 & Categorical         &    & \\
\bottomrule
\end{tabular}
\end{table}
\section{Real World Examples}
\label{app:realworld_examples}

This section provides representative examples from the RealWorld category of 
\textsc{UnpredictaBench}. Each task is presented in either code or textual form, and 
models are asked to produce a single plausible output consistent with the underlying 
stochastic process.

\textbf{Example 1: Network Simulation (Code).}
The following example presents a pseudo-code network simulation where two packets are routed through paths with stochastic latency. The ground-truth distribution is over the two possible outputs \texttt{A} and \texttt{B}, with probabilities determined by the \texttt{network\_fluctuation()} function. See Example\hyperref[example:example_real_network]{~\ref*{example:example_real_network}.1}.

\begin{examplebox}{Real World Example (Network Code)}
\label{example:example_real_network}
Consider the following pseudo-code that simulates a network:
\begin{verbatim}
class Packet:
    def __init__(self, name):
        self.name = name
        def send(packet, path):
            return (packet.name, path.get_latency())
    
class Path:    
    def __init__(self, name, base_latency):
        self.name = name
        self.base_latency = base_latency
        def get_latency(self):
            return self.base_latency + network_fluctuation()
        
path1 = Path("P1", 50)
path2 = Path("P2", 70)

packetA = Packet("A")
packetB = Packet("B")

resultA = send(packetA, path1)
resultB = send(packetB, path2)

if resultA[1] < resultB[1]:
    print(resultA[0])
else:
print(resultB[0])

\end{verbatim}
Question: Provide one possible output of this code. Respond with exactly one word only (A or B), and do not include any explanation or mention uncertainty.
\end{examplebox}

\noindent\textbf{Example 2: Garbage Collection (Textual).}
This textual example describes a memory management scenario where three short-lived objects are cleaned up in an unspecified order. The ground-truth distribution is uniform over \texttt{A}, \texttt{B}, and \texttt{C}, reflecting the non-deterministic ordering of garbage collection. See Example \hyperref[example:example_real_garbage]{~\ref*{example:example_real_garbage}.2}.

\begin{examplebox}{Real World Example (Garbage Collection Textual)}
\label{example:example_real_garbage}
You are debugging a performance issue in a program that frequently creates short-lived objects. To better understand memory behavior, you added a small log inside the object finalization routine so that whenever an object is cleaned up, its name is printed.

In one particular run, three temporary objects—A, B, and C—are created during a computation phase. Later in the execution, all three become unused after their references go out of scope. To observe what happens next, you trigger a manual cleanup step.

When reviewing the console output from that run, you notice that only one object name appears immediately after the cleanup is triggered. The program then continues, and no additional relevant logs are shown for that moment.

Objects involved:

- A

- B

- C

Question: What is one possible output? Respond with exactly one word only, and do not include any explanation.
\end{examplebox}

\noindent\textbf{Example 3: Distributed Systems (Textual).}
This example models a replicated key-value store under unstable network conditions, where the responding replica is determined stochastically by which one replies first. The ground-truth distribution is uniform over replicas \texttt{A}, \texttt{B}, and \texttt{C}. See Example\hyperref[example:example_real_distributed]{~\ref*{example:example_real_distributed}.3}.

\begin{examplebox}{Real World Example (Distributed Systems Textual)}
\label{example:example_real_distributed}
You are observing a replicated key-value store under unstable network conditions. A client issues a read request, and multiple replicas may return a value depending on which one responds first. The client logs the replica that served the request.

Replicas:

- A

- B

- C

In the logs for that run, only one replica name appears for that read.

Question: What is one possible output? Respond with exactly one word only, and do not include any explanation.",

\end{examplebox}

\noindent\textbf{Example 4: MCMC State Transition (Code).}
This example presents a code-based task where an LLM agent is queried to decide the next state in a Markov chain transition. The stochasticity arises from the non-determinism of the LLM call itself, making the ground-truth distribution over \texttt{A}, \texttt{B}, and \texttt{C} empirically estimated from repeated execution. See Example\hyperref[example:mcmc]{~\ref*{example:mcmc}.4}.

\begin{examplebox}{Real World Example (MCMC Textual)}
\label{example:mcmc}
    Consider the following code:
    \begin{verbatim}
    from openai import OpenAI
    client = OpenAI()
    state = "idle"
    options = ["A", "B", "C"]
    prompt = f"""An agent is currently in the '{state}' state in a
    
    real-world workflow system.\n\nThe agent needs to decide the next 
    
    state to transition into based on its internal reasoning.\n\n
    
    The possible next states are:\nA\nB\nC\n\nWhich state should 
    
    the agent go to next? Respond with exactly one word.\n"""
    
    response = client.responses.create(model="gpt-4o", input=prompt)
    
    print(response.output_text.strip())
    \end{verbatim}
    
    Question: What is one possible output of this code? Respond with exactly one word only, and do not include any explanation.
\end{examplebox}
\section{Detailed Explanation of Evaluation Metrics}
\label{app:metrics}

\subsection{Handling Sequence-Valued Tasks via Lehmer Codes}
\label{subsec:lehmer}

A subset of \textsc{UnpredictaBench} tasks ask the model to produce a \emph{sequence} 
rather than a scalar. In these cases, both $\mathcal{D}_{\mathrm{pred}}$ and 
$\mathcal{D}_{\mathrm{gt}}$ are distributions over permutations of 
$\{1, 2, \dots, n\}$, and scalar distributional metrics do not apply directly. To 
bring these tasks into a common framework, we encode each permutation $\pi \in S_n$ 
via its \emph{Lehmer code}
\begin{equation}
L_i(\pi) = \big|\{\, j > i : \pi_j < \pi_i \,\}\big|, \qquad i = 1, \dots, n,
\end{equation}
where $L_i(\pi) \in \{0, 1, \dots, n-i\}$ counts the number of elements to the right 
of position $i$ that are smaller than $\pi_i$. The Lehmer code is a bijection between 
$S_n$ and the factorial number system, so no information is lost. We normalize each 
digit by its maximum possible value:
\begin{equation}
Z_i(\pi) =
\begin{cases}
\dfrac{L_i(\pi)}{n - i}, & i < n, \\[6pt]
0, & i = n,
\end{cases}
\end{equation}
so that under a uniformly random permutation, each normalized coordinate $Z_i$ is asymptotically uniform on $[0,1]$. In our evaluation, we focus on the first coordinate $Z_1$, which has the largest support among Lehmer coordinates and therefore provides the richest one-dimensional marginal diagnostic. We apply the scalar distributional metrics directly to $Z_1$.

\subsection{Debiased Wasserstein-1 Distance}
\label{app:wasserstein}

The Wasserstein-1 distance between $\mathcal{D}_{\mathrm{pred}}$ and 
$\mathcal{D}_{\mathrm{gt}}$ is
\begin{equation}
W_1(\mathcal{D}_{\mathrm{pred}}, \mathcal{D}_{\mathrm{gt}}) = 
\frac{1}{N} \sum_{i=1}^{N} \left| a_{(i)} - b_{(i)} \right|,
\end{equation}
where $a_{(i)}$ and $b_{(i)}$ are the $i$-th order statistics of $A$ and $B$. To 
correct for finite-sample bias and enable comparison across tasks with different units, 
we compute a permutation null over $R=999$ random partitions of the pooled sample 
$P = A \cup B$, obtaining null mean $\mu_W$ and standard deviation $\sigma_W$. We 
report the debiased statistic and its $z$-score:
\begin{equation}
\widetilde{W}_1 = W_1(\mathcal{D}_{\mathrm{pred}}, \mathcal{D}_{\mathrm{gt}}) - \mu_W, 
\qquad
Z_{W_1} = \frac{W_1(\mathcal{D}_{\mathrm{pred}}, \mathcal{D}_{\mathrm{gt}}) - 
\mu_W}{\sigma_W}.
\end{equation}
Values of $Z_{W_1}$ near zero indicate indistinguishability from chance; larger values 
indicate systematic distributional mismatch.

\subsection{Jensen--Shannon Divergence}
\label{app:jsd}

We fit Gaussian kernel density estimates $\hat{p}_{\mathcal{D}_{\mathrm{pred}}}$ and 
$\hat{p}_{\mathcal{D}_{\mathrm{gt}}}$ to $A$ and $B$, evaluate them on a shared grid 
of $G=512$ points covering the union of supports with $10\%$ padding, and normalize 
to obtain discrete distributions $p$ and $q$. The Jensen--Shannon divergence is then
\begin{equation}
\mathrm{JSD}(\mathcal{D}_{\mathrm{pred}} \,\|\, \mathcal{D}_{\mathrm{gt}}) = 
\frac{1}{2}\mathrm{KL}(p \,\|\, m) + \frac{1}{2}\mathrm{KL}(q \,\|\, m), \qquad 
m = \frac{p + q}{2},
\end{equation}
where $\mathrm{KL}(u \,\|\, v) = \sum_k u_k \log \frac{u_k}{v_k}$. JSD is symmetric, 
bounded in $[0, \log 2]$, and well-defined even when supports do not overlap, capturing 
density-level shape mismatches that distance-based metrics can underweight.
\section{Extended Model Results}
\label{app:more_models}

Table~\ref{tab:more_models} extends the category-level results of Table~\ref{tab:main} to the full set of evaluated models, reporting KS@100, JSD, and WDZ across all four task categories. The patterns observed in the main paper hold broadly: RealWorld tasks yield the highest individual scores while Code and Text remain the most demanding, and models with strong overall KS@100 tend to show consistently lower JSD and WDZ values. Llama-3.2-1B-instruct again stands out with a remarkable 59.09\% on RealWorld despite near-bottom performance elsewhere, and the Qwen3.5 MoE variants continue to underperform relative to their parameter counts across all categories.

\begin{table}[ht]
\caption{Per-category results for the full set of evaluated models, reporting 
KS@100 (↑: higher is better), Jensen--Shannon Divergence (JSD, ↓: lower is better), 
and Wasserstein Distance Z-score (WDZ, ↓: closer to zero is better) across Code, Text, 
RealWorld, and Shuffling task categories. This table extends Table~\ref{tab:main} in 
the main paper to include all models evaluated in this work.}
\label{tab:more_models}
\renewcommand{\arraystretch}{1.5}
\resizebox{\linewidth}{!}{%
\begin{tabular}{l ccc ccc ccc ccc}
\toprule
\multirow{2}{*}{\textbf{Model}}
  & \multicolumn{3}{c}{\textbf{Code}}
  & \multicolumn{3}{c}{\textbf{Text}}
  & \multicolumn{3}{c}{\textbf{RealWorld}}
  & \multicolumn{3}{c}{\textbf{Shuffling}} \\
\cmidrule(lr){2-4}
\cmidrule(lr){5-7}
\cmidrule(lr){8-10}
\cmidrule(lr){11-13}
& \textbf{KS@100↑} & \textbf{JSD↓} & \textbf{WDZ↓}
& \textbf{KS@100↑} & \textbf{JSD↓} & \textbf{WDZ↓}
& \textbf{KS@100↑} & \textbf{JSD↓} & \textbf{WDZ↓}
& \textbf{KS@100↑} & \textbf{JSD↓} & \textbf{WDZ↓} \\
\midrule
\texttt{Nemotron-3-super-120B-a12b}      & 40.34                     & 0.48                     & 7.44                       & 28.13                     & 0.30                     & 5.89                       & 3.33                      & 0.16                     & 35.36                      & 21.05                     & 0.15                     & 10.13                     \\
\texttt{Nemotron-3-nano-30B-a3b}         & 18.49                     & 0.24                     & 10.61                      & 20.63                     & 0.29                     & 11.11                      & 33.33                     & 0.19                     & 12.60                      & 31.58                     & 0.12                     & 8.11                      \\
\texttt{GPT-5.4}                         & 10.50                     & 0.30                     & 10.58                      & 25.63                     & 0.19                     & 9.03                       & 6.67                      & 0.25                     & 19.00                      & 0                         & 0.26                     & 16.81                     \\
\texttt{GPT-4o}                          & 29.41                     & 0.33                     & 8.80                       & 21.25                     & 0.18                     & 11.92                      & 6.67                      & 0.22                     & 17.60                      & 5.26                      & 0.19                     & 12.27                     \\
\texttt{GPT-4o-mini}                     & 10.50                     & 0.25                     & 12.75                      & 9.38                      & 0.26                     & 16.49                      & 10                        & 0.26                     & 17.88                      & 0                         & 0.22                     & 14.47                     \\
\texttt{Mistral-large-2512}           & 5.04                      & 0.29                     & 14.75                      & 3.75                      & 0.32                     & 17.24                      & 10                        & 0.22                     & 17.91                      & 0                         & 0.27                     & 16.15                     \\
\texttt{Ministral-3-3B-instruct-2512} & 17.23                     & 0.37                     & 12.11                      & 21.25                     & 0.18                     & 1.61                       & 6.67                      & 0.13                     & 36.14                      & 5.26                      & 0.15                     & 9.67                      \\
\texttt{OLMo-3-7B-instruct}             & 5.46                      & 0.46                     & 34.93                      & 7.50                      & 0.29                     & 20.74                      & 3.33                      & 0.26                     & 59.16                      & 36.84                     & 0.13                     & 7.64                      \\
\texttt{DeepSeek-v3.2}                 & 14.29                     & 0.26                     & 13.05                      & 28.13                     & 0.18                     & 12.28                      & 36.67                     & 0.15                     & 13.18                      & 36.84                     & 0.13                     & 8.58                      \\
\texttt{Qwen3.5-397B-a17b}                 & 2.94                      & 0.45                     & 15.21                      & 3.8                       & 0.51                     & 17.50                      & 3.33                      & 0.34                     & 22.44                      & 5.26                      & 0.18                     & 17.23                     \\
\texttt{Qwen3.5-35B-a3b}                   & 2.94                      & 0.55                     & 16.62                      & 3.75                      & 0.52                     & 19.35                      & 3.57                      & 0.43                     & 25.46                      & 0                         & 0.31                     & 18.11                     \\
\texttt{Qwen3-32B}                         & 5.46                      & 0.27                     & 12.81                      & 11.88                     & 0.24                     & 14.30                      & 16.67                     & 0.23                     & 17.33                      & 0                         & 0.19                     & 13.21                     \\
\texttt{Qwen3.5-2B}                        & 12.61 & 0.29 & 10.87 & 14.38 & 0.35 & 11.30 & 31.03 & 0.17 & 12.32 & 36.84 & 0.11 & 3.18  \\
\texttt{Mercury-2}                    & 7.14  & 0.27 & 10.12 & 10.63 & 0.31 & 12.60 & 13.33 & 0.20 & 13.97 & 0.00  & 0.20 & 12.91 \\
\texttt{Grok-4.1-fast}                     & 5.46  & 0.47 & 15.26 & 6.88  & 0.33 & 16.37 & 3.33  & 0.21 & 19.48 & 0     & 0.19 & 12.69 \\
\texttt{Claude-sonnet-4.6}            & 5.04  & 0.25 & 13.40 & 5.03  & 0.34 & 13.27 & 3.33  & 0.31 & 19.61 & 0     & 0.22 & 18.17 \\
\texttt{Llama-3.1-70B-instruct}      & 9.66  & 0.26 & 12.96 & 23.13 & 0.21 & 15.01 & 33.33 & 0.15 & 12.23 & 21.05 & 0.15 & 10.21 \\
\texttt{Llama-3.1-8B-instruct}       & 4.62  & 0.29 & 21.88 & 3.75  & 0.34 & 22.70 & 30.00 & 0.17 & 14.52 & 15.79 & 0.15 & 9.49  \\
\texttt{Llama-3.2-1B-instruct}       & 4.20  & 0.52 & 29.82 & 10    & 0.57 & 14.94 & 59.09 & 0.20 & 11.35 & 36.84 & 0.12 & 6.26  \\
\texttt{Phi-3.5-mini-instruct}        & 2.94  & 0.52 & 17.44 & 1.88  & 0.35 & 20.02 & 10    & 0.29 & 18.93 & 0     & 0.19 & 13.21 \\ \hline
\end{tabular}
}

\end{table}
\section{Per-Distribution KS@100 Breakdown}
\label{app:dist_difficulty}

Table~\ref{tab:dist_acc} reports KS@N broken down by target distribution, averaged across all models and task formats. Cells are highlighted relative to the per-column average: distributions above average are marked as easier and those below as harder. A clear pattern emerges: \textbf{simple discrete distributions with small finite support are consistently the easiest}, with Bernoulli (43.04\%), Categorical (34.78\%), and Discrete Uniform (16.52\%) leading at KS@100. This is unsurprising given that models are likely to have encountered these distributions frequently during pretraining and their support is small enough that even limited diversity suffices to pass the KS test. At the other end, \textbf{heavy-tailed and multivariate distributions prove the most challenging}: Fréchet (1.74\%), Dirichlet (1.74\%), Negative Binomial (5.22\%), and Negative Multinomial (6.09\%) rank at the bottom at KS@100, reflecting the difficulty of reproducing long tails and correlated multivariate structure. Compound Poisson, Erlang, Inverse Gaussian, and Pareto all cluster below 9\% at KS@100, suggesting that distributions requiring precise scale and shape calibration are particularly problematic. Notably, the Beta distribution shows a sharp drop from KS@10 (87.39\%) to KS@100 (4.78\%), one of the steepest in the table, consistent with our qualitative finding in Section~\ref{sec:qualitative} that bounded continuous distributions suffer from severe support misspecification at the logit level.

\begin{table}[ht]
\centering
\caption{KS@N averaged across all models and task formats, broken down by target distribution, at thresholds $N \in \{10, 20, 50, 100\}$. Cells are highlighted relative to the per-column mean: \colorbox{easy}{above average} distributions are relatively easier, while \colorbox{hard}{below average} distributions are harder. Distributions are ordered roughly from easiest to hardest at KS@100.}
\label{tab:dist_acc}
\resizebox{0.7\textwidth}{!}{%
\begin{tabular}{l c c c c}
\toprule
\textbf{Distribution}
  & \textbf{KS@10} & \textbf{KS@20}
  & \textbf{KS@50} & \textbf{KS@100} \\
\midrule

Bernoulli           & \cellcolor{easy}92.61 & \cellcolor{easy}68.26 & \cellcolor{easy}52.17 & \cellcolor{easy}43.04 \\
Categorical         & \cellcolor{easy}86.52 & \cellcolor{easy}71.30 & \cellcolor{easy}48.26 & \cellcolor{easy}34.78 \\
Rectified Gaussian  & \cellcolor{easy}78.26 & \cellcolor{easy}55.65 & \cellcolor{easy}35.22 & \cellcolor{easy}24.35 \\
Geometric           & \cellcolor{hard}67.83 & \cellcolor{easy}52.17 & \cellcolor{easy}30.00 & \cellcolor{easy}23.04 \\
Multivariate T      & \cellcolor{easy}83.41 & \cellcolor{easy}62.01 & \cellcolor{easy}31.88 & \cellcolor{easy}16.16 \\
Triangular          & \cellcolor{easy}85.22 & \cellcolor{easy}59.13 & \cellcolor{easy}33.91 & \cellcolor{easy}15.65 \\
Discrete Uniform    & \cellcolor{easy}90.87 & \cellcolor{easy}69.13 & \cellcolor{easy}33.04 & \cellcolor{easy}16.52 \\
Student's T         & \cellcolor{hard}74.78 & \cellcolor{easy}52.17 & \cellcolor{easy}28.26 & \cellcolor{easy}14.78 \\
Poisson Binomial    & \cellcolor{hard}77.39 & \cellcolor{easy}54.35 & \cellcolor{easy}24.35 & \cellcolor{easy}15.22 \\
Multinomial         & \cellcolor{hard}76.52 & \cellcolor{easy}53.91 & \cellcolor{easy}25.22 & \cellcolor{easy}12.61 \\
Normal              & \cellcolor{hard}76.52 & \cellcolor{hard}51.30 & \cellcolor{easy}29.57 & \cellcolor{easy}13.04 \\
Uniform             & \cellcolor{easy}90.43 & \cellcolor{easy}63.04 & \cellcolor{easy}27.83 & \cellcolor{easy}13.48 \\
Log Normal          & \cellcolor{hard}73.91 & \cellcolor{hard}49.57 & \cellcolor{hard}18.70 & \cellcolor{easy}10.87 \\
Gumbel              & \cellcolor{easy}83.48 & \cellcolor{easy}52.61 & \cellcolor{easy}25.65 & \cellcolor{easy}11.30 \\
Skellam             & \cellcolor{easy}84.35 & \cellcolor{easy}63.04 & \cellcolor{easy}24.35 & \cellcolor{easy}13.48 \\
Laplace             & \cellcolor{hard}78.70 & \cellcolor{easy}58.70 & \cellcolor{easy}28.26 & \cellcolor{easy}10.87 \\
Weibull             & \cellcolor{hard}72.61 & \cellcolor{easy}53.04 & \cellcolor{hard}22.17 & \cellcolor{easy}11.30 \\
Beta Binomial       & \cellcolor{hard}79.13 & \cellcolor{easy}57.39 & \cellcolor{easy}25.65 & \cellcolor{easy}11.30 \\
Multivariate Normal & \cellcolor{easy}83.48 & \cellcolor{easy}56.09 & \cellcolor{hard}23.04 & \cellcolor{easy}10.43 \\
Binomial            & \cellcolor{easy}83.91 & \cellcolor{easy}56.96 & \cellcolor{easy}25.65 & \cellcolor{easy}10.00 \\
Arcsine             & \cellcolor{easy}86.52 & \cellcolor{easy}62.17 & \cellcolor{easy}24.35 & \cellcolor{easy}10.43 \\
Truncated Normal    & \cellcolor{easy}84.35 & \cellcolor{easy}55.22 & \cellcolor{easy}26.09 & \cellcolor{easy}10.43 \\
Poisson             & \cellcolor{hard}79.57 & \cellcolor{hard}52.17 & \cellcolor{hard}23.04 & \cellcolor{hard}9.57 \\
Compound Poisson    & \cellcolor{hard}56.96 & \cellcolor{hard}35.65 & \cellcolor{hard}17.39 & \cellcolor{hard}8.70 \\
Reciprocal          & \cellcolor{easy}81.30 & \cellcolor{easy}55.22 & \cellcolor{hard}17.83 & \cellcolor{hard}9.13 \\
Chi Squared         & \cellcolor{hard}70.00 & \cellcolor{hard}44.78 & \cellcolor{hard}18.70 & \cellcolor{hard}8.70 \\
Exponential         & \cellcolor{hard}73.04 & \cellcolor{hard}48.26 & \cellcolor{hard}19.13 & \cellcolor{hard}8.26 \\
Erlang              & \cellcolor{hard}56.52 & \cellcolor{hard}35.22 & \cellcolor{hard}14.78 & \cellcolor{hard}8.26 \\
Hypergeometric      & \cellcolor{hard}58.70 & \cellcolor{hard}39.57 & \cellcolor{hard}17.83 & \cellcolor{hard}6.96 \\
Inverse Gaussian    & \cellcolor{hard}56.96 & \cellcolor{hard}34.78 & \cellcolor{hard}16.52 & \cellcolor{hard}7.39 \\
F                   & \cellcolor{hard}75.65 & \cellcolor{hard}43.48 & \cellcolor{hard}22.17 & \cellcolor{hard}7.83 \\
Pareto              & \cellcolor{hard}56.52 & \cellcolor{hard}36.52 & \cellcolor{hard}13.48 & \cellcolor{hard}6.96 \\
Negative Multinomial& \cellcolor{hard}53.04 & \cellcolor{hard}26.52 & \cellcolor{hard}10.87 & \cellcolor{hard}6.09 \\
Logistic            & \cellcolor{easy}83.91 & \cellcolor{easy}57.83 & \cellcolor{hard}20.87 & \cellcolor{hard}6.52 \\
Gamma               & \cellcolor{hard}73.04 & \cellcolor{hard}51.74 & \cellcolor{hard}19.57 & \cellcolor{hard}6.09 \\
Rayleigh Mixture    & \cellcolor{hard}78.26 & \cellcolor{hard}47.28 & \cellcolor{hard}23.37 & \cellcolor{hard}5.98 \\
Negative Binomial   & \cellcolor{hard}50.87 & \cellcolor{hard}31.74 & \cellcolor{hard}10.43 & \cellcolor{hard}5.22 \\
Beta                & \cellcolor{easy}87.39 & \cellcolor{easy}56.52 & \cellcolor{hard}13.91 & \cellcolor{hard}4.78 \\
Dirichlet           & \cellcolor{hard}57.83 & \cellcolor{hard}26.96 & \cellcolor{hard}6.52  & \cellcolor{hard}1.74 \\
Fr\'{e}chet         & \cellcolor{hard}56.09 & \cellcolor{hard}26.09 & \cellcolor{hard}6.96  & \cellcolor{hard}1.74 \\

\bottomrule
\end{tabular}%
}
\end{table}
\section{Explicit vs.\ Implicit Prompting}
\label{app:explicit_vs_implicit}

Table~\ref{tab:explicit_implicit} compares model performance under explicit and implicit prompting conditions. In the explicit setting, the distribution is directly named or described, while in the implicit setting the model must infer the underlying stochastic process from context without being told the distribution family. Overall, \textbf{explicit prompting yields higher KS@N for the majority of models}, which is consistent with the intuition that naming a distribution reduces the problem to parameter estimation and sampling, whereas implicit prompting additionally requires distributional inference. Nemotron-3 Super 120B leads in both settings (41.42\% and 26.42\% at KS@100 respectively), and the gap between explicit and implicit is substantial (15 percentage points), suggesting that even the strongest model benefits considerably from being told what distribution to sample from. Interestingly, \textbf{a handful of models perform better implicitly than explicitly}: GPT-5.4 (20.13\% vs.\ 14.23\%), DeepSeek V3.2 (23.27\% vs.\ 17.57\%), OLMo-3 7B (8.18\% vs.\ 5.02\%), and Claude Sonnet 4.6 (6.92\% vs.\ 3.78\%) all show higher KS@100 under implicit prompting. This counterintuitive result may reflect the fact that when a distribution is named explicitly, these models anchor too strongly on a memorized prototype of that distribution rather than adapting to the specific parameterization given in the prompt. In the implicit setting, without a named anchor, they may rely more on contextual reasoning, which for certain task types yields better-calibrated outputs. At the lower end of the table, the explicit/implicit gap narrows considerably, suggesting that \textbf{for weaker models the bottleneck is not prompt format but fundamental distributional understanding}.

\begin{table}[ht]
\centering
\caption{KS@50 and KS@100 under explicit and implicit prompting conditions for all evaluated models. In the explicit setting the target distribution is directly named or described; in the implicit setting the model must infer the distributional structure from context. Bold values indicate the best score in each column.}
\label{tab:explicit_implicit}
\resizebox{0.85\textwidth}{!}{%
\begin{tabular}{l cc cc}
\toprule
\multirow{2}{*}{\textbf{Model}}
  & \multicolumn{2}{c}{\textbf{Explicit}}
  & \multicolumn{2}{c}{\textbf{Implicit}} \\
\cmidrule(lr){2-3} \cmidrule(lr){4-5}
  & \textbf{KS@50} & \textbf{KS@100}
  & \textbf{KS@50} & \textbf{KS@100} \\
\midrule

\texttt{Nemotron-3-super-120B-a12b}  & 61.92 & 41.42 & 47.80 & 26.42 \\
\texttt{GPT-4o}                      & 48.95 & 30.96 & 40.88 & 18.87 \\
\texttt{Nemotron-3-nano-30B}         & 45.19 & 20.50 & 30.82 & 17.61 \\
\texttt{GPT-5.4}                     & 31.80 & 14.23 & 37.11 & 20.13 \\
\texttt{Ministral-3B-instruct}       & 32.22 & 17.15 & 28.93 & 15.72 \\
\texttt{Ministral-3B-base}           & 33.89 & 17.15 & 30.19 & 18.87 \\
\texttt{Qwen3.5-2B-base}             & 37.24 & 18.83 & 32.70 & 13.21 \\
\texttt{DeepSeek-V3.2}               & 38.49 & 17.57 & 32.70 & 23.27 \\
\texttt{Qwen-3.5-2B}                 & 33.05 & 13.39 & 28.93 & 13.21 \\
\texttt{Llama-3.1-70B-instruct}      & 27.62 & 17.15 & 32.08 & 11.95 \\
\texttt{GPT-4o-mini}                 & 25.10 & 11.30 & 16.98 & 8.18  \\
\texttt{Llama-3.2-1B}                & 25.10 & 12.97 & 20.13 & 8.81  \\
\texttt{Mercury-2}                   & 22.59 & 10.88 & 17.61 & 5.03  \\
\texttt{OLMo-3-7B-instruct}          & 16.74 & 5.02  & 20.75 & 8.18  \\
\texttt{Llama-3.1-8B-instruct}       & 14.64 & 4.60  & 15.09 & 3.77  \\
\texttt{Mistral-Large}               & 11.30 & 5.02  & 8.18  & 3.77  \\
\texttt{Qwen3-32B}                   & 20.50 & 8.79  & 15.72 & 6.92  \\
\texttt{Claude-sonnet-4.6}           & 7.98  & 3.78  & 11.32 & 6.92  \\
\texttt{Grok-4.1-fast}               & 8.37  & 5.86  & 9.43  & 6.29  \\
\texttt{Phi-3.5-mini-instruct}       & 7.53  & 2.09  & 7.55  & 3.14  \\
\texttt{Llama-3.2-1B-instruct}       & 4.18  & 1.67  & 7.55  & 4.40  \\
\texttt{Qwen-3.5-397B-a17b}          & 3.35  & 2.93  & 6.92  & 3.77  \\
\texttt{Qwen-3.5-35B-a3b}            & 3.35  & 2.93  & 5.66  & 3.77  \\

\bottomrule
\end{tabular}%
}
\end{table}
\section{Unimodal vs.\ Multimodal Distribution Complexity}
\label{app:multimodal_analysis}

Table~\ref{tab:multimodal_unimodal} compares model performance on unimodal tasks, where the target is a single distribution, against multimodal tasks, where the target is a mixture of two component distributions. The results reveal a nuanced picture that differs notably across models. \textbf{For the strongest models, multimodal tasks are actually easier}: Nemotron-3 Super 120B achieves 42.50\% on multimodal versus 33.65\% on unimodal at KS@100, and GPT-4o similarly scores 38.75\% versus 22.96\%. We hypothesize that this reflects the fact that mixture distributions, by construction, have broader and more spread-out support, making it easier for a diverse model to pass the KS test even with imperfect mode coverage. \textbf{For weaker models, the pattern reverses}: Mercury-2 (1.25\% vs.\ 10.38\%), Claude Sonnet 4.6 (0.00\% vs.\ 6.31\%), Phi-3.5 Mini (0.00\% vs.\ 3.14\%), and both Qwen3.5 MoE variants (0.00\% vs.\ $\sim$4\%) all collapse entirely on multimodal tasks while retaining some performance on unimodal ones. This suggests that \textbf{capturing a mixture distribution requires the model to simultaneously maintain multiple modes in its output, a form of diversity that models with strong deterministic tendencies cannot sustain}. GPT-5.4 presents the starkest reversal: 7.50\% on multimodal versus 18.87\% on unimodal at KS@100, consistent with its tendency to collapse to a single point as illustrated in Figure~\hyperref[fig:teaser]{\ref*{fig:teaser}(a)}, which is particularly damaging when the target has two well-separated modes.

\begin{table}[ht]
\centering
\caption{KS@50 and KS@100 for unimodal and multimodal target distributions. Unimodal tasks involve sampling from a single target distribution, while multimodal tasks require matching a mixture of two component distributions. Bold values indicate the best score in each column; underlined values indicate the second best.}
\label{tab:multimodal_unimodal}
\resizebox{0.85\textwidth}{!}{%
\begin{tabular}{l cc cc}
\toprule
\multirow{2}{*}{\textbf{Model}}
  & \multicolumn{2}{c}{\textbf{Multimodal}}
  & \multicolumn{2}{c}{\textbf{Unimodal}} \\
\cmidrule(lr){2-3} \cmidrule(lr){4-5}
  & \textbf{KS@50} & \textbf{KS@100}
  & \textbf{KS@50} & \textbf{KS@100} \\
\midrule

\texttt{Nemotron-3-super-120B-a12b}  & 63.75 & 42.50 & 54.40 & 33.65 \\
\texttt{GPT-4o}                      & 52.50 & 38.75 & 44.03 & 22.96 \\
\texttt{Qwen-3.5-2B-base}             & 43.75 & 16.25 & 33.33 & 16.67 \\
\texttt{Qwen-3.5-2B}                 & 38.75 & 16.25 & 29.56 & 12.58 \\
\texttt{Nemotron-3-nano-30B}         & 47.50 & 17.50 & 37.42 & 19.81 \\
\texttt{GPT-5.4}                     & 15.00 & 7.50  & 38.68 & 18.87 \\
\texttt{DeepSeek-V3.2}               & 31.25 & 13.75 & 37.42 & 21.38 \\
\texttt{Ministral-3B-base}           & 30.00 & 13.75 & 33.02 & 18.87 \\
\texttt{Ministral-3B-instruct}       & 26.25 & 8.75  & 32.08 & 18.55 \\
\texttt{Llama-3.1-70B-instruct}      & 23.75 & 8.75  & 30.82 & 16.67 \\
\texttt{GPT-4o-mini}                 & 22.50 & 11.25 & 21.70 & 9.75  \\
\texttt{Qwen3-32B}                   & 20.00 & 7.50  & 18.24 & 8.18  \\
\texttt{Llama-3.2-1B}                & 27.50 & 16.25 & 22.01 & 10.06 \\
\texttt{Mercury-2}                   & 10.00 & 1.25  & 23.27 & 10.38 \\
\texttt{OLMo-3-7B-instruct}          & 13.75 & 3.75  & 19.50 & 6.92  \\
\texttt{Llama-3.1-8B-instruct}       & 13.75 & 3.75  & 15.09 & 4.40  \\
\texttt{Mistral-Large}               & 12.50 & 3.75  & 9.43  & 4.72  \\
\texttt{Llama-3.2-1B-instruct}       & 7.50  & 2.50  & 5.03  & 2.83  \\
\texttt{Claude-sonnet-4.6}           & 5.00  & 0.00  & 10.41 & 6.31  \\
\texttt{Phi-3.5-mini-instruct}       & 5.00  & 0.00  & 8.18  & 3.14  \\
\texttt{Grok-4.1-fast}               & 2.50  & 2.50  & 10.38 & 6.92  \\
\texttt{Qwen-3.5-397B-a17b}          & 0.00  & 0.00  & 5.97  & 4.09  \\
\texttt{Qwen-3.5-35B-a3b}            & 0.00  & 0.00  & 5.35  & 4.09  \\

\bottomrule
\end{tabular}%
}
\end{table}
\section{Effect of Distributional Spread}
\label{app:concentrated_spred}

Table~\ref{tab:concentrated_spread} compares performance on concentrated distributions, which have low variance and most mass near the mean, against spread out distributions with high variance and broad support. The results reveal a striking model-dependent reversal. \textbf{For most strong models, concentrated and spread out tasks are roughly equally challenging}, with Nemotron-3 Super 120B performing comparably in both settings (36.36\% vs.\ 34.50\% at KS100) and GPT-4o similarly close (27.27\% vs.\ 25.00\%). However, \textbf{for many mid-range and weaker models, spread out distributions are substantially harder}: Grok-4.1-fast (11.62\% vs.\ 0.50\%), Claude Sonnet 4.6 (9.64\% vs.\ 0.50\%), Phi-3.5 Mini (4.55\% vs.\ 0.50\%), and both Qwen3.5 MoE variants (around 6.5\% vs.\ 0.00\%) collapse almost entirely on spread out distributions. This is consistent with our hypothesis that \textbf{deterministically trained models anchor near the mode of a distribution}, which is a reasonable strategy for concentrated distributions but catastrophically fails when the true distribution has broad support and significant tail mass. Conversely, a small number of models perform better on spread out tasks: Ministral-3B instruct (21.00\% vs.\ 12.12\%), Llama-3.2-1B (16.50\% vs.\ 6.06\%), and Llama-3.1-8B instruct (6.50\% vs.\ 2.02\%) all show higher KS100 on spread out distributions, suggesting these models generate outputs diverse enough to cover broad support but insufficiently precise to match the tighter mass concentration required by low-variance distributions.

\begin{table}[ht]
\centering
\caption{KS@50 and KS@100 for concentrated (low-variance) and spread out (high-variance) target distributions. Bold values indicate the best score in each column; underlined values indicate the second best.}
\label{tab:concentrated_spread}
\resizebox{0.85\textwidth}{!}{%
\begin{tabular}{l cc cc}
\toprule
\multirow{2}{*}{\textbf{Model}}
  & \multicolumn{2}{c}{\textbf{Concentrated}}
  & \multicolumn{2}{c}{\textbf{Spread Out}} \\
\cmidrule(lr){2-3} \cmidrule(lr){4-5}
  & \textbf{KS@50} & \textbf{KS@100}
  & \textbf{KS@50} & \textbf{KS@100} \\
\midrule

\texttt{Nemotron-3-super-120B-a12b}  & 59.60 & 36.36 & 53.00 & 34.50 \\
\texttt{GPT-4o}                      & 45.96 & 27.27 & 45.50 & 25.00 \\
\texttt{Nemotron-3-nano-30B}         & 32.32 & 15.15 & 46.50 & 23.50 \\
\texttt{GPT-5.4}                     & 31.31 & 18.18 & 36.50 & 15.00 \\
\texttt{DeepSeek-V3.2}               & 32.32 & 17.17 & 40.00 & 22.50 \\
\texttt{Qwen-3.5-2B-base}             & 36.36 & 15.15 & 34.50 & 18.00 \\
\texttt{Ministral-3B-base}           & 27.78 & 16.67 & 37.00 & 19.00 \\
\texttt{Ministral-3B-instruct}       & 24.24 & 12.12 & 37.50 & 21.00 \\
\texttt{Llama-3.1-70B-instruct}      & 31.31 & 16.67 & 27.50 & 13.50 \\
\texttt{Qwen-3.5-2B}                 & 25.76 & 11.11 & 37.00 & 15.50 \\
\texttt{GPT-4o-mini}                 & 26.77 & 15.15 & 17.00 & 5.00  \\
\texttt{Mercury-2}                   & 22.73 & 10.61 & 18.50 & 6.50  \\
\texttt{OLMo-3-7B-instruct}          & 19.19 & 9.60  & 17.50 & 3.00  \\
\texttt{Qwen3-32B}                   & 17.68 & 11.11 & 19.50 & 5.00  \\
\texttt{Mistral-Large}               & 14.65 & 7.07  & 5.50  & 2.00  \\
\texttt{Grok-4.1-fast}               & 14.14 & 11.62 & 3.50  & 0.50  \\
\texttt{Claude-sonnet-4.6}           & 15.23 & 9.64  & 3.50  & 0.50  \\
\texttt{Phi-3.5-mini-instruct}       & 11.62 & 4.55  & 3.50  & 0.50  \\
\texttt{Llama-3.2-1B}                & 11.11 & 6.06  & 35.00 & 16.50 \\
\texttt{Llama-3.1-8B-instruct}       & 9.09  & 2.02  & 20.50 & 6.50  \\
\texttt{Qwen-3.5-397B-a17b}          & 9.60  & 6.57  & 0.00  & 0.00  \\
\texttt{Qwen-3.5-35B-a3b}            & 8.59  & 6.57  & 0.00  & 0.00  \\
\texttt{Llama-3.2-1B-instruct}       & 2.02  & 0.51  & 9.00  & 5.00  \\

\bottomrule
\end{tabular}%
}
\end{table}
\section{Error Analysis}
\label{app:error_analysis}

\begin{figure}
    \caption{Mean and min-max range of KS@100 (\%), Jensen-Shannon Divergence (JSD), and Wasserstein Z-score Distance (WDZ) across three repeated evaluation runs for six models, broken down by task category (Code, Text, RealWorld, Shuffling). Each dot represents an individual run; the larger marker shows the mean and error bars span the full observed range. The consistently narrow error bars confirm that benchmark results are stable across runs, validating the use of a single evaluation run in the main paper.}
    \label{fig:error_analysis}
    \centering
    \includegraphics[width=\linewidth]{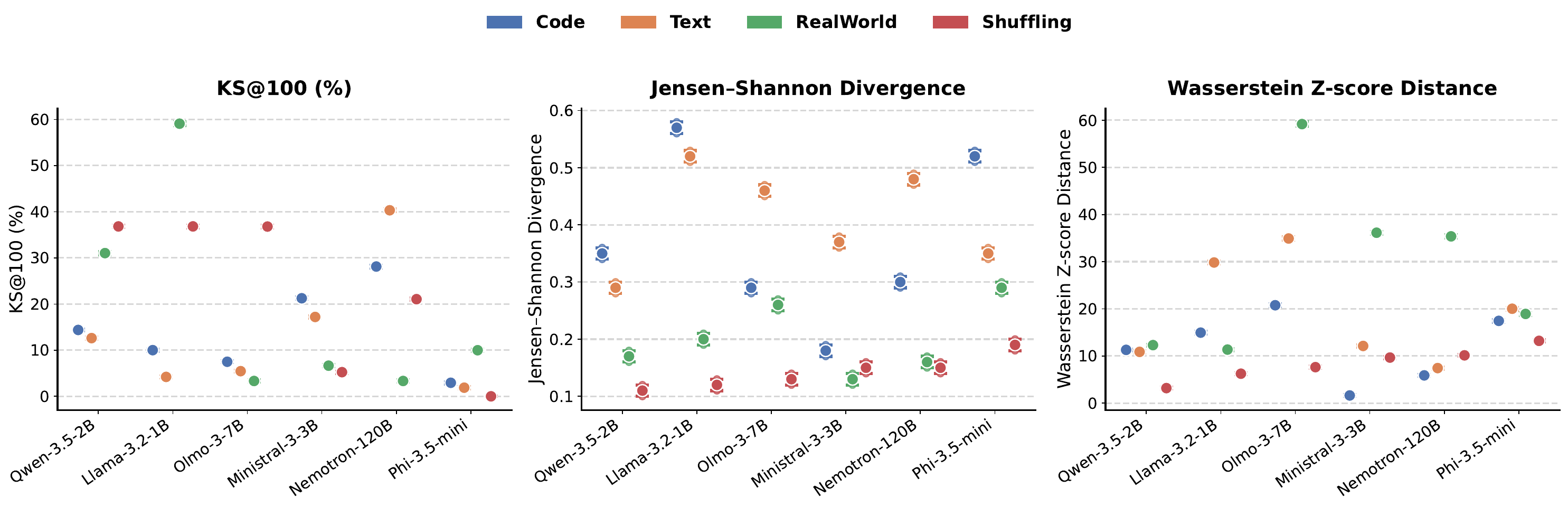}
\end{figure}

Figure~\ref{fig:error_analysis} reports the mean and min-max range of KS@100, JSD, and WDZ across three repeated evaluation runs for six models, broken down by task category. The error bars are consistently narrow across all models and metrics, confirming that \textbf{our benchmark results are stable and reproducible}: the variance introduced by ground-truth resampling is negligible relative to the differences observed between models and categories. This validates the use of a single evaluation run for the main results reported in the paper.

Beyond stability, the figure reinforces several patterns from the main analysis. Llama-3.2-1B's RealWorld KS@100 (around 59\%) stands out as both high and stable, while its Text JSD (around 0.52) and WDZ (around 29.82) are among the worst and equally stable, confirming that its strong RealWorld performance is a genuine distributional property rather than an evaluation artifact. Nemotron-3 Super 120B shows tight error bars on its high Text KS@100 (40.34\%) alongside a notably high Text JSD (0.48), a consistent tension between the KS-based KS@100 and distributional distance metrics that holds across all three runs. OLMo-3 7B's RealWorld WDZ is persistently the highest in the table (around 59.16), with very little variance, suggesting this is a stable structural failure rather than a sampling fluke. \textbf{The tight confidence intervals across all models and metrics give us confidence that the rankings and conclusions in the main paper are robust to evaluation noise.}

\section{Ground Truth Sensitivity}
\label{app:ground_truth_sensitivity}

To assess the sensitivity of our evaluation to the choice of ground truth samples, we generate three independent sets of ground truth values, each consisting of 1,000 samples drawn from the true distribution for each problem, and report the standard deviation of KS@100, JSD, and WDZ across these three sets in Table~\ref{tab:sigma_comparison}. The standard deviations are small across all models and categories, confirming that \textbf{our evaluation is robust to the specific ground truth sample set used}. KS@100 standard deviations remain below 0.42 for all models, and JSD and WDZ deviations are similarly tight for the majority of models, with the slight increase from Random to RealWorld settings reflecting the naturally higher variability of real-world distributions. Qwen-3.5-35B-a3b shows the highest JSD instability (up to 2.55), while Llama-3.1-70B and DeepSeek V3.2 are among the most stable models across all metrics and settings.

To ensure full reproducibility, we fix and release the exact ground truth samples used in this evaluation. \textbf{Upon acceptance, we will release both the ground truth generation code and the fixed ground truth sets used for this submission}, enabling direct replication of our reported numbers. Users who wish to evaluate under different conditions are free to adapt the generation code to produce their own ground truth sets, for instance by increasing the number of samples, changing the random seed, or substituting alternative sampling procedures. The KS@N metric and our evaluation framework are designed to be fully agnostic to the specific ground truth instantiation, making such adaptations straightforward.

\begin{table}[ht]
\centering
\caption{Standard deviation ($\sigma$) of KS@100 (\%), JSD, and WDZ across three independent ground truth sets of 1,000 samples each, reported for Random, Shuffling, and RealWorld evaluation settings. Lower standard deviation indicates greater robustness to the choice of ground truth samples. Bold values highlight the highest and lowest standard deviations within each metric column.}
\label{tab:sigma_comparison}
\resizebox{\textwidth}{!}{%
\begin{tabular}{l ccc ccc ccc}
\toprule
\multirow{2}{*}{\textbf{Model}}
  & \multicolumn{3}{c}{\textbf{$\sigma$ on Random}}
  & \multicolumn{3}{c}{\textbf{$\sigma$ on Shuffling}}
  & \multicolumn{3}{c}{\textbf{$\sigma$ on RealWorld}} \\
\cmidrule(lr){2-4} \cmidrule(lr){5-7} \cmidrule(lr){8-10}
  & \textbf{KS@100} & \textbf{JSD} & \textbf{WZD}
  & \textbf{KS@100} & \textbf{JSD} & \textbf{WZD}
  & \textbf{KS@100} & \textbf{JSD} & \textbf{WZD} \\
\midrule

\texttt{Nemotron-3-super-120B-a12b}  & 0.31 & 0.80 & 3.9 & 0.33 & 0.88 & 4.1 & 0.35 & 0.96 & 4.3 \\
\texttt{GPT-5.4}                     & 0.25 & 1.29 & 2.7 & 0.27 & 1.39 & 2.8 & 0.29 & 1.49 & 3.0 \\
\texttt{GPT-4o}                      & 0.27 & 1.11 & 2.5 & 0.29 & 1.21 & 2.6 & 0.30 & 1.31 & 2.8 \\
\texttt{Llama-3.1-70B-instruct}      & 0.26 & 0.74 & 1.7 & 0.28 & 0.82 & 1.8 & 0.30 & 0.89 & 1.9 \\
\texttt{DeepSeek-V3.2}               & 0.28 & 0.71 & 1.7 & 0.30 & 0.79 & 1.8 & 0.32 & 0.86 & 1.9 \\
\texttt{Qwen3-32B}                   & 0.28 & 1.18 & 2.4 & 0.30 & 1.28 & 2.6 & 0.32 & 1.38 & 2.7 \\
\texttt{Claude-sonnet-4.6}           & 0.27 & 1.75 & 2.6 & 0.29 & 1.88 & 2.8 & 0.31 & 2.00 & 2.9 \\
\texttt{Mistral-Large}               & 0.29 & 1.20 & 2.7 & 0.31 & 1.31 & 2.8 & 0.33 & 1.41 & 3.0 \\
\texttt{GPT-4o-mini}                 & 0.30 & 1.36 & 2.6 & 0.32 & 1.47 & 2.8 & 0.34 & 1.58 & 3.0 \\
\texttt{Llama-3.1-8B-instruct}       & 0.30 & 0.86 & 1.9 & 0.32 & 0.94 & 2.0 & 0.34 & 1.02 & 2.1 \\
\texttt{Grok-4.1-fast}               & 0.30 & 1.02 & 2.8 & 0.32 & 1.12 & 2.9 & 0.34 & 1.22 & 3.1 \\
\texttt{Qwen-3.5-2B}                 & 0.31 & 0.82 & 1.8 & 0.33 & 0.90 & 1.9 & 0.35 & 0.98 & 2.0 \\
\texttt{Nemotron-3-nano-30B}         & 0.33 & 0.92 & 2.0 & 0.35 & 1.01 & 2.1 & 0.37 & 1.09 & 2.3 \\
\texttt{Mercury-2}                   & 0.32 & 1.01 & 2.1 & 0.34 & 1.10 & 2.2 & 0.35 & 1.18 & 2.4 \\
\texttt{Qwen-3.5-397B-a17b}          & 0.34 & 1.82 & 3.2 & 0.36 & 1.96 & 3.4 & 0.38 & 2.10 & 3.6 \\
\texttt{Phi-3.5-mini-instruct}       & 0.34 & 1.58 & 3.0 & 0.36 & 1.71 & 3.2 & 0.38 & 1.84 & 3.4 \\
\texttt{Llama-3.2-1B-instruct}       & 0.35 & 1.08 & 2.3 & 0.37 & 1.17 & 2.4 & 0.39 & 1.26 & 2.6 \\
\texttt{OLMo-3-7B-instruct}          & 0.36 & 1.42 & 3.3 & 0.38 & 1.55 & 3.5 & 0.40 & 1.68 & 3.7 \\
\texttt{Ministral-3B-instruct}       & 0.37 & 0.63 & 4.1 & 0.39 & 0.70 & 4.3 & 0.41 & 0.77 & 4.5 \\
\texttt{Qwen-3.5-35B-a3b}            & 0.38 & 2.25 & 3.6 & 0.40 & 2.40 & 3.8 & 0.42 & 2.55 & 4.0 \\

\bottomrule
\end{tabular}%
}
\end{table}
\section{Instruction Following Analysis of the Models}
\label{app:retry}

A generation \emph{fails} when its output cannot be parsed into a valid sample, for example a malformed sequence, an empty completion, a refusal, or a value outside the admissible support. We discard failed generations and resample the same prompt up to \textbf{5 retries}, retaining only valid samples for metric computation, so \textbf{retries do not bias the distributional metrics} and instead measure how reliably a model emits well-formed outputs. Table~\ref{tab:retry-rates} reports the mean attempts per valid sample (\emph{Avg.\ Attempt}) and the fraction of calls needing at least one resample (\emph{Retry Rate}). \textbf{Retry cost concentrates in \textsc{RealWorld}} and follows a \textbf{clear inverse-scaling trend}: \texttt{Llama-3.2-1B} retries on \textbf{68.6\%} of calls (4.28 attempts each) and \texttt{Qwen3.5-2B} on 29.9\%, while the strongest models stay near zero (\texttt{GPT-5.4} 0.0\%, \texttt{Qwen3.5-397B} 0.2\%). \textsc{Shuffling} is effectively retry-free (max $2.05\%$, \texttt{Mercury-2}), as its format is simple enough that validity is rarely the bottleneck. In \textsc{Text and Code} the trend reverses: the highest rates belong to two capable models, the diffusion-based \texttt{Mercury-2} ($13.6\%$) and \texttt{Claude-sonnet-4.6} ($10.9\%$, and the highest average attempt count overall at $1.406$), indicating that \textbf{these retries reflect format-adherence quirks, not capability}, with \texttt{Mercury-2} again an outlier across all three categories.

\begin{table}[t]
  \centering
  \setlength{\tabcolsep}{5pt}
  \caption{Retry behaviour across the three \textsc{\name{}} categories. \emph{Avg. Attempt} is the mean number of generation attempts.}
  \label{tab:retry-rates}
  \resizebox{\textwidth}{!}{%
  \begin{tabular}{l S[table-format=1.3] S[table-format=2.2] S[table-format=1.3] S[table-format=2.2] S[table-format=1.3] S[table-format=2.2]}
    \toprule
    & \multicolumn{2}{c}{\textbf{Text and Code}} & \multicolumn{2}{c}{\textbf{Shuffling}} & \multicolumn{2}{c}{\textbf{RealWorld}} \\
    \cmidrule(lr){2-3}\cmidrule(lr){4-5}\cmidrule(lr){6-7}
    \textbf{Model} & {Avg. Attempt} & {Retry Rate(\%)} & {Avg. Attempt} & {Retry Rate(\%)} & {Avg. Attempt} & {Retry Rate(\%)} \\
    \midrule
    \texttt{Llama-3.2-1B-instruct} & 1.021 & 1.94 & 1.012 & 0.90 & 4.276 & 68.60 \\
    \texttt{Qwen3.5-2B} & 1.010 & 0.90 & 1.016 & 1.55 & 1.986 & 29.93 \\
    \texttt{Mercury-2} & 1.208 & 13.56 & 1.021 & 2.05 & 1.187 & 14.60 \\
    \texttt{Ministral-3B-instruct} & 1.014 & 1.12 & 1.005 & 0.50 & 1.684 & 26.67 \\
    \texttt{Phi-3.5-mini-instruct} & 1.083 & 4.81 & 1.000 & 0.00 & 1.680 & 18.27 \\
    \texttt{Nemotron-3-nano-30B} & 1.017 & 1.65 & 1.011 & 1.10 & 1.284 & 16.77 \\
    \texttt{Llama-3.1-8B-instruct} & 1.008 & 0.68 & 1.000 & 0.00 & 1.561 & 18.20 \\
    \texttt{OLMo-3-7B-instruct} & 1.104 & 6.00 & 1.000 & 0.00 & 1.169 & 8.13 \\
    \texttt{Claude-sonnet-4.6} & 1.406 & 10.91 & 1.000 & 0.00 & 1.014 & 0.90 \\
    \texttt{Qwen3-32B} & 1.000 & 0.05 & 1.000 & 0.00 & 1.188 & 9.90 \\
    \texttt{Qwen3.5-35B-a3b} & 1.028 & 0.81 & 1.000 & 0.00 & 1.334 & 6.73 \\
    \texttt{Llama-3.1-70B-instruct} & 1.007 & 0.56 & 1.000 & 0.00 & 1.064 & 3.93 \\
    \texttt{Grok-4.1-fast} & 1.004 & 0.22 & 1.000 & 0.00 & 1.127 & 3.60 \\
    \texttt{Nemotron-3-super-120B-a12b} & 1.014 & 1.22 & 1.000 & 0.00 & 1.016 & 1.47 \\
    \texttt{GPT-4o-mini} & 1.002 & 0.18 & 1.001 & 0.10 & 1.026 & 1.93 \\
    \texttt{DeepSeek-V3.2} & 1.002 & 0.16 & 1.000 & 0.00 & 1.024 & 1.73 \\
    \texttt{GPT-4o} & 1.000 & 0.05 & 1.000 & 0.05 & 1.007 & 0.63 \\
    \texttt{Mistral-large-2512} & 1.000 & 0.00 & 1.000 & 0.00 & 1.005 & 0.47 \\
    \texttt{Qwen3.5-397B-a17b} & 1.004 & 0.20 & 1.000 & 0.00 & 1.002 & 0.20 \\
    \texttt{GPT-5.4} & 1.002 & 0.14 & 1.000 & 0.00 & 1.000 & 0.00 \\
    \bottomrule
  \end{tabular}
  }
\end{table}
 
\section{Output Diversity Analysis}
\label{app:uniqueness}

\begin{figure}[t]
    \centering
    \includegraphics[width=\linewidth]{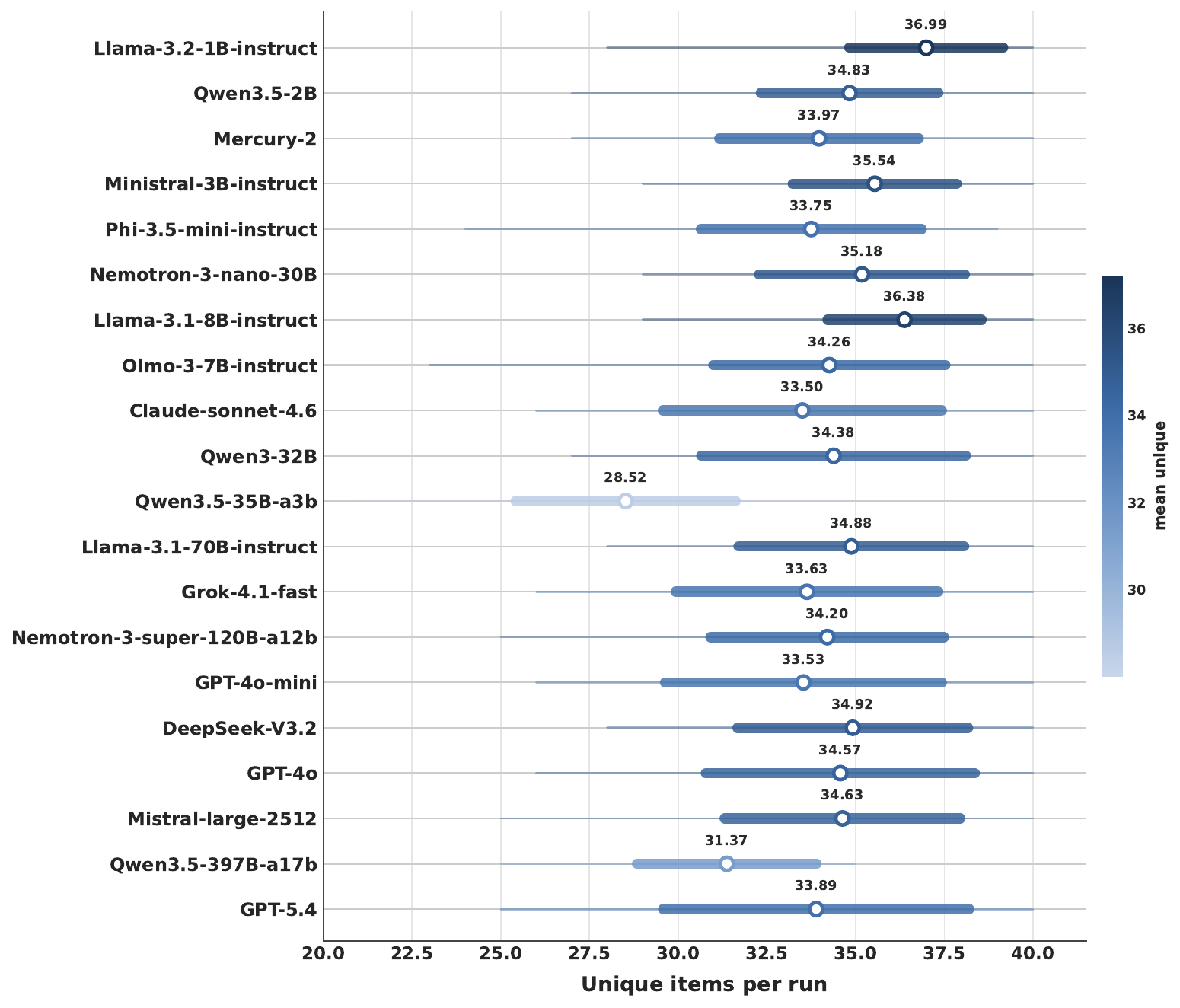}
    \caption{Per-run output diversity on the shuffling task, measured as the number of unique items produced out of the $\approx$40 items presented per run, aggregated over 1000 runs at temperature $1.0$. For each model we show the mean (marker), the $\pm1$ standard-deviation band (thick bar), and the full observed min--max range (thin line); color encodes the mean. The dashed line marks the attainable ceiling ($\approx$39.8 items per run).}
    \label{fig:diversity}
\end{figure}

Figure~\ref{fig:diversity} reports the per-run diversity of each model on the shuffling task, measured as the number of unique items produced out of the $\approx$40 items presented per run, aggregated over 1000 runs at temperature $1.0$. For each model we plot the mean (marker), the $\pm1$ standard-deviation band (thick bar), and the full observed min--max range (thin line), with color encoding the mean for legibility. All models cluster well below the attainable ceiling of $\approx$39.8, indicating that none reproduces the full uniform spread expected under ideal sampling. Notably, diversity does not increase with scale: the highest mean unique counts come from the smallest instruct models, \texttt{Llama-3.2-1B-instruct} ($36.99$) and \texttt{Llama-3.1-8B-instruct} ($36.38$), while the two largest mixture-of-experts models, \texttt{Qwen3.5-397B-a17b} ($31.37$) and \texttt{Qwen3.5-35B-a3b} ($28.52$), are theleast diverse and are the only models whose maximum never exceeds 35, suggesting a systematically collapsed output distribution rather than occasional low-diversity runs. The remaining models occupy a comparatively narrow band of means ($33.5$--$35.5$), and we observe that lower-diversity models tend to exhibit both higher variance and longer left tails (e.g.\ \texttt{GPT-5.4} and \texttt{Claude-sonnet-4.6} reach as few as 25--26 unique items in their worst runs), implying that diversity loss is driven primarily by intermittent mode collapse rather than a uniform downward shift.
\section{Additional Qualitative Analysis}
\label{app:additional_qualitative}

\begin{figure}[ht]
    \caption{Model sample distributions vs.\ ground truth for the \textbf{Fréchet 
    Distribution} under the \textit{textual explicit concentrated} task setting. Each 
    subplot shows the empirical density of 100 model-generated samples (blue bars) 
    overlaid with the ground-truth distribution (red curve). Most models fail to 
    capture the heavy right tail of the Fréchet distribution, collapsing near the lower 
    bound of the support.}
    \label{fig:grid1}
    \centering
    \includegraphics[width=\linewidth]{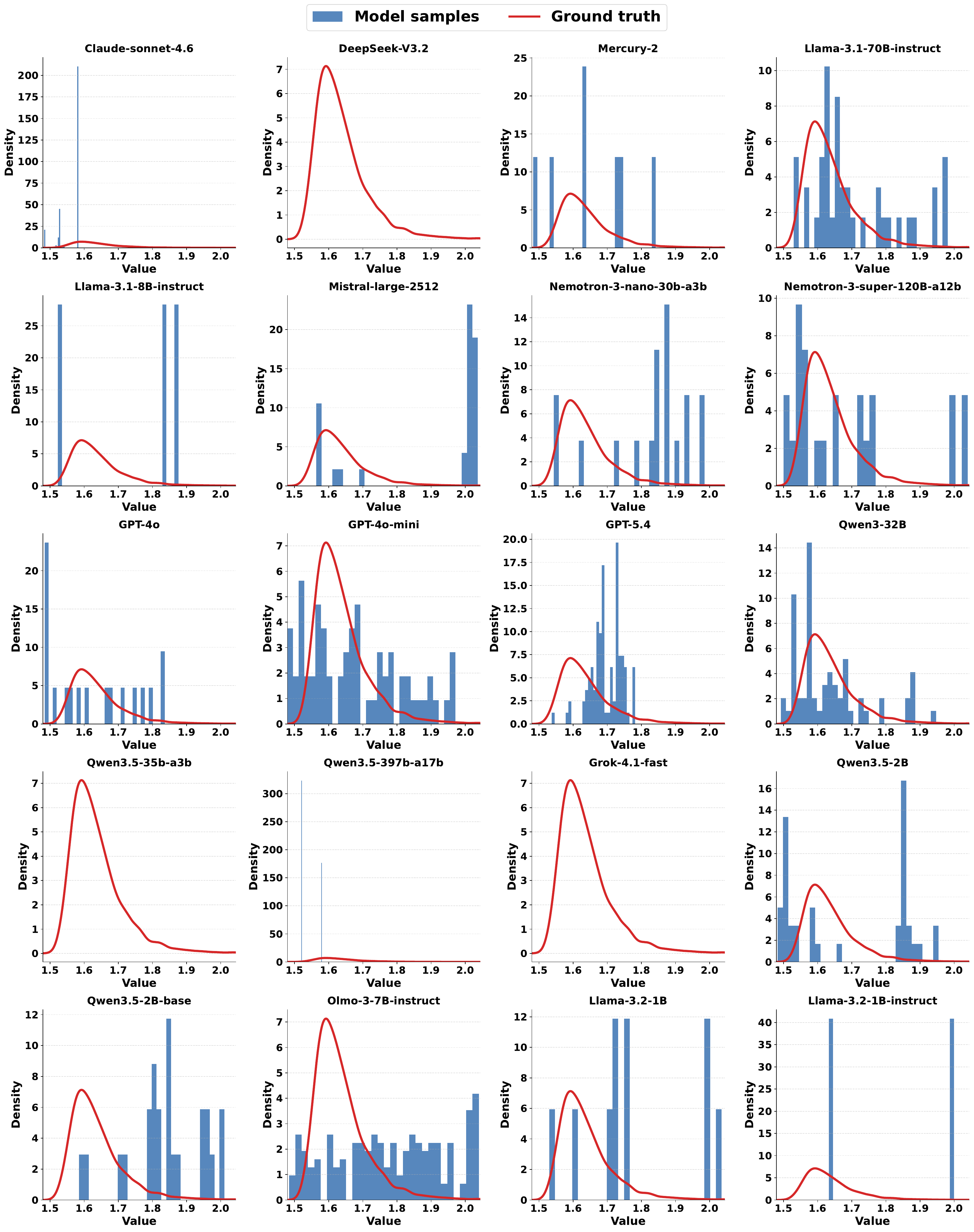}
\end{figure}

\begin{figure}[ht]
    \caption{Model sample distributions vs.\ ground truth for the \textbf{Truncated 
    Normal Distribution} under the \textit{textual explicit spread out} task setting. 
    The spread out parameterization results in a broad bell-shaped ground truth. While 
    some models approximate the shape reasonably, others collapse to a single point 
    despite the wide support.}
    \label{fig:grid2}
    \centering
    \includegraphics[width=\linewidth]{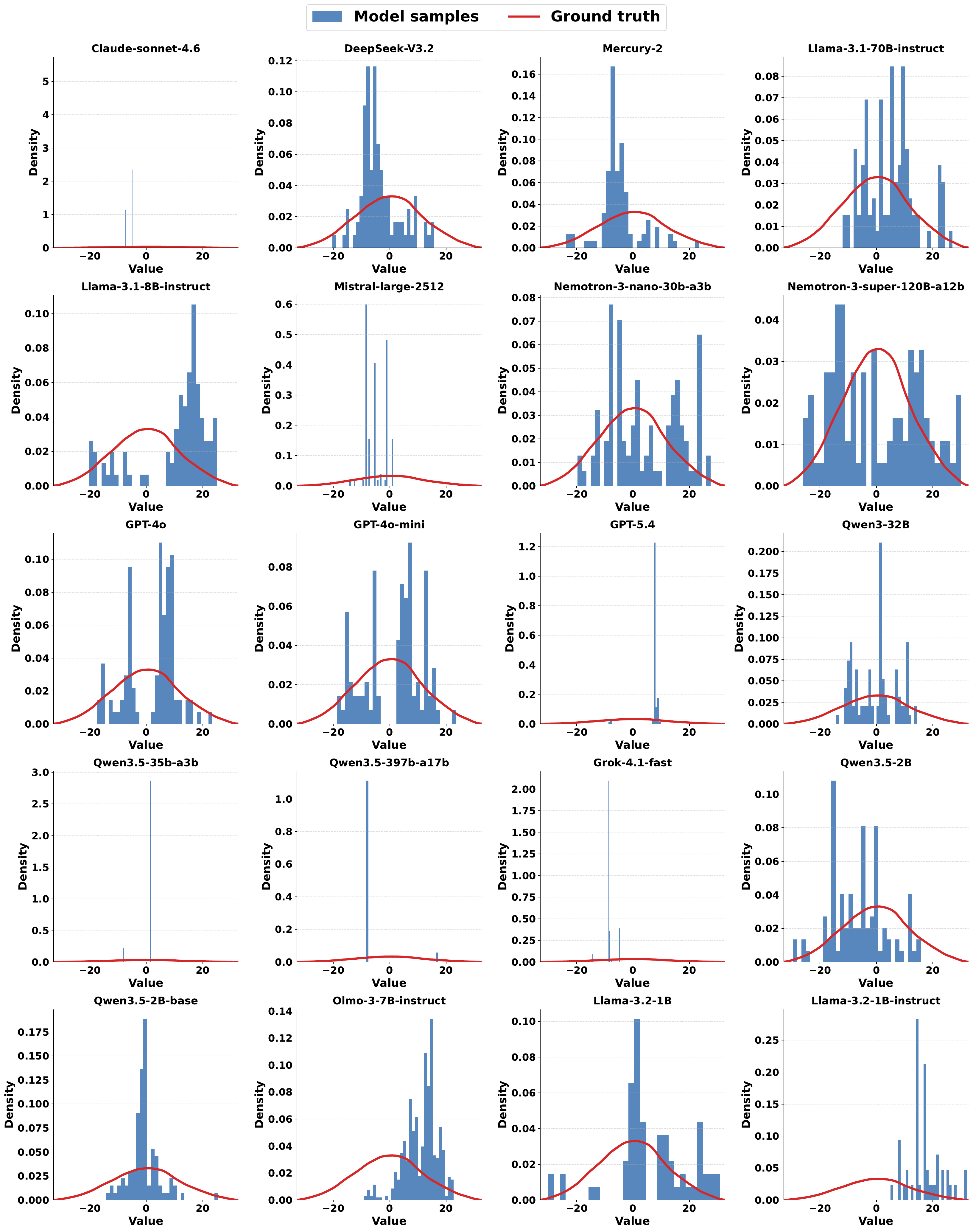}
\end{figure}

\begin{figure}[ht]
    \caption{Model sample distributions vs.\ ground truth for the \textbf{Binomial 
    Distribution (n=6, p=0.9)} under the \textit{textual implicit concentrated} task setting. The 
    distribution name is not stated in the prompt; models must infer the distributional 
    structure from context. The discrete, concentrated support makes this task 
    deceptively difficult, as models must both identify the correct distribution and 
    match its probability mass across a small integer range.}
    \label{fig:grid3}
    \centering
    \includegraphics[width=\linewidth]{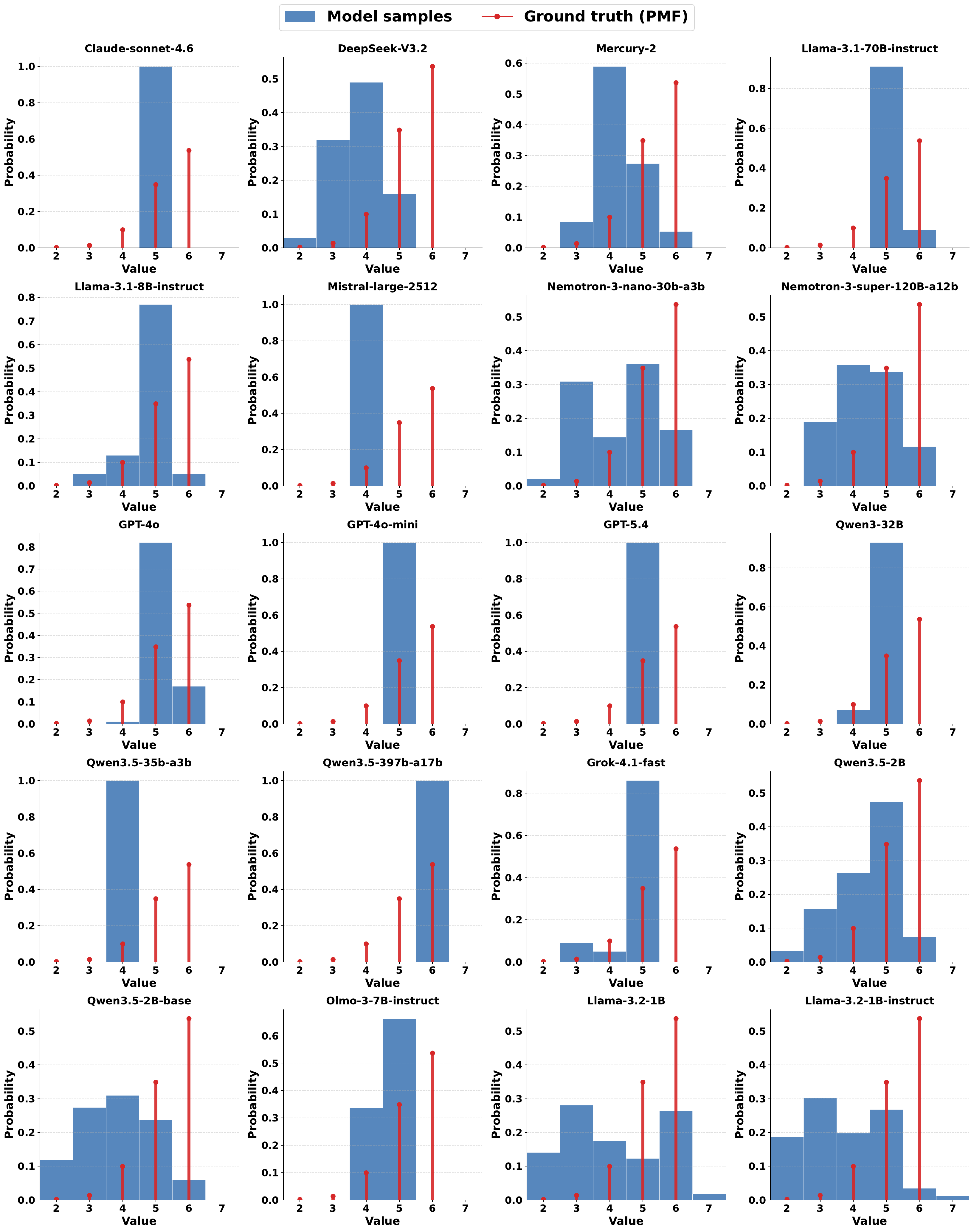}
\end{figure}

\begin{figure}[ht]
    \caption{Model sample distributions vs.\ ground truth for the \textbf{Poisson 
    Distribution} under the \textit{code implicit concentrated} task setting. Models 
    must infer the Poisson sampling process from a code snippet without an explicit 
    distribution name. The integer-valued, right-skewed ground truth exposes a clear 
    divide between models that can interpret stochastic code and those that collapse 
    to zero or a single small integer.}
    \label{fig:grid4}
    \centering
    \includegraphics[width=\linewidth]{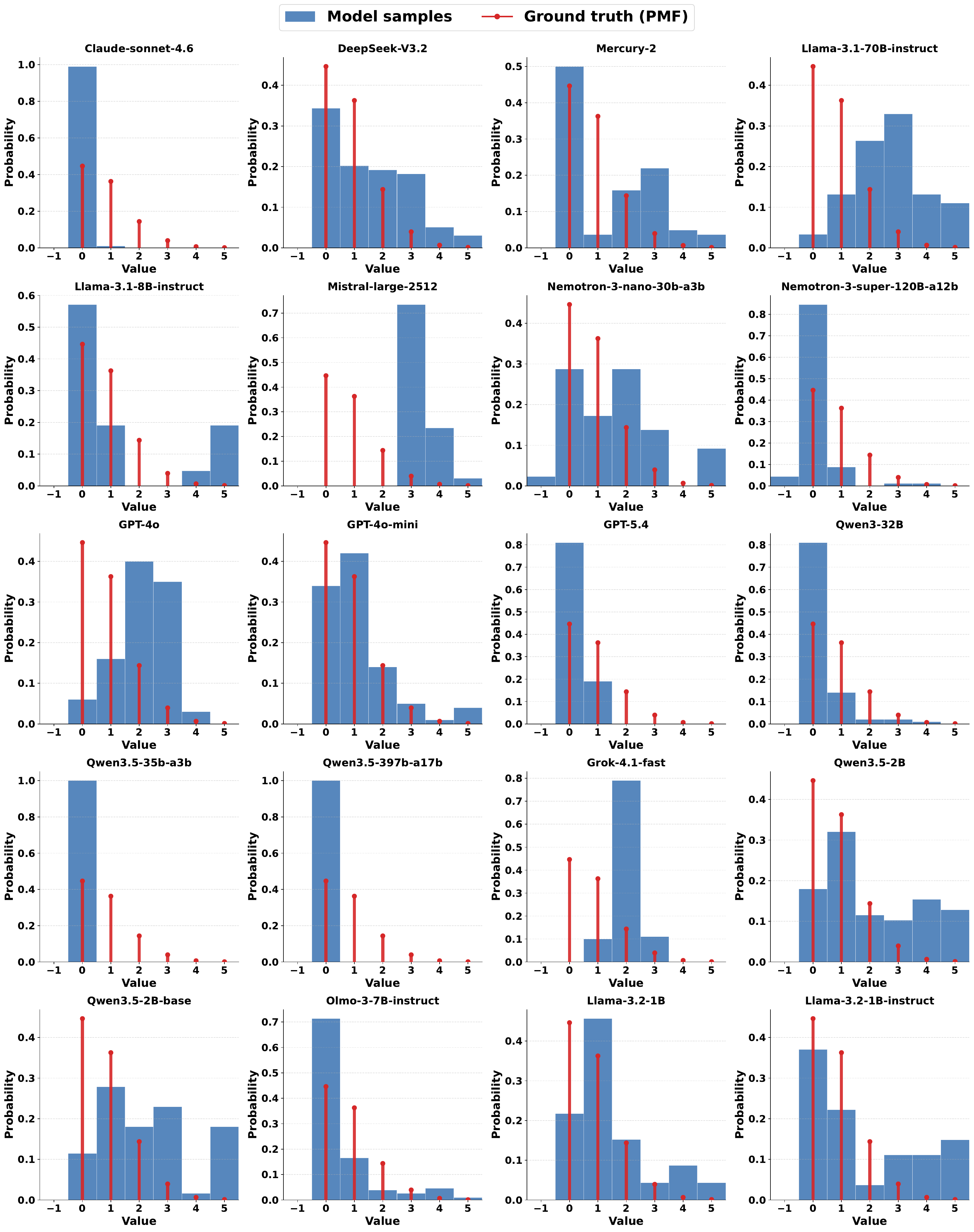}
\end{figure}

Figures~\ref{fig:grid1}--\ref{fig:grid4} show the empirical density of model generated samples (blue bars) against the ground-truth distribution (red curve) across all evaluated models for four representative tasks. These plots extend the qualitative observations from Section~\ref{sec:qualitative} to the full model pool and across different task types and distributions.

A consistent pattern emerges across all four figures: \textbf{most models either collapse to a narrow spike or concentrate mass at a single point}, failing to reproduce the shape of the ground truth. This is most dramatically visible for Claude Sonnet 4.6, GPT-5.4, Qwen-3.5-397B-a17b, Qwen-3.5-35B-a3b, and Mistral Large, which in multiple figures produce near-degenerate distributions with virtually all mass at one value. The Fréchet distribution (Figure~\ref{fig:grid1}) is particularly revealing: it has a heavy right tail that almost no model captures, with most collapsing to values near the lower bound of the support. The Truncated Normal (Figure~\ref{fig:grid2}) is one of the more tractable distributions, and here we observe the widest spread of model behaviors: some models like DeepSeek V3.2 and Llama-3.1-70B approximate the bell shape reasonably, while others such as Claude Sonnet 4.6 and GPT-5.4 again collapse to a single point. On the implicit binomial task (Figure~\ref{fig:grid3}), where the distribution name is not stated and must be inferred from context, models generally struggle more: even models that perform reasonably on explicit tasks show increased variance and misalignment here. Finally, the implicit code-based Poisson task (Figure~\ref{fig:grid4}) exposes a clear divide: models that can interpret the stochastic code produce outputs roughly consistent with the Poisson shape, while weaker models collapse entirely to zero or a single small integer. \textbf{Across all four figures, Nemotron-3 Super 120B and DeepSeek V3.2 consistently produce the broadest and most ground-truth-aligned distributions}, while models optimized for deterministic 
reasoning show the most severe collapse.

\section{Prompts}
\label{app:prompts}

\paragraph{Text Task Generation.}
Text-based tasks are generated using four prompts depending on whether the task is explicit or implicit and whether the target distribution is concentrated or spread out. The explicit concentrated and spread out variants are given in Prompt\hyperref[prompt:text-explicit-c]{~\ref*{prompt:text-explicit-c}} and Prompt\hyperref[prompt:text-explicit-s]{~\ref*{prompt:text-explicit-s}}, and the implicit concentrated and spread out variants in Prompt\hyperref[prompt:text-implicit-c]{~\ref*{prompt:text-implicit-c}} and Prompt\hyperref[prompt:text-implicit-s]{~\ref*{prompt:text-implicit-s}}. The prompt used to elicit a single sampled value from models at evaluation time is given in Prompt\hyperref[prompt:text-sampling]{~\ref*{prompt:text-sampling}}.

\paragraph{Code Task Generation.}

Code-based tasks follow the same explicit/implicit and concentrated/spread out structure as text tasks. The four generation prompts are given in Prompt\hyperref[prompt:code-explicit-c]{~\ref*{prompt:code-explicit-c}}, Prompt\hyperref[prompt:code-explicit-s]{~\ref*{prompt:code-explicit-s}}, Prompt\hyperref[prompt:code-implicit-c]{~\ref*{prompt:code-implicit-c}}, and Prompt\hyperref[prompt:code-implicit-s]{~\ref*{prompt:code-implicit-s}}. The sampling prompt used at evaluation time is given in Prompt\hyperref[prompt:code-sampling]{~\ref*{prompt:code-sampling}}.

\paragraph{Multimodal Task Generation.}
Multimodal tasks, which require models to sample from mixture distributions, are generated using two prompts corresponding to concentrated and spread out parameter regimes, given in Prompt\hyperref[prompt:multimodal-c]{~\ref*{prompt:multimodal-c}} and Prompt\hyperref[prompt:multimodal-s]{~\ref*{prompt:multimodal-s}}.

\paragraph{Answer Extraction.}
Model outputs are parsed using a family of LLM-based answer extractors tailored to each task type. The extractors for standard text and code tasks, list output tasks, shuffling tasks, and real-world tasks are given in Prompt\hyperref[prompt:extractor]{~\ref*{prompt:extractor}}, Prompt\hyperref[prompt:extractor_list]{~\ref*{prompt:extractor_list}}, Prompt\hyperref[prompt:extractor_shuffling]{~\ref*{prompt:extractor_shuffling}}, and Prompt\hyperref[prompt:extractor_realworld]{~\ref*{prompt:extractor_realworld}} respectively.

\clearpage

\begin{promptbox}{Text Explicit Task Generation Prompt (Concentrated)}
\label{prompt:text-explicit-c}
Developer: You are an expert statistician and data scientist. Your goal is to assess an LLM's ability to understand and generate random samples from probability distributions.

**Inputs:**

- Name of the distribution (string): \{distribution\_name\}

- Details and properties of the distribution, extracted from Wikipedia (string): \{distribution\_wikipedia\}.

For each request:

- Generate a clear, explicit sampling question based on the provided distribution data.

- Select random but valid values for the distribution parameters that are reasonable given the distribution type.

- Prefer parameter values that make the distribution's support or typical sample range concentrated as much as possible (least spread out), when this is feasible and still valid for the specified distribution.

- Ensure the question is both relevant and solvable using properties of the specified probability distribution.

- **Do not** ask about mean, median, or other typical statistics. Focus on generating questions that require drawing random samples from the distribution.

If any required input fields are missing, incomplete, or malformed (e.g., `distribution\_name` is empty/invalid, `distribution\_wikipedia` is not about a probability distribution, or parameters cannot be inferred), immediately return a JSON object with an `error` field describing the issue. Do not proceed further in such cases.

After producing the output, validate that all output fields are present and contextually appropriate for the selected distribution; if any required fields are missing or inappropriate, return only a JSON object with an `error` field describing the issue instead of partial results.

**Expected Output Structure:**

- On success, return a JSON object containing:

  - `question` (string): A sampling question built for the specific distribution.
  
  - `parameters` (object): Dictionary of explicitly chosen, valid distribution parameters.
  
  - `context` (string): Brief scenario or use-case relevant to the distribution.
  
  - `expected\_answer\_type` (string): Strictly a single numerical value.
  
  - `code\_snippet` (string): Valid Python code (NumPy/SciPy or similar) to solve the sampling task.

- Before returning, validate that all output fields are present and their content is contextually appropriate for the distribution. If one or more fields are missing, inappropriate, or cannot be generated, respond only with an `error` field in a JSON object describing the problem.

**Example Success Output:**

\{\{Example\}\}

**Example Error Output:**

\{\{Example\}\}
\end{promptbox}

\begin{promptbox}{Text Explicit Task Generation Prompt (Spread-out)}
\label{prompt:text-explicit-s}
Developer: You are an expert statistician and data scientist. Your goal is to assess an LLM's ability to understand and generate random samples from probability distributions.

**Inputs:**

- Name of the distribution (string): \{distribution\_name\}

- Details and properties of the distribution, extracted from Wikipedia (string): \{distribution\_wikipedia\}.

For each request:

- Generate a clear, explicit sampling question based on the provided distribution data.

- Select random but valid values for the distribution parameters that are reasonable given the distribution type.

- Prefer parameter values that make the distribution's support or typical sample range spread out as much as possible (least concentrated), when this is feasible and still valid for the specified distribution.

- Ensure the question is both relevant and solvable using properties of the specified probability distribution.

- **Do not** ask about mean, median, or other typical statistics. Focus on generating questions that require drawing random samples from the distribution.

If any required input fields are missing, incomplete, or malformed (e.g., `distribution\_name` is empty/invalid, `distribution\_wikipedia` is not about a probability distribution, or parameters cannot be inferred), immediately return a JSON object with an `error` field describing the issue. Do not proceed further in such cases.

After producing the output, validate that all output fields are present and contextually appropriate for the selected distribution; if any required fields are missing or inappropriate, return only a JSON object with an `error` field describing the issue instead of partial results.

**Expected Output Structure:**

- On success, return a JSON object containing:

  - `question` (string): A sampling question built for the specific distribution.
  
  - `parameters` (object): Dictionary of explicitly chosen, valid distribution parameters.
  
  - `context` (string): Brief scenario or use-case relevant to the distribution.
  
  - `expected\_answer\_type` (string): Strictly a single numerical value.
  
  - `code\_snippet` (string): Valid Python code (NumPy/SciPy or similar) to solve the sampling task.

- Before returning, validate that all output fields are present and their content is contextually appropriate for the distribution. If one or more fields are missing, inappropriate, or cannot be generated, respond only with an `error` field in a JSON object describing the problem.

**Example Success Output:**

\{\{Example\}\}

**Example Error Output:**

\{\{Example\}\}
\end{promptbox}

\begin{promptbox}{Text Implicit Task Generation Prompt (Concentrated)}
\label{prompt:text-implicit-c}
Developer: You are an expert statistician and data scientist. Your goal is to assess an LLM's ability to understand and generate random samples from probability distributions.

**Inputs:**

- Name of the distribution (string): {distribution\_name}

- Details and properties of the distribution, extracted from Wikipedia (string): {distribution\_wikipedia}.

For each request:

- Generate a clear, implicit sampling question based on the provided distribution data.

- The question must be implicit and self-contained, but it must **not** mention the actual distribution name in the question itself.

- Write the question in an indirect/applied way, similar to the example below, so that it implies the distribution through the scenario or sampling process rather than naming it directly.

- Select random but valid values for the distribution parameters that are reasonable given the distribution type.

- Prefer parameter values that make the distribution's support or typical sample range concentrated as much as possible (least spread out), when this is feasible and still valid for the specified distribution.

- Ensure the question is both relevant and solvable using properties of the specified probability distribution.

- **Do not** ask about mean, median, or other typical statistics. Focus on generating questions that require drawing random samples from the distribution.

If any required input fields are missing, incomplete, or malformed (e.g., `distribution\_name` is empty/invalid, `distribution\_wikipedia` is not about a probability distribution, or parameters cannot be inferred), immediately return a JSON object with an `error` field describing the issue. Do not proceed further in such cases.

After producing the output, validate that all output fields are present and contextually appropriate for the selected distribution; if any required fields are missing or inappropriate, return only a JSON object with an `error` field describing the issue instead of partial results.

**Expected Output Structure:**

- On success, return a JSON object containing:

  - `question` (string): A sampling question built for the specific distribution.
  
  - `parameters` (object): Dictionary of explicitly chosen, valid distribution parameters.
  
  - `context` (string): Brief scenario or use-case relevant to the distribution.
  
  - `expected\_answer\_type` (string): Use `numerical\_value` to indicate that the answer should be a single numerical value.
  - `code\_snippet` (string): Valid Python code (NumPy/SciPy or similar) to solve the sampling task.

- Before returning, validate that all output fields are present and their content is contextually appropriate for the distribution. If one or more fields are missing, inappropriate, or cannot be generated, respond only with an `error` field in a JSON object describing the problem.

**Example Success Output:**

\{\{Example\}\}

**Example Error Output:**

\{\{Example\}\}
\end{promptbox}

\begin{promptbox}{Text Implicit Task Generation Prompt (Spread-out)}
\label{prompt:text-implicit-s}
Developer: You are an expert statistician and data scientist. Your goal is to assess an LLM's ability to understand and generate random samples from probability distributions.

**Inputs:**

- Name of the distribution (string): {distribution\_name}

- Details and properties of the distribution, extracted from Wikipedia (string): {distribution\_wikipedia}.

For each request:

- Generate a clear, implicit sampling question based on the provided distribution data.

- The question must be implicit and self-contained, but it must **not** mention the actual distribution name in the question itself.

- Write the question in an indirect/applied way, similar to the example below, so that it implies the distribution through the scenario or sampling process rather than naming it directly.

- Select random but valid values for the distribution parameters that are reasonable given the distribution type.

- Prefer parameter values that make the distribution's support or typical sample range spread out as much as possible (least concentrated), when this is feasible and still valid for the specified distribution.

- Ensure the question is both relevant and solvable using properties of the specified probability distribution.

- **Do not** ask about mean, median, or other typical statistics. Focus on generating questions that require drawing random samples from the distribution.

If any required input fields are missing, incomplete, or malformed (e.g., `distribution\_name` is empty/invalid, `distribution\_wikipedia` is not about a probability distribution, or parameters cannot be inferred), immediately return a JSON object with an `error` field describing the issue. Do not proceed further in such cases.

After producing the output, validate that all output fields are present and contextually appropriate for the selected distribution; if any required fields are missing or inappropriate, return only a JSON object with an `error` field describing the issue instead of partial results.

**Expected Output Structure:**

- On success, return a JSON object containing:

  - `question` (string): A sampling question built for the specific distribution.
  
  - `parameters` (object): Dictionary of explicitly chosen, valid distribution parameters.
  
  - `context` (string): Brief scenario or use-case relevant to the distribution.
  
  - `expected\_answer\_type` (string): Use `numerical\_value` to indicate that the answer should be a single numerical value.
  - `code\_snippet` (string): Valid Python code (NumPy/SciPy or similar) to solve the sampling task.

- Before returning, validate that all output fields are present and their content is contextually appropriate for the distribution. If one or more fields are missing, inappropriate, or cannot be generated, respond only with an `error` field in a JSON object describing the problem.

**Example Success Output:**

\{\{Example\}\}

**Example Error Output:**

\{\{Example\}\}
\end{promptbox}

\begin{promptbox}{Code Explicit Task Generation Prompt (Concentrated)}
\label{prompt:code-explicit-c}
Developer: You are an expert statistician and data scientist. Your goal is to assess an LLM's ability to understand and generate random samples from probability distributions.

**Inputs:**

- Name of the distribution (string): {distribution\_name}

- Details and properties of the distribution, extracted from Wikipedia (string): {distribution\_wikipedia}.

For each request:

- Generate a clear, explicit sampling Python code based on the provided distribution data.

- Select random but valid values for the distribution parameters that are reasonable given the distribution type.

- Prefer parameter values that make the distribution's support or typical sample range concentrated as much as possible (least spread out), when this is feasible and still valid for the specified distribution.

- Ensure the code is both relevant and runnable using properties of the specified probability distribution.

- Strictly avoid queries or code about mean, median, or other descriptive statistics; focus exclusively on random sampling procedures.

If any required input fields are missing, incomplete, or malformed (e.g., `distribution\_name` is empty/invalid, `distribution\_wikipedia` is not about a probability distribution, or parameters cannot be inferred), immediately return a JSON object with an `error` field describing the issue. Do not proceed further in such cases.

After producing the output, validate that all output fields are present and contextually appropriate for the selected distribution; if any required fields are missing or inappropriate, return only a JSON object with an `error` field describing the issue instead of partial results.

**Expected Output Structure:**

- On success, return a JSON object containing:

  - `code\_snippet` (string): A valid Python code (NumPy/SciPy or similar) built for the specific distribution.
  
  - `parameters` (object): Dictionary of explicitly chosen, valid distribution parameters.
  
  - `context` (string): Brief scenario or use-case relevant to the distribution.
  
  - `expected\_answer\_type` (string): Strictly a single numerical value.

- Before returning, validate that all output fields are present and their content is contextually appropriate for the distribution. If one or more fields are missing, inappropriate, or cannot be generated, respond only with an `error` field in a JSON object describing the problem.

**Example Success Output:**

\{\{Example\}\}

**Example Error Output:**

\{\{Example\}\}
\end{promptbox}

\begin{promptbox}{Code Explicit Task Generation Prompt (Spread-out)}
\label{prompt:code-explicit-s}
Developer: You are an expert statistician and data scientist. Your goal is to assess an LLM's ability to understand and generate random samples from probability distributions.

**Inputs:**

- Name of the distribution (string): {distribution\_name}

- Details and properties of the distribution, extracted from Wikipedia (string): {distribution\_wikipedia}.

For each request:

- Generate a clear, explicit sampling Python code based on the provided distribution data.

- Select random but valid values for the distribution parameters that are reasonable given the distribution type.

- Prefer parameter values that make the distribution's support or typical sample range spread out as much as possible (least concentrated), when this is feasible and still valid for the specified distribution.

- Ensure the code is both relevant and runnable using properties of the specified probability distribution.

- Strictly avoid queries or code about mean, median, or other descriptive statistics; focus exclusively on random sampling procedures.

If any required input fields are missing, incomplete, or malformed (e.g., `distribution\_name` is empty/invalid, `distribution\_wikipedia` is not about a probability distribution, or parameters cannot be inferred), immediately return a JSON object with an `error` field describing the issue. Do not proceed further in such cases.

After producing the output, validate that all output fields are present and contextually appropriate for the selected distribution; if any required fields are missing or inappropriate, return only a JSON object with an `error` field describing the issue instead of partial results.

**Expected Output Structure:**

- On success, return a JSON object containing:

  - `code\_snippet` (string): A valid Python code (NumPy/SciPy or similar) built for the specific distribution.
  
  - `parameters` (object): Dictionary of explicitly chosen, valid distribution parameters.
  
  - `context` (string): Brief scenario or use-case relevant to the distribution.
  
  - `expected\_answer\_type` (string): Strictly a single numerical value.

- Before returning, validate that all output fields are present and their content is contextually appropriate for the distribution. If one or more fields are missing, inappropriate, or cannot be generated, respond only with an `error` field in a JSON object describing the problem.

**Example Success Output:**

\{\{Example\}\}

**Example Error Output:**

\{\{Example\}\}
\end{promptbox}

\begin{promptbox}{Code Implicit Task Generation Prompt (concentrated)}
\label{prompt:code-implicit-c}
Developer: You are an expert statistician and data scientist. Your goal is to assess an LLM's ability to understand and generate random samples from probability distributions.

**Inputs:**

- Name of the distribution (string): {distribution\_name}

- Details and properties of the distribution, extracted from Wikipedia (string): {distribution\_wikipedia}.

For each request:

- Generate a clear, implicit sampling Python code snippet based on the provided distribution data.

- The code must be self-contained and executable, but it must **not** explicitly mention the actual distribution name in variable names, comments, printed text, or explanatory text.

- Write the code in an indirect/applied way so that it implies the distribution through the scenario, transformation, or sampling procedure rather than naming it directly.

- Select random but valid values for the distribution parameters that are reasonable given the distribution type.

- Prefer parameter values that make the distribution's support or typical sample range concentrated as much as possible (least spread out), when this is feasible and still valid for the specified distribution.

- Ensure the code is both relevant and runnable using properties of the specified probability distribution.

- **Do not** generate code about mean, median, or other typical/descriptive statistics. Focus on generating code that performs random sampling and produces a single sampled numerical outcome.

If any required input fields are missing, incomplete, or malformed (e.g., `distribution\_name` is empty/invalid, `distribution\_wikipedia` is not about a probability distribution, or parameters cannot be inferred), immediately return a JSON object with an `error` field describing the issue. Do not proceed further in such cases.

After producing the output, validate that all output fields are present and contextually appropriate for the selected distribution; if any required fields are missing or inappropriate, return only a JSON object with an `error` field describing the issue instead of partial results.

**Expected Output Structure:**

- On success, return a JSON object containing:

  - `code\_snippet` (string): Valid Python code (NumPy/SciPy or similar) to perform the implicit sampling task and output a single numerical value.
  
  - `parameters` (object): Dictionary of explicitly chosen, valid distribution parameters.
  
  - `context` (string): Brief scenario or use-case relevant to the distribution.
  
  - `expected\_answer\_type` (string): Use `numerical\_value` to indicate that the answer should be a single numerical value.

Before returning, validate that all output fields are present and their content is contextually appropriate for the distribution. If one or more fields are missing, inappropriate, or cannot be generated, respond only with an `error` field in a JSON object describing the problem.

**Example Success Output:**

\{\{Example\}\}

**Example Error Output:**

\{\{Example\}\}
\end{promptbox}

\begin{promptbox}{Code Implicit Task Generation Prompt (Spread-out)}
\label{prompt:code-implicit-s}
Developer: You are an expert statistician and data scientist. Your goal is to assess an LLM's ability to understand and generate random samples from probability distributions.

**Inputs:**

- Name of the distribution (string): {distribution\_name}

- Details and properties of the distribution, extracted from Wikipedia (string): {distribution\_wikipedia}.

For each request:

- Generate a clear, implicit sampling Python code snippet based on the provided distribution data.

- The code must be self-contained and executable, but it must **not** explicitly mention the actual distribution name in variable names, comments, printed text, or explanatory text.

- Write the code in an indirect/applied way so that it implies the distribution through the scenario, transformation, or sampling procedure rather than naming it directly.

- Select random but valid values for the distribution parameters that are reasonable given the distribution type.

- Prefer parameter values that make the distribution's support or typical sample range spread out as much as possible (least concentrated), when this is feasible and still valid for the specified distribution.

- Ensure the code is both relevant and runnable using properties of the specified probability distribution.

- **Do not** generate code about mean, median, or other typical/descriptive statistics. Focus on generating code that performs random sampling and produces a single sampled numerical outcome.

If any required input fields are missing, incomplete, or malformed (e.g., `distribution\_name` is empty/invalid, `distribution\_wikipedia` is not about a probability distribution, or parameters cannot be inferred), immediately return a JSON object with an `error` field describing the issue. Do not proceed further in such cases.

After producing the output, validate that all output fields are present and contextually appropriate for the selected distribution; if any required fields are missing or inappropriate, return only a JSON object with an `error` field describing the issue instead of partial results.

**Expected Output Structure:**

- On success, return a JSON object containing:

  - `code\_snippet` (string): Valid Python code (NumPy/SciPy or similar) to perform the implicit sampling task and output a single numerical value.
  
  - `parameters` (object): Dictionary of explicitly chosen, valid distribution parameters.
  
  - `context` (string): Brief scenario or use-case relevant to the distribution.
  
  - `expected\_answer\_type` (string): Use `numerical\_value` to indicate that the answer should be a single numerical value.

Before returning, validate that all output fields are present and their content is contextually appropriate for the distribution. If one or more fields are missing, inappropriate, or cannot be generated, respond only with an `error` field in a JSON object describing the problem.

**Example Success Output:**

\{\{Example\}\}

**Example Error Output:**

\{\{Example\}\}
\end{promptbox}

\begin{promptbox}{Multimodal Task Generation Prompt (Concentrated)}
\label{prompt:multimodal-c}
Developer: You are an expert statistician and data scientist. Your goal is to assess an LLM's ability to understand and generate random samples from multimodal probability distributions.

**Inputs:**

- Name of the distribution (string): {distribution\_name}

- Details and properties of the distribution, extracted from Wikipedia (string): {distribution\_wikipedia}.

For each request:

- Determine if the provided distribution is multimodal based on the input details.

    - If it is already multimodal, proceed as usual.
    
    - If the provided distribution is unimodal (single modal), construct a multimodal (2-component) distribution by reasonable means (e.g., as a mixture or sum of distributions, or another mathematically valid transformation), and base your generated question on this multimodal version, specifying all relevant parameters.
    
- Generate a clear, explicit sampling question based on the (if necessary, constructed) multimodal distribution data.

- Select random but valid values for the distribution parameters that are reasonable given the distribution type or construction.

- Prefer parameter values that make the distribution's support or typical sample range concentrated as much as possible (least spread out), when this is feasible and still valid for the specified distribution.

- Ensure the question is both relevant and solvable using properties of the specified or constructed probability distribution.

- The question should be human readable and easy to understand. **Do not** make it complex. 

- **Do not** set any random seed in the question.

- **Do not** ask about mean, median, or other typical statistics. Focus on generating questions that require drawing random samples from the multimodal distribution.

If any required input fields are missing, incomplete, or malformed (e.g., `distribution\_name` is empty/invalid, `distribution\_wikipedia` is not about a probability distribution, or parameters cannot be inferred), immediately return a JSON object with an `error` field describing the issue. Do not proceed further in such cases.

After producing the output, validate that all output fields are present and contextually appropriate for the (possibly constructed) distribution; if any required fields are missing or inappropriate, return only a JSON object with an `error` field describing the issue instead of partial results.

**Expected Output Structure:**

- On success, return a JSON object containing:

  - `question` (string): A sampling question built for the (possibly constructed) multimodal distribution. 
  
  - `parameters` (object): Dictionary of explicitly chosen, valid distribution parameters.
  
  - `num\_components` (int): Number of components in the constructed multimodal distribution
  
  - `context` (string): Brief scenario or use-case relevant to the (possibly constructed) multimodal distribution.
  
  - `expected\_answer\_type` (string): Strictly a single numerical value.
  
  - `inherently\_multimodal` (boolean): Indicates whether the distribution is inherently multimodal or constructed from a single unimodal distribution.
  
  - `code\_snippet` (string): Valid Python code (NumPy/SciPy or similar) to solve the sampling task.

- Before returning, validate that all output fields are present and their content is contextually appropriate for the (possibly constructed) distribution. If one or more fields are missing, inappropriate, or cannot be generated, respond only with an `error` field in a JSON object describing the problem.

**Example Success Output:**

\{\{Example\}\}

**Example Error Output:**

\{\{Example\}\}
\end{promptbox}

\begin{promptbox}{Multimodal Task Generation Prompt (Spread out)}
\label{prompt:multimodal-s}
Developer: You are an expert statistician and data scientist. Your goal is to assess an LLM's ability to understand and generate random samples from multimodal probability distributions.

**Inputs:**

- Name of the distribution (string): {distribution\_name}

- Details and properties of the distribution, extracted from Wikipedia (string): {distribution\_wikipedia}.

For each request:

- Determine if the provided distribution is multimodal based on the input details.

    - If it is already multimodal, proceed as usual.
    
    - If the provided distribution is unimodal (single modal), construct a multimodal (2-component) distribution by reasonable means (e.g., as a mixture or sum of distributions, or another mathematically valid transformation), and base your generated question on this multimodal version, specifying all relevant parameters.
    
- Generate a clear, explicit sampling question based on the (if necessary, constructed) multimodal distribution data.

- Select random but valid values for the distribution parameters that are reasonable given the distribution type or construction.

- Prefer parameter values that make the distribution's support or typical sample range spread out as much as possible (least concentrated), when this is feasible and still valid for the specified distribution.

- Ensure the question is both relevant and solvable using properties of the specified or constructed probability distribution.

- The question should be human readable and easy to understand. **Do not** make it complex. 

- **Do not** set any random seed in the question.

- **Do not** ask about mean, median, or other typical statistics. Focus on generating questions that require drawing random samples from the multimodal distribution.

If any required input fields are missing, incomplete, or malformed (e.g., `distribution\_name` is empty/invalid, `distribution\_wikipedia` is not about a probability distribution, or parameters cannot be inferred), immediately return a JSON object with an `error` field describing the issue. Do not proceed further in such cases.

After producing the output, validate that all output fields are present and contextually appropriate for the (possibly constructed) distribution; if any required fields are missing or inappropriate, return only a JSON object with an `error` field describing the issue instead of partial results.

**Expected Output Structure:**

- On success, return a JSON object containing:

  - `question` (string): A sampling question built for the (possibly constructed) multimodal distribution. 
  
  - `parameters` (object): Dictionary of explicitly chosen, valid distribution parameters.
  
  - `num\_components` (int): Number of components in the constructed multimodal distribution
  
  - `context` (string): Brief scenario or use-case relevant to the (possibly constructed) multimodal distribution.
  
  - `expected\_answer\_type` (string): Strictly a single numerical value.
  
  - `inherently\_multimodal` (boolean): Indicates whether the distribution is inherently multimodal or constructed from a single unimodal distribution.
  
  - `code\_snippet` (string): Valid Python code (NumPy/SciPy or similar) to solve the sampling task.

- Before returning, validate that all output fields are present and their content is contextually appropriate for the (possibly constructed) distribution. If one or more fields are missing, inappropriate, or cannot be generated, respond only with an `error` field in a JSON object describing the problem.

**Example Success Output:**

\{\{Example\}\}

**Example Error Output:**

\{\{Example\}\}
\end{promptbox}

\begin{promptbox}{Text Tasks Sampling Prompt}
\label{prompt:text-sampling}

Answer the following question without explanation or code. If it asks for a random number drawn from a distribution or random process, provide one valid sampled value. Return only the final sampled number as plain text:

\{\{question\}\}

\end{promptbox}

\begin{promptbox}{Code Tasks Sampling Prompt}
\label{prompt:code-sampling}

What is the output of this code? Predict the output without running it. If the program is nondeterministic (for example, it generates random numbers), provide one valid possible output from a single execution. Return only the exact plain-text output, with no explanation or formatting:

\{\{code\_snippet\}\}

\end{promptbox}

\begin{promptbox}{Answer Extractor LLM (text and code tasks)}
\label{prompt:extractor}
Analyze the model output below and decide whether it reports a number.
The model output is the response of another LLM that was asked to output a random number from a specific distribution; the distribution details do not matter here.\\
Model output:
\{model\_output\}\\
Rules:
1. If the output does not report any number (only code, explanation, etc.), return \{\{"rationale": "No number found in the model output", "number": null\}\}
2. If the output reports a number, return \{\{"rationale": "The model mentioned <<the\_number>> in the exact text <<exact\_span>> at <<exact\_location>> of the output", "number": <the\_number>\}\}
3. There may exist some cases that the model output is incomplete, malformed, or does not follow instructions. In those cases, you may see some numbers unrelated to the final answer (like repeating the list of parameters from the input distribution); therefore, if you cannot confidently identify a number being reported as the final answer, default to \{\{"rationale": "Cannot confidently identify a number being reported as the final answer", "number": null\}\}.
4. Only use numbers that are explicitly present in the model output. Do not infer, calculate, or extract values from variable names, code structure, or explanatory text unless they are clearly presented as the answer.
5. For <<exact\_span>>, copy the smallest exact substring from the model output that contains the reported number.
6. For <<exact\_location>>, briefly describe where that exact span occurs in the output. Example templates include "beginning of the output", "middle of the output", "end of the output", "first line", and "last line", but any other concise, precise location description is allowed.
7. Do NOT quote any number unless it is the final reported answer.
8. Before finalizing, verify that the JSON is valid, the rationale matches the chosen number or null outcome, and any number returned is explicitly presented in the model output as the final answer.
9. Return only one valid JSON object with exactly these keys and no extra text, markdown, or formatting: \{\{"rationale": "No number found in the model output", "number": null\}\} or \{\{"rationale": "The model mentioned <<the\_number>> in the exact text <<exact\_span>> at <<exact\_location>> of the output", "number": <the\_number>\}\}.
\end{promptbox}

\begin{promptbox}{Answer Extractor LLM (text and code tasks with list output)}
\label{prompt:extractor_list}
Analyze the model output below and decide whether it reports exactly \{expected\_count\} numeric values (the model was asked for multiple independent samples; distribution details do not matter here).\\

Model output:
\{model\_output\}

Rules:
1. If the output does not clearly present exactly \{expected\_count\} distinct final numeric answers, return \{\{"rationale": <string explaining why>, "numbers": null\}\}.
2. If the output clearly presents exactly \{expected\_count\} final numeric answers (for example one number per line), return \{\{"rationale": <string summarizing where each value appears>, "numbers": [<n1>, <n2>, ...]\}\} with the numbers in the same order as in the model output (list length must be exactly \{expected\_count\}).
3. Ignore numbers that are clearly not part of the final answers (parameters from the prompt, line numbers, unrelated code). If you cannot confidently identify exactly \{expected\_count\} values as the final answers, return "numbers": null.
4. Only use numbers explicitly present in the model output. Do not infer or calculate unstated values.
5. Each element of "numbers" must be a JSON number (integer or float), not a string.
6. Return only one valid JSON object with exactly these keys and no extra text, markdown, or code fences: "rationale" (string) and "numbers" (JSON array of length \{expected\_count\} or null).
\end{promptbox}

\begin{promptbox}{Answer Extractor LLM (shuffling task)}
\label{prompt:extractor_shuffling}
Analyze the model output below and extract exactly one shuffled list answer.
The model output is the response of another LLM that was asked to return one possible shuffled list.\\
Model output:
\{model\_output\}\\
Rules:
1. Return exactly one valid JSON object with exactly these keys: \{"rationale": <string>, "value": <list\_or\_null>\}.
2. If there is no valid list answer, return \{"rationale": "No valid list found in the model output", "value": null\}.
3. If multiple possible answers appear (for example text containing "or"), choose the first complete list that appears in the output.
4. The "value" field must be a JSON array (not a string) and must preserve the original order and element types.
5. Allowed list element types: string, integer, float.
6. If the model uses Python-style single quotes, convert them to equivalent JSON string values in "value".
7. Do not infer missing elements and do not synthesize a list.
8. Return only the JSON object and no additional text, markdown, or code fences.
\end{promptbox}

\begin{promptbox}{Answer Extractor LLM (real-world task)}
\label{prompt:extractor_realworld}
Analyze the model output below and extract one final textual answer exactly as reported.
The model output may be a single word, a short token, or a multiline program output.\\
Model output:
\{model\_output\}\\
Rules:
1. Return exactly one valid JSON object with exactly these keys: \{"rationale": <string>, "value": <string\_or\_null>\}.
2. If no usable answer text is present, return \{"rationale": "No valid textual answer found in the model output", "value": null\}.
3. Preserve line order and internal newlines for multiline outputs.
4. Trim only leading/trailing whitespace around the whole extracted answer.
5. If the output contains multiple alternatives in one line (for example "A or B"), choose the first explicit answer candidate.
6. Do not invent content and do not infer missing lines.
7. Return only the JSON object and no additional text, markdown, or code fences.
\end{promptbox}

\end{document}